\documentclass[english]{article}
\usepackage{geometry}

\usepackage[T1]{fontenc}
\usepackage[latin9]{inputenc}
\usepackage{bm}
\usepackage{amsmath,mathtools}
\usepackage{amssymb}
\usepackage[unicode=true,
 bookmarks=false,
 breaklinks=false,pdfborder={0 0 1},colorlinks=false]
 {hyperref}
\hypersetup{
colorlinks,citecolor=blue,filecolor=blue,linkcolor=blue,urlcolor=blue}

\makeatletter
\usepackage{amsthm}
\usepackage{comment}
\usepackage{natbib}
\usepackage{booktabs}

\usepackage{graphicx}

\usepackage[linesnumbered,ruled,vlined]{algorithm2e}

\SetCommentSty{mycommfont}
\usepackage{algorithmic}

\usepackage{float}
\usepackage{multirow}
 
\usepackage{dsfont}
\usepackage{tcolorbox}

\usepackage{color}
\definecolor{yxc}{RGB}{255,0,0}
\definecolor{yjc}{RGB}{125,0,0}
\definecolor{ytw}{RGB}{255,69,0}
\definecolor{gen}{RGB}{0,0,200}

\usepackage{bm}
\usepackage{mathtools}
\usepackage{dsfont}
\usepackage{amsmath}
\usepackage{amssymb}
\usepackage{amsfonts}

\usepackage{booktabs}
\usepackage[T1]{fontenc}
\usepackage[latin9]{inputenc}

\usepackage[linesnumbered,ruled,vlined]{algorithm2e}
\usepackage{algorithmic}
\usepackage{multirow}
\usepackage{subcaption}

\usepackage{amsthm}

\usepackage{makecell}
\renewcommand{\paragraph}[1]{\medskip\noindent
{\bf #1}}

\allowdisplaybreaks

\usepackage{color}
\definecolor{yxccolor}{RGB}{255,0,0}
\definecolor{ycjcolor}{RGB}{125,0,0}
\definecolor{ytwcolor}{RGB}{255,69,0}
\definecolor{gencolor}{RGB}{0,0,200}

\title{Provably Efficient Q-learning for Robust Reinforcement Learning}

\author{%
Yuchen Jiao\thanks{Department of Statistics and Data Science, Chinese University of Hong Kong.}
\and
Gen Li\footnotemark[1] 
}
\date{March 2025;~~ Revised: October 2026}

\makeatother

\begin{document}

\theoremstyle{assumption}\newtheorem{assumption}{\textbf{Assumption}}
\newtheorem{thm}{Theorem}
\newtheorem{cor}{Corollary}
\newtheorem{lem}{Lemma}
\newtheorem{rem}{Remark}
\newcommand{\revision}[1]{\textcolor{black}{#1}}

\maketitle

\begin{abstract}
Standard reinforcement learning (RL) algorithms based on Markov Decision Processes (MDPs) are vulnerable to sim-to-real gaps. To address this issue, distributionally robust RL algorithms based on robust MDPs (RMDPs) have been proposed, for ensuring that agents can perform reliably to unpredictable conditions in real-world environments.
Despite of the rich literature, \revision{an efficient model-free method is still lacking} for robust RL, particularly for Markovian data, which is more relevant in practical applications.
In this work, we take a step towards settling this problem by proposing a sample efficient (asynchronous Q-learning) algorithm for robust RL. 
In particular, consider a $\gamma$-discounted infinite-horizon RMDP with state space $\mathcal{S}$ and action space $\mathcal{A}$, where the uncertainty set is specified by the total variation (TV) distance with uncertainty level $\sigma$.
Assuming access to a single trajectory of Markovian samples induced by a behavior policy and the nominal transition kernel, the sample complexity of the proposed Q-learning algorithm --- to achieve an entrywise $\varepsilon$-approximation of the optimal Q-function --- is bounded by 
$$
\frac{1}{\mu_{\mathsf{min}}\varepsilon^2(1-\gamma)^3\max\{1-\gamma,\sigma\}} + \frac{t_{\mathsf{mix}}}{\mu_{\mathsf{min}}(1-\gamma)}
$$
up to logarithmic factors. Here, $t_{\mathsf{mix}}$ and $\mu_{\mathsf{min}}$ denote the mixing time and the minimum state-action occupancy probability of the
sample trajectory, respectively.
This result is nearly minimax optimal, with an additional factor of $\frac{1}{1-\gamma}$.
Moreover, it achieves a significant improvement over existing bounds.
\end{abstract}



\section{Introduction}
\label{sec:intro}

As a fast-growing subfield of artificial intelligence, reinforcement learning (RL) demonstrated impressive success across various domains, including networked systems \citep{qu2022scalable}, trading \citep{park2015adaptive}, operations research \citep{de2003greedy,pan2023adjustable,zhao2021learning}, large language model alignment \citep{ziegler2019fine}, healthcare \citep{fatemi2021medical,liu2019deep}, \revision{and} robotics and control \citep{kober2013reinforcement,mnih2013playing}. Due to the unprecedented dimensionality of the state-action space, enhancing sample, space and computational efficiency has become a central challenge in contemporary RL applications. As a result, significant research efforts have been devoted to developing sample-efficient and memory-efficient algorithms and to characterizing the sample complexity and space complexity \citep{li2021sample,li2024breaking,liminimax2022}
\revision{
    across diverse RL settings, including different algorithm paradigms (model-based \citep{li2024breaking}; model-free \citep{li2021breaking,zhang2021model}) and different data-access regimes (online \citep{zhang2024settling,li2020sample}; offline \citep{li2024settling,rashidinejad2022bridging}). 
}

Despite this progress, practical applications of standard RL techniques can still be greatly affected by the sim-to-real gap or environmental uncertainty \citep{bertsimas2019adaptive,zhang2020robust}. 
For instance, a standard RL agent trained in an ideal environment might fail dramatically when the real-world environment is subject to small adversarial disruptions \citep{klopp2017robust,mahmood2018benchmarking}. Consequently, in addition to maximizing the expected total reward, robustness has emerged as an important goal for an RL agent, especially in high-stakes applications like robotics, autonomous driving, clinical trials, and financial investments. 
To address these challenges, distributionally robust RL \citep{bauerle2022distributionally,cai2016robust,iyengar2005robust,nilim2005robust,xu2010distributionally,shi2024breaking,shi2024sample,jiao2024minimax}, offers a flexible framework by drawing on principles from distributionally robust optimization and supervised learning \citep{bertsimas2018data,blanchet2019quantifying,chen2019distributionally,duchi2021learning,gao2023finite,lam2019recovering,rahimian2019distributionally}.
This approach aims to develop policies capable of maintaining high performance even when the deployed environment deviates from the nominal model due to environmental uncertainty.

\subsection{Challenges}
\label{subsec:challenges}
\revision{
We study robust Markov decision processes (RMDPs) in the discounted infinite-horizon setting, where the learner observes a single trajectory generated by the nominal transition kernel.
Our goal is to develop a model-free algorithm that achieves $O(SA)$ space complexity and a sample complexity comparable to that of standard Q-learning.
The main difficulty is to implement robust Bellman updates using nominal trajectory samples without explicitly estimating the transition kernel.
}

\revision{
To explain this difficulty, consider the standard Q-learning update:
\begin{align}\label{eq:update-qlearning}
Q_{t+1}(s_t,a_t)
=
(1-\eta_t)Q_t(s_t,a_t)
+
\eta_t\bigl(r(s_t,a_t)+\gamma V_t(s_{t+1})\bigr),
\end{align}
where $V_t(s)=\max_{a\in\mathcal{A}}Q_t(s,a)$ and the observed next state satisfies
$s_{t+1}\sim P^0(\cdot\mid s_t,a_t)$.
The sampled target therefore estimates the nominal Bellman update.
In an RMDP, however, the Bellman update involves the smallest expected next-state value over an uncertainty set centered around $P^0$ (cf.~\eqref{eq:def-setU}).
Thus, directly using the observed next state generally does not estimate the desired robust Bellman update.
}

\revision{
Our approach is to replace $s_{t+1}$ with a surrogate next state $s_{t+1}^{\prime}$ drawn from a worst-case transition kernel associated with the value function estimate.
The resulting ideal update is
\begin{align}\label{eq:update-robust-qlearning}
Q_{t+1}(s_t,a_t)
=
(1-\eta_t)Q_t(s_t,a_t)
+
\eta_t\bigl(r(s_t,a_t)+\gamma V_t(s_{t+1}^{\prime})\bigr).
\end{align}
The challenge is to construct such a surrogate state using only observations from the nominal kernel.
Moreover, the worst-case kernel depends on the value function, so the state construction must be adjusted as the value estimate changes.
}

\revision{
Existing approaches address this problem with different resource requirements.
Model-based methods estimate and store the nominal transition kernel and then perform robust planning, requiring $O(S^2A)$ space
\citep{shi2024curious,zhou2021finite,panaganti2022sample,yang2022toward}.
The model-free method in \citet{wang2024sample} reduces the space requirement to $O(SA)$, but relies on generative-model access to query arbitrary state-action pairs.
Such access is not assumed in our trajectory setting.
For Markovian data, \citet{yang2023robust} use stochastic gradient ascent to update a dual variable in a penalized Bellman formulation alongside the Q-function.
Their reported Q-function convergence rate for the $\chi^2$-divergence formulation is $O(t^{-1/3})$, which is significantly slower compared with the $O(t^{-1/2})$ rate of standard Q-learning.
These limitations motivate the following question:
\begin{quote}
Can a model-free algorithm for RMDPs learn from a single trajectory with $O(SA)$ space complexity and a sample complexity comparable to that of standard Q-learning?
\end{quote}
}

\revision{
To answer this question, we address two challenges: constructing surrogate state with worst-case expected value from nominal trajectory data without estimating the transition kernel, and controlling the resulting estimation error when the worst-case kernel changes across stages.
Our approach uses resampling rather than the dual optimization of \citet{yang2023robust}.
Compared with the standard Q-learning analysis of \citet{li2024q}, our analysis needs to additionally account for resampling bias and error propagation under changing worst-case kernels.
We summarize these contributions in Section \ref{subsec:contributions}.
}

\subsection{Contributions}
\label{subsec:contributions}

In this paper, we propose an efficient Q-learning algorithm for RMDPs that integrates several key advantages. Specifically, the algorithm achieves the optimal space complexity of $O(SA)$, and operates effectively in the asynchronous setting in terms of sample complexity.
To the best of our knowledge, this is the first algorithm that combines all these advantages
\revision{and is supported by a rigorous theoretical analysis.}

In particular, assume that we have access to a single trajectory of the Markov decision processes (MDPs) with a transition nominal kernel induced by a stationary behavior policy (also called an asynchronous setting).
Consider an uncertainty set with the shape and size of the ball specified by the total variation (TV) distance between probability distributions and some uncertainty level $\sigma\in(0,1)$.
Let $S$ be the size of the state space, $A$ the size of the action space, $\gamma$ the discount factor. We are interested in how the space complexity and the sample complexity --- the number of samples required for an algorithm to output a Q-function whose entrywise error with the optimal Q-function is at most $\varepsilon$ --- scales with all these salient problem parameters.
Our main contributions are as follows.
\begin{itemize}
\item 
\revision{
    We design an efficient Q-learning algorithm for RMDPs that operates on a single trajectory and achieves \(O(SA)\) space complexity. 
    At each time step, the agent selects an action according to a behavior policy and observes the next state $s_{t+1}\sim P^0(\cdot|s_{t},a_{t})$ according to the nominal transition kernel. We observe that the worst-case transition kernel can be realized by thresholding the value function of \(s_{t+1}\) and redirecting state samples with large values to a fixed state.
    This allows us to construct a surrogate state \(s_{t+1}^{\prime}\) and update the Q-function following the same form as \eqref{eq:update-robust-qlearning}, but with the value function from an earlier stage to avoid using the current Q-estimate when constructing \(s_{t+1}^{\prime}\).
}
The resulting space complexity is \(O(SA)\), which is optimal in the tabular setting where Q-values are explicitly stored.
\revision{Moreover, the algorithm processes samples as they arrive along a single trajectory generated by a fixed behavior policy. It does not require a generative model or the ability to query arbitrary state-action pairs.}
\item We prove that the proposed Q-learning algorithm is sample-efficient.
Specifically, it can find an $\varepsilon$-approximation of the optimal Q-function for RMDPs with sample complexity (modulo log factors) of
$$
\widetilde{O}\left(\frac{1}{\mu_{\mathsf{min}}\varepsilon^2(1-\gamma)^3\max\{1-\gamma,\sigma\}} + \frac{t_{\mathsf{mix}}}{\mu_{\mathsf{min}}(1-\gamma)}\right),
$$
where $t_{\mathsf{mix}}$ denotes the mixing time that characterizes how fast the trajectory converges to stationary, and $\mu_{\mathsf{min}}$ denotes the minimum state-action occupancy probability of the stationary distribution of the trajectory.
\revision{Our theoretical results establish a nearly minimax-optimal sample complexity for model-free robust RL, substantially improve existing bounds, and achieve a sample complexity comparable to that of standard Q-learning.}
As indicated in \citet{shi2024curious}, 
the main term in our bound is nearly minimax-optimal with an additional factor of $\frac{1}{1-\gamma}$,
which is shown unavoidable for Q-learning for standard RL \citep{li2021sample}, and we think it is also unavoidable for robust MDP case.
In addition, our result improves significantly over the existing bound of $\widetilde{O}\left(\frac{1}{\mu_{\mathsf{min}}^9\varepsilon^3(1-\gamma)^{9}}\left(\frac{1}{(1-\gamma)^6}+t_{\mathsf{mix}}^3\right)\right)$ \citep{yang2023robust} by a factor of $\frac{\mu_{\mathsf{min}}^8\varepsilon(1-\gamma)^{12}}{\max\{1-\gamma,\sigma\}}$.
Furthermore, our result is no larger than the sample complexity of Q-learning for standard RL in \citet{li2024q}, $\widetilde{O}\left(\frac{1}{\mu_{\mathsf{min}}\varepsilon^2(1-\gamma)^4} + \frac{t_{\mathsf{mix}}}{\mu_{\mathsf{min}}(1-\gamma)}\right)$.
In particular, if $\sigma\gtrsim 1-\gamma$,
then the sample complexity for RMDPs is improved by a factor of $\sigma/(1-\gamma)$ when $\varepsilon$ is small enough.
\item As a corollary, the proposed algorithm can be naturally extended to the generative model, which assumes access to a simulator (also called a synchronous setting).
In this context, we prove that the sample complexity of the proposed synchronous Q-learning algorithm is $\widetilde{O}\left(\frac{SA}{\varepsilon^2(1-\gamma)^3\max\{1-\gamma,\sigma\}}\right)$ (modulo log factors).
This result is nearly minimax optimal with an additional factor of $\frac{1}{1-\gamma}$ inherent to the design of Q-learning algorithms as stated before.
Moreover, similar insights arise when comparing with synchronous Q-learning for standard RL $\widetilde{O}\left(\frac{SA}{\varepsilon^2(1-\gamma)^4}\right)$ \citep{li2021sample}.
\end{itemize}

\subsection{Related Works}
\label{subsec:related-works}

Robust MDPs have been extensively studied, both empirically and theoretically, under various forms of uncertainty sets \citep{badrinath2021robust,ding2024seeing,goyal2023robust,ho2018fast,ho2021partial,iyengar2005robust,xu2010distributionally,tamar2014scaling}.
However, most existing theoretical studies either focus on planning algorithms \citep{tamar2014scaling,xu2010distributionally,iyengar2005robust} or adopt an asymptotic perspective \citep{roy2017reinforcement}.

Recently, several robust RL algorithms have been developed, focusing on finite-sample performance guarantees under various sampling mechanisms, as show in Table \ref{tab:related-works}.
A line of work focuses on model-based algorithms.
\revision{
In particular, \citet{yang2022toward,panaganti2022sample} studied model-based robust RL algorithms with an \((s,a)\)-rectangular uncertainty set specified by the TV distance, which matches our uncertainty-set specification, and used data collected from a generative model, corresponding to the synchronous setting in this work.
They also considered other uncertainty sets, including those specified by the \(\chi^2\)-divergence or the KL divergence, and \citet{yang2022toward} additionally considered the \(s\)-rectangular case under these measures.
\citet{zhou2021finite} assumed that the reward follows a certain distribution on a bounded set and conducted an offline analysis where states are uniformly sampled by a known policy, focusing on uncertainty sets defined by the KL divergence.
\citet{badrinath2021robust} proposed a policy iteration algorithm based on linear function approximation and analyzed its convergence asymptotically under certain dedicated assumptions on the iterative policy and the induced stationary distribution.
Though an asynchronous setting is considered, its space complexity in the tabular case is $O(S^2A^2)$ to store the feature matrix.
Moreover, \citet{shi2024curious} provided the near-optimal sample complexity bounds for model-based robust RL algorithms with access to a generative model when the uncertainty set is measured by the TV distance or the $\chi^2$-divergence.
Specifically, for a TV-based uncertainty set, the minimax-optimal sample complexity of $\widetilde{O}\left(\frac{SA}{\varepsilon^2(1-\gamma)^2\max\{1-\gamma,\sigma\}}\right)$ is achieved by robust value iteration.
However, these algorithms require at least $O(S^2A)$ space complexity to store the transition probability kernel, which is prohibitive for large-scale problems.
}

Another line of research focuses on memory-efficient algorithms with a reduced space complexity of $O(SA)$ and their finite-sample performance analysis.
\revision{
    \citet{wang2023finite,wang2024sample} considered a randomized reward function and an $(s,a)$-rectangular uncertainty set under the KL divergence with a sufficiently small uncertainty level $\sigma\lesssim p_{\min}$, where $p_{\min}$ denotes the minimum nominal transition probability.
    They reduced the space complexity by collecting a large batch of samples and using them to update the Q-function for one state-action pair \((s,a)\) at a time, then discarding the data before moving on to the next one.
    This design achieves \(O(SA)\) space complexity; however, it implicitly assumes that the learner can query any state-action pair on demand, an assumption that holds under a generative model (synchronous setting) but fails in the asynchronous setting, where samples arrive sequentially along a single trajectory induced by the behavior policy.
In addition, the restriction \(\sigma\lesssim p_{\min}\) on the uncertainty level limits its applicability.
    In contrast, \citet{yang2023robust} proposed an algorithm for the asynchronous setting that performs Q-learning-style stochastic approximation updates on the penalized dual formulation of the RMDP. Their analysis relies on two key assumptions: that the Markov chain induced by the behavior policy is uniformly ergodic, which is also assumed in our work, and that the objective in the Bellman operator is strongly convex at its optimum. 
    Under these conditions, they derived sample complexity bounds for various uncertainty sets, with an explicit bound of $\widetilde{O}\left(\frac{1}{\mu_{\mathsf{min}}^9\varepsilon^3(1-\gamma)^{9}}\left(\frac{1}{(1-\gamma)^6}+t_{\mathsf{mix}}^3\right)\right)$ for the $\chi^2$-divergence. 
    This bound has worse dependence on the accuracy and mixing parameters than ours, although the two results concern different uncertainty sets.
}

\begin{table}
\begin{small}
\centering
\renewcommand\arraystretch{1.5}
\begin{tabular}{c|c|c|c|c}
\toprule
Algorithm & setting & uncertainty set & \makecell{space\\compl.} & sample compl.  \\
\hline
\citet{yang2022toward} & model-based & TV distance & $S^2A$ & $\frac{S^2A}{\varepsilon^2(1-\gamma)^4\sigma^2}$\\
\citet{panaganti2022sample} & model-based & TV distance & $S^2A$ & $\frac{S^2A}{\varepsilon^2(1-\gamma)^4}$\\
\citet{zhou2021finite} & model-based & KL divergence & $S^2A$ & $\frac{C_{P^0}S^2A}{\varepsilon^2(1-\gamma)^2\sigma^2}$\\
\citet{shi2024curious} & model-based & TV distance & $S^2A$ & $\frac{SA}{\varepsilon^2(1-\gamma)^2\max\{1-\gamma,\sigma\}}$\\
\citet{wang2023finite} & synchronous & KL-divergence & $SA$ &$\frac{SA}{\varepsilon^2(1-\gamma)^5\sigma^4p_{\min}^6}$ \\
\citet{wang2024sample} & synchronous & KL-divergence & $SA$ & $\frac{SA}{\varepsilon^2(1-\gamma)^4p_{\min}^3}$\\
\citet{yang2023robust} & asynchronous & \makecell{$\chi^2$-divergence} & $SA$ & $\frac{1}{\mu_{\mathsf{min}}^9\varepsilon^3(1-\gamma)^{9}}\left(\frac{1}{(1-\gamma)^6}+t_{\mathsf{mix}}^3\right)$  \\
\hline
\multirow{2}{*}{\textbf{this work}} & synchronous & TV distance & $SA$ & $\frac{SA}{\varepsilon^2(1-\gamma)^3\max\{1-\gamma,\sigma\}}$\\
\cline{2-5}
 & asynchronous & TV distance & $SA$ & $\frac{1}{\mu_{\mathsf{min}}}\left(\frac{1}{\varepsilon^2(1-\gamma)^3\max\{1-\gamma,\sigma\}}\!+\!\frac{t_{\mathsf{mix}}}{1-\gamma}\right)$\\
\bottomrule
\end{tabular}
\caption{Our results and comparisons with prior art (up to log factors) for robust RL algorithms in terms of setting, uncertainty set, space complexity and the sample complexity.
Here $t_{\mathrm{mix}}$ denotes the mixing time, $\mu_{\mathsf{min}}$ denotes the  smallest state-action occupancy probability of the stationary
distribution of the trajectory, $p_{\min}$ denotes the minimum of nominal transition probability, and $C_{P^0}$ is determined by the nominal transition probability.}
\label{tab:related-works}
\end{small}
\end{table}


\paragraph{Notation and paper organization.} 
Throughout this paper, we denote $\Delta(\mathcal{S})$ as the probability simplex over a set $\mathcal{S}$.
The notation $x =
[x(s,a)]_{(s,a)\in \mathcal{S}\times \mathcal{A}} \in \mathbb{R}^{SA}$ (resp. $x = [x(s)]_{s\in\mathcal{S}} \in \mathbb{R}^{S}$) denotes a vector assigning values to each state-action pair (resp. each state). 
The Hadamard product of two vectors $x$ and $y$ is defined as $x\circ y = [x(s) \cdot y(s)]_{s\in\mathcal{S}}$ or  $x\circ y = [x(s,a) \cdot y(s,a)]_{(s,a)\in\mathcal{S}\times\mathcal{A}}$ depending on the context.
We also define $\mathcal{X} := \{S,A, 1/(1-\gamma),\sigma, 1/\delta, 1/\varepsilon\}$. 
The notation $f(\mathcal{X}) = O(g(\mathcal{X}))$ or $f(\mathcal{X}) \lesssim g(\mathcal{X})$ implies that there exists a universal constant $C > 0$ such that $f(\mathcal{X}) \le Cg(\mathcal{X})$. 
Similarly, $f(\mathcal{X}) \gtrsim g(\mathcal{X})$ indicates that $g(\mathcal{X}) = O(f(\mathcal{X}))$. Additionally, the notation $\widetilde{O}(\cdot)$ follows the same definition as $O(\cdot)$ but omits logarithmic factors.

The remainder of this paper is organized as follows. Section \ref{sec:formulation} provides an overview of discounted infinite-horizon standard MDPs and formulates the RMDP problem.
Section \ref{sec:alg} proposes an asynchronous Q-learning algorithm for robust RL, with its sample complexity presented in Section \ref{sec:mainthm}. 
Section \ref{sec:analysis}  outlines the main steps of the theoretical analysis.
Finally, we conclude the paper in Section \ref{sec:conclusion} with further discussions.
Detailed proofs are included in the appendix.

\section{Problem formulation}
\label{sec:formulation}

In this section, we formulate robust Markov decision processes (RMDPs) within the infinite-horizon setting and present two sampling mechanisms.

\paragraph{Infinite-horizon tabular MDPs.} We first introduce the standard framework of MDPs to facilitate understanding of RMDPs.
A discounted infinite-horizon MDP is represented by a quintuple $\mathcal{M}=
\{ \mathcal{S},\mathcal{A}, \gamma, P, r\}$, 
where $\mathcal{S} = \{1, \cdots, S\}$ and $\mathcal{A} = \{1, \cdots,A\}$ are the state and action spaces, respectively.
The parameter $\gamma\in[0,1)$ represents the \revision{discount} factor, $P : \mathcal{S}\times\mathcal{A} \rightarrow \Delta(\mathcal{S})$ represents the probability transition kernel, and $r : \mathcal{S}\times\mathcal{A} \rightarrow [0,1]$ stands for the reward function, \revision{which is assumed to be deterministic and known.}

More specifically, $P(\cdot|s,a) \in \Delta(\mathcal{S})$ denotes the transition probability vector from state $s$ when action $a$ is taken, and $r(s,a)$ denotes the immediate reward collected in state $s$ when action $a$ is taken.
Moreover, let $\pi : \mathcal{S} \rightarrow \mathcal{A}$ represent a deterministic policy, where $\pi(s)$ specifies the action chosen in state $s$.

To evaluate the cumulative reward, we define the value function $V^{\pi,P}$ for any policy $\pi$ under the transition kernel $P$ as
 \begin{align*}
 V^{\pi,P}(s) \coloneqq \mathbb{E}_{s_{t+1}\sim P(\cdot|s_t,\pi(s_t))}\left[\sum_{t=0}^{\infty}\gamma^tr(s_t,\pi(s_t))|s_0 = s\right],
 \end{align*}
 which is the expected discounted cumulative reward under policy $\pi$ and transition kernel $P$, starting from $s_0=s$.
The action
value function (a.k.a. Q-function) $Q^{\pi,P}(s,a)$ for a policy $\pi$ under the transition kernel $P$ is similarly defined, except that the action $a_0$ is fixed to be $a$:
\revision{
 \begin{align*}
 Q^{\pi,P}(s,a) \coloneqq \mathbb{E}\Big[\sum_{t=0}^{\infty}\gamma^tr(s_t,a_t)\,\big|\,s_0 = s,a_0 = a\Big],\quad \text{where}~s_{t+1}\sim P(\cdot|s_{t},a_{t}), a_{t+1}=\pi(s_{t+1}),t\ge 0.
 \end{align*}
}

\revision{
    For a fixed transition kernel $P$, let $\pi^\star$ be an optimal deterministic stationary policy that maximizes the value function simultaneously over all states. The optimal value function and Q-function are defined by
 \begin{align*}
 \forall s\in\mathcal{S}:\quad V^{\star,P}(s) \coloneqq V^{\pi^{\star},P}(s) = \max_{\pi}V^{\pi,P}(s),\\
 \forall (s,a)\in\mathcal{S}\times\mathcal{A}:\quad Q^{\star,P}(s,a) \coloneqq Q^{\pi^{\star},P}(s,a) = \max_{\pi}Q^{\pi,P}(s,a).
 \end{align*}
 The optimal Q-function satisfies the Bellman optimality equation
\begin{align}\label{eq:Bellman}
Q^{\star,P}=\mathcal{T}(Q^{\star,P}),
\end{align}
where the Bellman operator $\mathcal{T}$ is defined by
\[
\mathcal{T}(Q)(s,a)
\coloneqq r(s,a)
+\gamma\sum_{s'\in\mathcal{S}}
P(s'\mid s,a)\max_{a'\in\mathcal{A}}Q(s',a').
\]
Q-learning approximates this fixed point using sampled transitions.
Given a trajectory
$\{(s_t,a_t,r(s_t,a_t),s_{t+1})\}_{t=0}^{T-1}$
with $s_{t+1}\sim P(\cdot\mid s_t,a_t)$, it updates the visited state-action pair as follows:
    \begin{align}\label{eq:qlearning}
    Q_{t+1}(s_t,a_t) = (1-\eta_t) Q_{t}(s_t,a_t) + \eta_t(r(s_t,a_t) + \gamma V_{t}(s_{t+1})),
    \end{align}
    where $\eta_t\in(0,1)$ is the stepsize,
$V_t(s)=\max_{a'\in\mathcal{A}}Q_t(s,a')$,
and all other entries of $Q_t$ remain unchanged.
Conditional on the history up to the current state-action pair, the sampled target
$r(s_t,a_t)+\gamma V_t(s_{t+1})$ is an unbiased estimate of $\mathcal{T}(Q_t)(s_t,a_t)$.
}

\paragraph{Robust MDPs.}
An RMDP in the infinite-horizon setting is represented by $\mathcal{M}_{\mathsf{rob}} = \{\mathcal{S},\mathcal{A}, \gamma, \mathcal{U}^{\sigma}_{\rho}(P^0), r\}$, where $\mathcal{S}$, $\mathcal{A}$, $\gamma$, $r$ are defined identically to those in standard infinite-horizon tabular MDPs. 
The primary distinction lies in the treatment of the transition kernel: instead of assuming it is fixed, the RMDP models uncertainty through an uncertainty set centered around a nominal transition kernel $P^0: \mathcal{S} \times \mathcal{A} \to \Delta(\mathcal{S})$, where the distance is measured in terms of total variation (TV) distance.
Specifically, given an uncertainty level $\sigma>0$ and the nominal transition kernel $P^0$, the uncertainty set is defined as
\begin{align}\label{eq:def-setU}
\mathcal{U}^\sigma(P^0) := \otimes \mathcal{U}^\sigma(P_{s,a}^0)~~{\rm with}~~\mathcal{U}^\sigma(P_{s,a}^0) = \{P_{s,a}\in\Delta(\mathcal{S}):\rho_{\mathsf{TV}}(P_{s,a},P_{s,a}^0)\le\sigma\},
\end{align}
where $\rho_{\mathsf{TV}}(P_{s,a}, P^0_{s,a})$ denotes the total variation distance between the transition probability distributions $P_{s,a}$ and $P^0_{s,a}$ for state-action pair $(s, a)$. 
This distance is formally expressed as
\begin{align}\label{eq:def-TV}
\rho_{\mathsf{TV}}(P_{s,a}, P^0_{s,a}):=
\frac12\left\|P_{s,a} - P^0_{s,a}\right\|_1
 =\frac12\sum_{s'\in\mathcal{S}} \left|P^0(s'|s, a) - P(s'|s,a)\right|,
\end{align}
which takes values in the interval $[0, 1]$.
Here
\begin{align*}
    P_{s,a} := P(\cdot\,|\,s,a)\in\mathbb{R}^{1\times S},\quad P_{s,a}^0 := P^0(\cdot\,|\,s,a)\in\mathbb{R}^{1\times S}
\end{align*}
denote the transition probability vectors for state-action pair $(s, a)$.
The uncertainty set satisfies the so-called  $(s, a)$-rectangularity \citep{wiesemann2013robust,zhou2021finite}.

In RMDPs, a primary objective is to evaluate the worst-case performance of a policy $\pi$ across all the possible transition kernels in the uncertainty set. 
This evaluation is quantified using the robust value function $V^{\pi,\sigma}$ and the robust Q-function $Q^{\pi,\sigma}$ 
, defined respectively as
\begin{align*}
\forall(s, a) \in \mathcal{S} \times \mathcal{A}:\quad 
V^{\pi,\sigma}(s) \coloneqq \inf_{P\in\mathcal{U}^{\sigma}(P^0)}
V^{\pi,P}(s), \quad Q^{\pi,\sigma}(s,a) \coloneqq \inf_{P\in\mathcal{U}^{\sigma}(P^0)}
Q^{\pi,P}(s,a).
\end{align*}

\paragraph{Optimal robust policy and robust Bellman operator.}
The robust value function $V^{\pi,\sigma}$ and the Q-function $Q^{\pi,\sigma}$ satisfy a robust Bellman consistency equation:
\begin{align*}
Q^{\pi,\sigma}(s,a) 
&= r(s,a) + \gamma\inf_{\mathcal{P}_{s,a}\in\mathcal{U}^{\sigma}(P^0_{s,a})}\mathcal{P}_{s,a}V^{\pi,\sigma}.
\end{align*}
 A policy $\pi^{\star}$ is termed optimal if it maximizes the value function simultaneously across all states and policies. The corresponding optimal value function 
 $V^{\star,\sigma}$ and optimal Q-functions $Q^{\star,\sigma}$ satisfy 
 \begin{align*}
 \forall s\in\mathcal{S}:\quad V^{\star,\sigma}(s) \coloneqq V^{\pi^{\star},\sigma}(s) = \max_{\pi}V^{\pi,\sigma}(s),\\
 \forall (s,a)\in\mathcal{S}\times\mathcal{A}:\quad \revision{Q^{\star,\sigma}(s,a) \coloneqq Q^{\pi^{\star},\sigma}(s,a) = \max_{\pi}Q^{\pi,\sigma}(s,a).}
 \end{align*}
 It is well known that the optimal policy always exists \citep{iyengar2005robust,nilim2005robust}, and the corresponding optimal robust Q-function satisfies the robust Bellman optimality equation:
 \begin{align}\label{eq:robust-Bellman}
 &\qquad\qquad Q^{\star,\sigma} = \mathcal{T}^{\sigma}(Q^{\star,\sigma}), \notag\\
 &\text{where}~ \mathcal{T}^{\sigma}(Q)(s,a)\coloneqq r(s,a) + \gamma\inf_{\mathcal{P}_{s,a}\in\mathcal{U}^{\sigma}(P^0_{s,a})}\mathcal{P}_{s,a}V,\quad V(s)\coloneqq \max_{a'\in\mathcal{A}} Q(s,a').
 \end{align}
 

\paragraph{Sampling mechanisms.} 
The asynchronous setting assumes access to a sample trajectory $\{s_t, a_t, r_t\}_{t=0}^{\infty}$, which is generated based on executing a given policy $\pi_{\mathsf{b}}$, referred to as a behavior policy, in the nominal MDP $\mathcal{M}=\{\mathcal{S},\mathcal{A},\gamma,P^0,r\}$ .
More specifically, the sample $s_{t+1}$ is generated according to
$$
a_{t}\sim\pi_{\mathsf{b}}(\cdot|s_{t}),\quad s_{t+1}\sim P^0(\cdot|s_{t},a_{t}).
$$
Throughout this paper, we make the following assumption to ensure sufficient coverage of the sample trajectory over the state/action space.

\begin{assumption}
    The Markov chain induced by the behavior policy $\pi_{\mathsf{b}}$ is uniformly ergodic (The definition of uniformly ergodic is in Section 2.1 of \citet{paulin2015concentration}).
\end{assumption}

We shall introduce two important quantities used to characterize the sample complexity of asynchronous Q-learning algorithms.
First, the minimum state-action occupancy probability of the sample trajectory is defined as
$$
\mu_{\mathsf{min}}\coloneqq\min_{(s,a)\in\mathcal{S}\times \mathcal{A}} \mu_{\pi_{\mathsf{b}}}(s,a).
$$
where $\mu_{\pi_{\mathsf{b}}}$ denotes the stationary distribution over all state-action pairs of the behavior Markov chain. 
This metric reflects an information bottleneck caused by the least frequently explored state-action pair.
Second, the mixing time of the chain is defined as
$$
t_{\mathsf{mix}}: = \min\left\{t\,|\,\max_{(s_0,a_0)\in\mathcal{S}\times\mathcal{A}}d_{\mathsf{TV}}(P_t(\cdot|s_0, a_0), \mu_{\pi_{\mathsf{b}}})\le\frac14\right\},
$$
where $d_{\mathsf{TV}}(\mu, \nu)$ denotes the TV distance between two measures $\mu$ and $\nu$ (see \eqref{eq:def-TV}), and $P_t(\cdot|s_0, a_0)$ denotes the distribution of $(s_t, a_t)$ when the initial state-action pair is $(s_0, a_0)$. 
The mixing time $t_{\mathsf{mix}}$ represents the time required for the Markov chain to become decorrelated from its initial state.

In contrast, the synchronous setting, also referred to as the generative model, assumes access to a simulator \citep{panaganti2022sample,zhou2021finite} such that: in each iteration $t$, we collect an independent sample $s_t(s,a)\sim P^0(\cdot|s, a)$ for every state-action pair $(s,a)\in\mathcal{S}\times\mathcal{A}$.

\paragraph{Goal.} Given a sample trajectory $\{s_t, a_t, r_t\}_{t=0}^{\infty}$ generated by the behavior policy $\pi_{\mathsf{b}}$, our goal is to approximate the optimal Q-function $Q^{\star,\sigma}$,
and analyze the sample efficiency of the proposed asynchronous Q-learning algorithm.
Furthermore, we extend both the algorithm and its theoretical analysis to the synchronous setting as a corollary.
\section{Algorithm}
\label{sec:alg}

\revision{Standard Q-learning \eqref{eq:qlearning} uses the observed next-state value $V_t(s_{t+1})$ to estimate $P^0_{s_t,a_t}V_t$. The robust Bellman operator \eqref{eq:robust-Bellman} instead minimizes this expectation over the uncertainty set, so the nominal sample generally gives a biased estimate of the robust target. For a given value function $V$, let $P^V(\cdot\mid s,a)$, also denoted by $P^V_{s,a}$, be a worst-case next-state distribution:
\begin{align}\label{eq:def-Pprime}
P^V(\cdot\mid s,a)\in\arg\min_{\mathcal P_{s,a}\in\mathcal U^\sigma(P^0_{s,a})}\mathcal P_{s,a}V.
\end{align}
Ideally, a surrogate state $s_{t+1}'\sim P^{V_t}(\cdot\mid s_t,a_t)$ would make $V_t(s_{t+1}')$ an unbiased estimate of the worst-case expectation. Our task is to approximate this construction from nominal trajectory samples with a limited sample budget and $O(SA)$ memory, without explicitly estimating the transition model.
Our key insight is that, for the TV-based uncertainty set, a worst-case transition kernel $P^{V}(\cdot|s,a)$ can be realized by resampling nominal next states according to thresholds determined by the value function $V$. 
To estimate the resampling parameters for a fixed value function before using them, we adopt a stage-wise implementation. 
We first describe the ideal resampling construction and then develop its online approximation using estimated parameters.}

\subsection{Ideal resampling construction}
\label{subsec:resample}

\revision{
Intuitively, for each $(s,a)\in\mathcal{S}\times\mathcal{A}$, the TV-based uncertainty set \eqref{eq:def-setU} allows a total probability mass of at most $\sigma$ to be redistributed among next states relative to the nominal distribution. The worst-case transition kernel defined in \eqref{eq:def-Pprime} can therefore be constructed by shifting probability mass from high-value states to a minimum-value state. This observation motivates the following resampling procedure, which uses the observed state $x\sim P_{s,a}^0$ to generate a sample from a worst-case transition kernel $x^{\prime}\sim P_{s,a}^V$ for a fixed value function $V$:
\begin{itemize}
\item[1.] Find a threshold \(\tau_{s,a}^{\star}\in[\min_{s\in\mathcal{S}}V(s),\max_{s\in\mathcal{S}}V(s)]\) satisfying
\begin{align}\label{eq:def-tau}
\sum_{u:V(u)> \tau_{s,a}^{\star}}P_{s,a}^{0}(u) < \sigma \le \sum_{u:V(u)\ge \tau_{s,a}^{\star}}P_{s,a}^{0}(u).
\end{align}
Such a $\tau_{s,a}^{\star}$ exists for any $\sigma\in(0,1]$ because $V(x)$ has a discrete distribution with finite support when $x\sim P_{s,a}^0$.
The detailed proof is deferred to Appendix \ref{app:proof-lem-ideal-resample}.
\item[2.] Compute the resampled state \(x^{\prime}\) as follows:
\begin{itemize}
\item If \(V(x) < \tau_{s,a}^{\star}\), set \(x^{\prime}=x\).
\item If \(V(x) = \tau_{s,a}^{\star}\), set \(x^{\prime}=s_{\min}\) with probability
\begin{align}\label{eq:def-psa}
p_{s,a}^{\star}\coloneqq \frac{\sigma-\sum_{u:V(u)> \tau_{s,a}^{\star}}P_{s,a}^{0}(u)}{\sum_{u:V(u)= \tau_{s,a}^{\star}}P_{s,a}^{0}(u)}
\end{align}
and \(x^{\prime}=x\) with probability \(1-p_{s,a}^{\star}\),
where
\begin{align}\label{eq:def-smin}
s_{\min} \coloneqq \arg\min_{s\in\mathcal{S}} V(s).
\end{align}
\item If \(V(x) > \tau_{s,a}^{\star}\), set \(x^{\prime}=s_{\min}\).
\end{itemize}
\end{itemize}
The distribution of $x'$ is a minimizer in \eqref{eq:def-Pprime}, as established in Lemma \ref{lem:ideal-resample}. Throughout the rest of the paper, we choose $P^V(\cdot\mid s,a)$ to be the distribution induced by this ideal resampling procedure.
}

\revision{
\begin{rem}\label{rem:tie}
    Throughout the paper, including all subroutines, we choose $s_{\min}$ to be the state with the largest index among the minimizers of $V$ whenever multiple states minimize the value function. This convention does not affect the Q-function updates, since all minimizers have the same value and the selected state enters each update only through that value.
    \end{rem}
}

\subsection{Proposed algorithm}
\label{subsec:alg}
\paragraph{Resampling with estimated parameters.}
\revision{The nominal transition kernel $P^0$ is unknown to the learner, who observes only a trajectory generated by it. We therefore approximate the ideal construction in Section \ref{subsec:resample} using parameters estimated from trajectory data. The ideal construction requires a threshold $\tau_{s,a}^{\star}$ and a randomization probability $p_{s,a}^{\star}$. However, the tail probability $\sum_{u:V(u)>\tau}P_{s,a}^{0}(u)$ is a step function of $\tau$, so a small threshold estimation error alone need not ensure a small bias in the expected next-state value. We instead estimate an interval $[\tau_{s,a}^{\mathsf{L}},\tau_{s,a}^{\mathsf{R}}]$ together with a resampling probability $p_{s,a}$.}

\revision{Algorithm \ref{alg:subroutine-resample} implements resampling with these estimated parameters. For a given state-action pair $(s,a)$, we suppress the parameter subscripts in the subroutine. Its inputs are a fixed value function $V$, a nominal next state $x$, and parameters $(\tau^{\mathsf{L}},\tau^{\mathsf{R}},p)$ satisfying $\tau^{\mathsf{L}}\le\tau^{\mathsf{R}}$ and $p\in[0,1]$; its output is a surrogate state $x'$. The subroutine retains low-value next states, redirects high-value next states to a minimum-value state, and randomizes between these choices within the threshold interval. Storing two thresholds and one resampling probability for each state-action pair requires $O(SA)$ space, the same as storing the Q-function.}

\begin{algorithm}[t]
\begingroup\color{black}
\SetAlgoLined
\DontPrintSemicolon
\SetAlFnt{\small\linespread{1.12}\selectfont}
\SetAlgoNlRelativeSize{-1}
\SetInd{0.6em}{0.9em}
\SetKwInOut{AlgInput}{Input}
\SetKwInOut{AlgOutput}{Output}
\caption{Resampling with estimated parameters (\textsc{Resample}).}
\label{alg:subroutine-resample}
\AlgInput{Value function $V$, nominal next state $x$, thresholds $\tau^{\mathsf{L}}\le\tau^{\mathsf{R}}$, and resampling probability $p\in[0,1]$.}
\AlgOutput{Surrogate state $x'$.}
Set $s_{\min}=\arg\min_{s\in\mathcal{S}}V(s)$, with ties broken as in Remark \ref{rem:tie}\;
\uIf{$V(x)\le\tau^{\mathsf{L}}$}{
  $x'\leftarrow x$\;
}
\uElseIf{$V(x)>\tau^{\mathsf{R}}$}{
  $x'\leftarrow s_{\min}$\;
}
\Else{
  $x'\leftarrow s_{\min}$ with probability $p$, and $x'\leftarrow x$ with probability $1-p$\;
}
Return $x'$\;
\endgroup
\end{algorithm}

\paragraph{Online estimation of resampling parameters.}
\revision{For a fixed value function $V$, we maintain thresholds $\tau_{s,a}^{\mathsf{L}}$ and $\tau_{s,a}^{\mathsf{R}}$ for each state-action pair $(s,a)$, together with estimates $\widehat{p}_{s,a}^{\mathsf{L}}$ and $\widehat{p}_{s,a}^{\mathsf{R}}$ of the corresponding tail probabilities
\[
\sum_{u:V(u)>\tau_{s,a}^{\mathsf{L}}}P_{s,a}^{0}(u),\qquad
\sum_{u:V(u)>\tau_{s,a}^{\mathsf{R}}}P_{s,a}^{0}(u).
\]
The resampling probability is computed from these estimates as
\begin{align*}
p_{s,a}\coloneqq\begin{cases}
\displaystyle\frac{\sigma-\widehat{p}_{s,a}^{\mathsf{R}}}{\widehat{p}_{s,a}^{\mathsf{L}}-\widehat{p}_{s,a}^{\mathsf{R}}},
& \text{if }\widehat{p}_{s,a}^{\mathsf{L}}>\widehat{p}_{s,a}^{\mathsf{R}}\text{ and }\widehat{p}_{s,a}^{\mathsf{R}}<\sigma,\\
0, & \text{otherwise.}
\end{cases}
\end{align*}}

\revision{We estimate these parameters by bisection, using $N$ nominal samples per state-action pair for each update. The thresholds, endpoint probability estimates, and midpoint are initialized as in Algorithm \ref{alg:asyn}, with the probability accumulator $\widehat{p}_{s,a}^{\mathsf{mid}}$ and sample counter $n_{s,a}$ set to zero. While the threshold interval exceeds the prescribed tolerance,
each observed transition from $(s,a)$ updates the accumulator as follows:
\begin{align}\label{eq:def-hatp}
\widehat{p}_{s,a}^{\mathsf{mid}}\leftarrow
\widehat{p}_{s,a}^{\mathsf{mid}}+\frac{\mathds{1}(V(x)>\tau_{s,a}^{\mathsf{mid}})}{N},
\end{align}
and increases $n_{s,a}$. Here, $\mathds{1}(\cdot)$ denotes the indicator function. The value function and midpoint remain fixed while these $N$ samples are collected. The subroutine suppresses the $(s,a)$ subscripts; each state-action pair maintains its own estimation state.}

\revision{Once $n_{s,a}=N$, Algorithm \ref{alg:subroutine-bisection} updates the endpoints, their probability estimates, and the midpoint, and resets the accumulator and the counter. Bisection stops when $\tau_{s,a}^{\mathsf{R}}-\tau_{s,a}^{\mathsf{L}}\le\varepsilon_{\tau}$, and the resulting thresholds and resampling probability are supplied to Algorithm \ref{alg:subroutine-resample}.}


\begin{algorithm}[t]
\begingroup\color{black}
\SetAlgoLined
\DontPrintSemicolon
\SetAlFnt{\small\linespread{1.12}\selectfont}
\SetAlgoNlRelativeSize{-1}
\SetInd{0.6em}{0.9em}
\SetKwInOut{AlgInput}{Input}
\SetKwInOut{AlgOutput}{Output}
\caption{Resampling parameter update (\textsc{ResampParamsUpd}).}
\label{alg:subroutine-bisection}
\AlgInput{Value function $V$, nominal next state $x$, uncertainty level $\sigma$, parameter tuple $(\tau^{\mathsf{mid}},\widehat{p}^{\mathsf{mid}},\tau^{\mathsf{L}},\tau^{\mathsf{R}},\widehat{p}^{\mathsf{L}},\widehat{p}^{\mathsf{R}})$, counter $n$, and sample count per bisection update $N$.}
\AlgOutput{Updated parameter tuple and counter $n$.}
$\widehat p^{\mathsf{mid}}\leftarrow\widehat p^{\mathsf{mid}}+\frac{\mathds{1}(V(x)>\tau^{\mathsf{mid}})}{N},\quad n\leftarrow n+1$\;
\If{$n=N$}{
\If{$\widehat p^{\mathsf{mid}}\ge\sigma$}{
  $\tau^{\mathsf{L}}\leftarrow\tau^{\mathsf{mid}},\quad\widehat p^{\mathsf{L}}\leftarrow\widehat p^{\mathsf{mid}}$\;
}
\If{$\widehat p^{\mathsf{mid}}\le\sigma$}{
  $\tau^{\mathsf{R}}\leftarrow\tau^{\mathsf{mid}},\quad\widehat p^{\mathsf{R}}\leftarrow\widehat p^{\mathsf{mid}}$\;
}
$\tau^{\mathsf{mid}}\leftarrow\frac12(\tau^{\mathsf{L}}+\tau^{\mathsf{R}}),\quad\widehat p^{\mathsf{mid}}\leftarrow0$\;
Reset $n\leftarrow0$\;
}
Return $(\tau^{\mathsf{mid}},\widehat{p}^{\mathsf{mid}},\tau^{\mathsf{L}},\tau^{\mathsf{R}},\widehat{p}^{\mathsf{L}},\widehat{p}^{\mathsf{R}})$, $n$\;
\endgroup
\end{algorithm}

\paragraph{Asynchronous Q-learning for robust RL.}
\revision{The resampling parameters depend on the value function. We therefore use $K$ stages to keep the target value function fixed during parameter estimation and separate estimation from subsequent use. The value function $V_{k-1}$ becomes available at the end of stage $k-1$, and its resampling parameters are estimated during stage $k$ for use in stage $k+1$. Thus, stage $k$ updates the Q-function using $V_{k-2}$ and the parameters estimated for it in stage $k-1$, while estimating new parameters for $V_{k-1}$.}

\revision{At each iteration in stage $k$, the algorithm observes a nominal next state $s_{t+1}$ and calls $\textsc{Resample}$ (Algorithm \ref{alg:subroutine-resample}) with $V_{k-2}$ and the previously estimated parameters to obtain $s_{t+1}'$. It then updates the Q-function as
\begin{align}\label{eq:alg-update}
Q_{t+1}(s_t,a_t)=(1-\eta_k)Q_t(s_t,a_t)+\eta_k\left(r(s_t,a_t)+\gamma V_{k-2}(s_{t+1}')\right).
\end{align}
The same nominal observation also supports parameter estimation for $V_{k-1}$ through $\textsc{ResampParamsUpd}$ (Algorithm \ref{alg:subroutine-bisection}). At the end of stage $k$, the algorithm computes the resampling probabilities $p_{k,s,a}$ from these estimates and sets
\[
V_k(s)=\max_{a\in\mathcal{A}}Q_{T_k}(s,a),\qquad s\in\mathcal{S}.
\]}

\revision{Given a total budget of $T$ observations, stage $k$ consists of iterations $T_{k-1},\ldots,T_k-1$, where
\begin{align}\label{eq:def-Tk}
T_k=\lfloor (1-\beta)^{K-k}T\rfloor,\quad 1\le k\le K,\qquad T_0=0,
\end{align}
and
\begin{align}\label{eq:def-beta}
\beta=\frac{1-\gamma}{4\log T}.
\end{align}
The number of observations per state-action pair used for each bisection update in stage $k$ is
\begin{align}\label{eq:def-Np}
N_k=\left\lfloor\frac{\mu_{\mathsf{min}}(T_{k}-T_{k-1})}{2\log_2\frac{2}{(1-\gamma)\varepsilon_{\tau}}}\right\rfloor.
\end{align}
Algorithm \ref{alg:asyn} combines the two subroutines and returns $Q_T$. The initialization and stage schedule are further explained in Remark \ref{rem:stage-schedule}.}

\begin{algorithm}[p]
\begingroup
\SetAlgoLined
\DontPrintSemicolon
\SetAlFnt{\small\linespread{1.12}\selectfont}
\renewcommand{\CommentSty}[1]{\textnormal{\textcolor{blue}{\texttt{#1}}}\unskip}
\SetAlgoNlRelativeSize{-1}
\SetInd{0.6em}{0.9em}
\SetKwInOut{AlgInput}{Input}
\SetKwInOut{AlgOutput}{Output}
\caption{Efficient asynchronous Q-learning for robust RL.}
\label{alg:asyn}
\AlgInput{Learning rates $\{\eta_k\}$, total budget $T$, number of stages $K$, tolerance $\varepsilon_\tau$, discount factor $\gamma$, uncertainty level $\sigma$, stage boundaries $\{T_k\}$ from \eqref{eq:def-Tk}, and sample counts $\{N_k\}$ from \eqref{eq:def-Np}.}
\AlgOutput{Q-function $Q_T$.}
Initialize $Q_0\leftarrow0$, $V_{-1},V_0\leftarrow0$, and $\tau_{0,s,a}^{\mathsf{L}},\tau_{0,s,a}^{\mathsf{R}},p_{0,s,a}\leftarrow0$ for all $(s,a)\in\mathcal{S}\times\mathcal{A}$\;
\For{$k=1,\ldots,K$}{
  Initialize the following for all state-action pairs:\;
  $\begin{aligned}[t]
    &\tau_{k,s,a}^{\mathsf{R}}\leftarrow(1-\gamma)^{-1},\quad
      \tau_{k,s,a}^{\mathsf{L}}\leftarrow-0.1,\quad
      \widehat p_{k,s,a}^{\mathsf{L}}\leftarrow1,\quad
      \widehat p_{k,s,a}^{\mathsf{R}}\leftarrow0,\\
    &\tau_{s,a}^{\mathsf{mid}}\leftarrow0.5/(1-\gamma)-0.05,\quad
      \widehat p_{s,a}^{\mathsf{mid}}\leftarrow0,\quad n_{s,a}\leftarrow0
  \end{aligned}$\;
  \For{$t=T_{k-1},\ldots,T_k-1$}{
    \tcp{Transition observation}
    Draw $a_t\sim\pi_{\mathsf b}(s_t)$; observe $r(s_t,a_t)$ and $s_{t+1}\sim P^0(\cdot\mid s_t,a_t)$\;
    \tcp{Resampling}
    $\begin{aligned}[t]
      s_{t+1}'\leftarrow{}&\textnormal{\textsc{Resample}}\bigl(V_{k-2},s_{t+1},
      \tau_{k-1,s_t,a_t}^{\mathsf{L}},\tau_{k-1,s_t,a_t}^{\mathsf{R}},p_{k-1,s_t,a_t}\bigr)
    \end{aligned}$\;
    \tcp{Q-function update}
    $\begin{aligned}[t]
      Q_{t+1}(s_t,a_t)\leftarrow{}&(1-\eta_k)Q_t(s_t,a_t)+\eta_k\bigl(r(s_t,a_t)+\gamma V_{k-2}(s_{t+1}')\bigr)
    \end{aligned}$\;
    \tcp{Online parameter estimation}
        \If{$\tau_{k,s_t,a_t}^{\mathsf{R}}-\tau_{k,s_t,a_t}^{\mathsf{L}}>\varepsilon_\tau$}{
$\begin{aligned}[t]
          &\bigl(\Theta_{s_t,a_t},n_{s_t,a_t}\bigr)\leftarrow\textnormal{\textsc{ResampParamsUpd}}\bigl(
            V_{k-1},s_{t+1},\sigma,\Theta_{s_t,a_t},n_{s_t,a_t},N_k\bigr),
        \end{aligned}$\;
        where $\Theta_{s_t,a_t} = (\tau_{s_t,a_t}^{\mathsf{mid}},\widehat p_{s_t,a_t}^{\mathsf{mid}}, \tau_{k,s_t,a_t}^{\mathsf{L}},\tau_{k,s_t,a_t}^{\mathsf{R}},
            \widehat p_{k,s_t,a_t}^{\mathsf{L}},\widehat p_{k,s_t,a_t}^{\mathsf{R}})$, and assignments to $\Theta_{s_t,a_t}$ update its components accordingly.
        }
  }
  \tcp{Value function update}
  Compute $V_k(s)\leftarrow\max_{a'\in\mathcal A}Q_{T_k}(s,a')$ for all $s\in\mathcal S$\;
  \tcp{Resampling probability computation}
  \For{$(s,a)\in\mathcal S\times\mathcal A$}{
    \eIf{$\widehat p_{k,s,a}^{\mathsf{L}}\le\widehat p_{k,s,a}^{\mathsf{R}}$}{
      $p_{k,s,a}\leftarrow0$\;
    }{
      $p_{k,s,a}\leftarrow(\sigma-\widehat p_{k,s,a}^{\mathsf{R}})/(\widehat p_{k,s,a}^{\mathsf{L}}-\widehat p_{k,s,a}^{\mathsf{R}})$\;
    }
  }
}
\endgroup
\end{algorithm}

\revision{
    \begin{rem}[staging schedule and initialization]\label{rem:stage-schedule}
    Since $V_{-1}=0$, the resampled state does not affect the first-stage Bellman target, so the initial resampling parameters can be set to zero. The bound $0\le V_{k-1}(s)\le(1-\gamma)^{-1}$ ensures that the initial bisection interval $[-0.1,(1-\gamma)^{-1}]$ contains the threshold in \eqref{eq:def-tau}. Its endpoint tail probabilities are exactly $1$ and $0$, respectively, which justifies the initialization of $\widehat{p}_{k,s,a}^{\mathsf{L}}$ and $\widehat{p}_{k,s,a}^{\mathsf{R}}$.
    Under the schedule \eqref{eq:def-Tk}, the first stage has length $T_1$, and the lengths of stages $2,\ldots,K$ increase geometrically. This schedule is motivated by the geometric convergence of exact value iteration: as the target error decreases, later stages use more samples to control the stochastic error.
    \end{rem}
}

\paragraph{Comparison with \citet{yang2023robust}.}
\revision{
Our algorithm differs from that of \citet{yang2023robust} in the Bellman formulation, the update mechanism, and the scope of uncertainty sets.
First, our Bellman operator minimizes the expected next-state value over an uncertainty set, whereas theirs penalizes the divergence from the nominal transition kernel. They show that, for a given initial-state distribution, a suitable penalty coefficient yields the same optimal objective value as the constrained formulation. The two formulations nevertheless lead to different update mechanisms.
Second, their method uses stochastic gradient ascent to update the dual variable alongside Q-function updates.
These dual-variable updates slow the overall convergence because the dual variable must converge faster than the Q-function to control the overall error. In the generative-model setting, this requirement is reflected in an additional inner optimization loop. Our method instead approximates the robust Bellman update using a single resampled next state and online estimates of the resampling parameters. This avoids dual optimization and achieves a smaller sample complexity bound (Table \ref{tab:related-works}).
Third, their framework accommodates general divergences under regularity conditions, whereas our resampling scheme is tailored to TV-based uncertainty sets. Table \ref{tab:related-works} compares their $\chi^2$-divergence formulation with our TV-based formulation. We use their $\chi^2$-divergence result because their paper explicitly provides a convergence rate for this case, but not for the others.
}

\section{Main results}
\label{sec:mainthm}

\subsection{Main results: sample complexity of Algorithm \ref{alg:asyn}}

The sample complexity bound for the proposed efficient Q-learning algorithm in asynchronous setting is summarized as follows.
 
\begin{thm}\label{thm:infinite-horizon}
Let the uncertainty set $\mathcal{U}^{\sigma}(\cdot)$ defined in \eqref{eq:def-setU}.
Consider any uncertainty level $\sigma\in(0,1)$, accuracy $\varepsilon\in(0,1)$, probability $\delta\in(0,1)$, and discounted factor $\gamma\ge 1/2$.
Suppose that the learning rates $\eta_k$, the number of stages $K$, and the threshold accuracy $\varepsilon_\tau$ satisfy
\begin{align}\label{eq:thm-condition}
\eta_k &= \frac{8\log T}{\mu_{\mathsf{min}}\beta T_k+8\log T},\quad
K =  \frac{2\log\frac{C_1}{(1-\gamma)\varepsilon}}{\log\frac{1}{\gamma}} \le \frac{2}{1-\gamma}\log\left(\frac{C_1}{(1-\gamma)\varepsilon}\right),\quad \varepsilon_{\tau} \le c(1-\gamma)\varepsilon.
\end{align}
Then with probability at least
$1-\delta$, Algorithm \ref{alg:asyn} achieves
\begin{align*}
\max_{(s,a)\in\mathcal{S}\times\mathcal{A}}\left|Q_T(s,a) - Q^{\star,\sigma}(s,a)\right|\le \varepsilon,
\end{align*}
as long as the total number of samples satisfy
\begin{align}\label{eq:thm-T}
T \ge \frac{C_2}{\mu_{\mathsf{min}}}\max\left\{\frac{1}{\varepsilon^2(1-\gamma)^3\max\left\{1-\gamma,\sigma\right\}},\frac{t_{\mathsf{mix}}}{1-\gamma}\right\}\log^2T\log\frac{SAT}{\delta}\log\frac{C_1}{(1-\gamma)\varepsilon}\log^2\frac{1}{(1-\gamma)\varepsilon_{\tau}}.
\end{align}
Here, $c$ is a small enough universal constant, and $C_2\gg C_1$ are two large enough universal constants.
\end{thm}

\paragraph{Sample complexity.}
Theorem \ref{thm:infinite-horizon} shows that the total number of samples required for asynchronous robust Q-learning to achieve $\varepsilon$-accuracy is given by
\begin{align}\label{eq:sample-complexity}
\widetilde{O}\left(\frac{1}{\mu_{\mathsf{min}}\varepsilon^2(1-\gamma)^3\max\{1-\gamma,\sigma\}}+\frac{t_{\mathsf{mix}}}{\mu_{\mathsf{min}}(1-\gamma)}\right).
\end{align}
In prior work \citep{shi2024curious}, 
the minimax lower bound for generative model is provided.
Fortunately, it can be generalized to asynchronous case directly by taking the non-uniform state-action pair distributions into account as 
\begin{align}\label{eq:sample-complexity-minimax}
\widetilde{O}\left(\frac{1}{\mu_{\mathsf{min}}\varepsilon^2(1-\gamma)^2\max\{1-\gamma,\sigma\}}\right).
\end{align}
In comparison, the main term of the sample complexity in \eqref{eq:sample-complexity} is nearly minimax optimal, with an additional factor of $\frac{1}{1-\gamma}$.
This additional factor, as shown in \citet{li2021sample}, is unavoidable for Q-learning in standard MDPs.
 While this has not been rigorously proven for robust MDPs, we conjecture that a similar conclusion holds.

In comparison, \citet{yang2023robust} proposed a model-free algorithm for robust RL in the asynchronous setting and established a sample complexity of
$$
\widetilde{O}\left(\frac{1}{\mu_{\mathsf{min}}^9\varepsilon^3(1-\gamma)^{9}}\left(\frac{1}{(1-\gamma)^6}+t_{\mathsf{mix}}^3\right)\right)
$$ 
for uncertainty set based on $\chi^2$-divergence.
However, their convergence rate of $O(t^{-1/3})$, is significantly slower than our result of $O(t^{-1/2})$.
Moreover, our result exhibits a marked improvement in its dependency on $\gamma$ and $\mu_{\mathsf{min}}$.

Comparing \eqref{eq:sample-complexity} with the sample complexity of standard MDPs derived in \citet{li2024q}:
\begin{align}\label{eq:sample-complexity-standard}
    \widetilde{O}\left(\frac{1}{\mu_{\mathsf{min}}\varepsilon^2(1-\gamma)^4}+\frac{t_{\mathsf{mix}}}{\mu_{\mathsf{min}}(1-\gamma)}\right),
\end{align}
we observe that the sample complexity of RMDPs is as least as well as the standard MDPs. 
Specifically, 
when $1-\gamma<\sigma<1$,
the sample complexity of RMDPs is improved than that of standard MDPs by a factor of $\sigma/(1-\gamma)$ when $\varepsilon$ is small enough.
\revision{When $\sigma\le 1-\gamma$, the sample complexity of RMDPs matches that of standard MDPs.
    These findings are consistent with those reported in \citet{shi2024curious}.}

\paragraph{Memory-efficiency.}
As a model-free algorithm, the proposed asynchronous Q-learning achieves a space complexity of $O(SA)$, which is essentially optimal in the tabular setting when Q-values are explicitly stored. 
Specifically, the algorithm maintains an estimate of the Q-function $Q_t$, value function estimates $V_k$ for the three most recent stages, threshold estimates $[\tau_{k,s,a}^{\mathsf{L}}]_{(s,a)\in\mathcal{S}\times\mathcal{A}}$ and $[\tau_{k,s,a}^{\mathsf{R}}]_{(s,a)\in\mathcal{S}\times\mathcal{A}}$,
and probability estimates $[p_{k,s,a}]_{(s,a)\in\mathcal{S}\times\mathcal{A}}$ for the current and the last stages.
The total space complexity is thus in the order of $SA$.
In comparison, although the previous model-based algorithms achieve smaller sample complexity, they suffer a significantly larger space complexity of $O(S^2A)$ due to the need to store and process transition probability matrices.

\paragraph{Comparison with analysis of standard Q-learning.}
\revision{
Compared with the standard Q-learning analysis of \citet{li2024q}, our analysis addresses two additional challenges. First, errors in the estimated resampling parameters cause the distribution of the surrogate state \(s_{t+1}^{\prime}\) to deviate from the target worst-case kernel \(P^{V_{k-2}}\). This introduces resampling bias into the Bellman update, which we isolate in \eqref{eq:proof-temp-12} and bound in Lemma \ref{lem:bound-resample-sample}. Second, the worst-case kernel depends on the value function and changes across stages. The resulting error propagation requires bounds on accumulated variance terms under these changing kernels, developed in Lemmas \ref{lem:var-V-pik-1} --\ref{lem:var-V-pistar-sub}. 
}

\subsection{Extension: sample complexity of synchronous setting}

A simpler scenario than the asynchronous setting is the synchronous setting, where, in each iteration $t$, independent samples $s_{t+1}(s,a)$ are collected for every state-action pair $(s,a)\in\mathcal{S}\times\mathcal{A}$.
In this section, we extend
our algorithm and analysis for asynchronous Q-learning to accommodate the synchronous setting.
The algorithm is shown in Algorithm \ref{alg} in Appendix \ref{subsec:alg-syn}, which also has a space complexity $O(SA)$.
The corresponding sample complexity bound is summarized as below.
The proof follows a similar analysis to that of Theorem \ref{thm:infinite-horizon} and is therefore omitted.

\begin{cor}\label{thm:infinite-horizon-syn}
Let the uncertainty set $\mathcal{U}^{\sigma}(\cdot)$ defined in \eqref{eq:def-setU}.
Consider any uncertainty level $\sigma\in(0,1)$, accuracy $\varepsilon\in(0,1)$, probability $\delta\in(0,1)$, and discounted factor $\gamma\ge 1/2$.
Suppose that the learning rates $\eta_k$, the number of stages $K$, and the threshold accuracy $\varepsilon_\tau$ satisfy
\begin{align}\label{eq:thm-condition-syn}
\eta_k &= \frac{8\log T}{8\log T+\beta T_k},\quad
K =  \frac{2\log\frac{C_1}{(1-\gamma)\varepsilon}}{\log\frac{1}{\gamma}} \le \frac{2}{1-\gamma}\log\left(\frac{C_1}{(1-\gamma)\varepsilon}\right),\quad \varepsilon_{\tau} \le c(1-\gamma)\varepsilon.
\end{align}
Then with probability at least
$1-\delta$, Algorithm \ref{alg} achieves
\begin{align*}
\max_{(s,a)\in\mathcal{S}\times\mathcal{A}}\left|Q_T(s,a) - Q^{\star,\sigma}(s,a)\right|\le \varepsilon,
\end{align*}
as long as the total number of samples
\begin{align}\label{eq:thm-T-syn}
TSA \ge \frac{C_2SA\log^2T}{\varepsilon^2(1-\gamma)^3\max\left\{1-\gamma,\sigma\right\}}\log\frac{SAT}{\delta}\log\frac{C_1}{(1-\gamma)\varepsilon}\log^2\frac{1}{(1-\gamma)\varepsilon_{\tau}}.
\end{align}
Here, $c$ is a small enough universal constant, and $C_2\gg C_1$ are two large enough universal constants.
\end{cor}

Corollary \ref{thm:infinite-horizon-syn} shows that the sample complexity for robust Q-learning in the synchronous setting is given by
$$
\widetilde{O}\left(\frac{SA}{\varepsilon^2(1-\gamma)^3\max\{1-\gamma,\sigma\}}\right).
$$
In comparison with the lower bound in \eqref{eq:sample-complexity-minimax},
this result is nearly minimax optimal with an additional factor of $\frac{1}{1-\gamma}$.
This suboptimality arises because Q-learning is not minimax-optimal for standard MDPs, falling short by a factor of $\frac{1}{1-\gamma}$, as demonstrated by \citet{li2021sample}.
We conjecture that this limitation also applies to  robust MDPs.

For standard MDPs, learning with a generative model requires a sample complexity of $
\widetilde{O}\left(\frac{SA}{\varepsilon^2(1-\gamma)^4}\right)
$
as shown in \citet{li2024q}.
In comparison, learning RMDPs under the TV distance in synchronous setting is no more challenging than learning standard MDPs. Moreover, 
if $1-\gamma < \sigma < 1$, the sample complexity for RMDPs is reduced by a factor of $\sigma / (1-\gamma)$ compared to that of standard MDPs.

\section{Analysis}
\label{sec:analysis}

Before delving into the proof details, we first introduce some notations that shall be used frequently throughout the proof.
\revision{They are summarized in Table \ref{tab:notation}.}

\begin{table}[htbp]
\centering
\caption{Notation summary.}
\label{tab:notation}
\begin{tabular}{c p{10cm} c}
\hline
\textbf{Notation} & \textbf{Explanation} & \textbf{Definition} \\
\hline
$T$        & Total number of iterations & -\\
$T_k$      & Iteration index marking the end of stage $k$ & \eqref{eq:def-Tk} \\
$\beta$    & Parameter controlling the stage schedule & \eqref{eq:def-beta} \\
$K$        & Total number of stages             & - \\
$\eta_k$ & Q-learning stepsize in stage $k$ & \eqref{eq:thm-condition}\\
$Q_t$      & Q-function estimate after iteration $t$ & - \\
$V_{k}$ & Value estimate at the end of stage $k$: $V_k(s)=\max_a Q_{T_k}(s,a)$ for $k\ge 1$ & -\\
$\pi_k$ & Greedy policy with respect to $Q_{T_k}$ for $k\ge 1$; $\pi_0$ is defined separately & \eqref{eq:def-pik}\\
$\tau_{k,s,a}^{\mathsf{L}},\tau_{k,s,a}^{\mathsf{R}},p_{k,s,a}$ & Thresholds and resampling probability estimated during stage $k$ for $(s,a)$ &-\\
$\widehat{p}^{\mathsf{mid}}$ & Estimate of the nominal probability that the next-state value exceeds the current bisection midpoint & \eqref{eq:def-hatp}\\
$\varepsilon_\tau$ & Bisection stops when $\tau_{k,s,a}^{\mathsf{R}}-\tau_{k,s,a}^{\mathsf{L}}\le \varepsilon_\tau$ & \eqref{eq:thm-condition}\\
$N_k$ & Number of samples per state-action pair used for each bisection update in stage $k$ & \eqref{eq:def-Np}\\
$\Lambda_t$ & Diagonal stepsize matrix for iteration $t$ & \eqref{eq:def-Lambda}\\
$P_t$ & Sample transition matrix constructed from the resampled state $s_{t+1}^{\prime}$ & \eqref{eq:def-Pt}\\
${P}_{s,a}^{V}$ & Worst-case transition kernel for value function $V$ at $(s,a)$ & \eqref{eq:def-Pprime}\\
$\overline{P}_{s,a}^{V_{k-2}}$ & Distribution of the resampled state $s_{t+1}^{\prime}$ conditional on $(s_{t},a_{t})=(s,a)$ in stage $k$ & \eqref{eq:def-barP}\\
$P_{\pi}$  & State-action transition matrix induced by kernel $P$ and policy $\pi$ & \eqref{eq:def-Ppi}\\
$P^{\star},P_{\pi}^{\star}$ & Worst-case kernel for the optimal robust value function $V^{\star,\sigma}$, and the corresponding transition matrix under policy $\pi$, respectively & \eqref{eq:def-Pstar}\\
\hline
\end{tabular}
\end{table}

\subsection{Notations}
\label{subsec:notations}

We introduce an empirical transition matrix constructed using \revision{the sample} collected in the $t$-th iteration, denoted by $P_{t+1}\in\{0,1\}^{SA\times S}$, such that 
\begin{align}\label{eq:def-Pt}
P_{t+1}((s_{t},a_{t}),s_{t+1}^{\prime})=1,\qquad \text{and}\quad P_{t+1}((s,a),s^{\prime})=0,\quad \text{for}~(s,a,s') \neq (s_{t},a_{t},s_{t+1}^{\prime}).
\end{align}
Let $\Lambda_t\in\mathbb{R}^{SA\times SA}$ be a diagonal matrix obeying
\begin{align}\label{eq:def-Lambda}
\Lambda_t((s,a),(s,a))\coloneqq\left\{
\begin{array}{ll}
    \eta_k, & {\rm if}~(s,a)=(s_{t-1},a_{t-1})~{\rm and}~T_{k-1}< t\le T_k,\\
    0, &{\rm otherwise}.
\end{array}
\right.
\end{align}
For convenience, we rewrite $\Lambda_t((s,a),(s,a))$ as $\Lambda_t(s,a)$ throughout the proof.
We further introduce two sequences of related quantities defined for any integer $T_k\ge 1$ and $t \ge1$:
\begin{equation}\label{eq:def-LamdatT}
\Lambda_t^{T_k}:=\left\{
\begin{array}{ll}
\Lambda_t\prod_{i=t+1}^{T_k}(I-\Lambda_i), & {\rm if}~t<T_k,\\
\Lambda_t, & {\rm if}~t=T_k,\\
0, & {\rm if}~t>T_k
\end{array}
\right.
\quad {\rm and}
\quad
\Lambda_0^{T_k}:=\prod_{i=1}^{T_k}(I-\Lambda_i).
\end{equation}
Denote policy $\pi_{k}$ for $k\ge 0$ as 
\begin{align}\label{eq:def-pik}
    \pi_{k}(s) = \left\{
    \begin{array}{ll}
    \arg\max_a Q_{T_{k}}(s,a) & k\ge 1,\\
    \arg\max_a Q_{(1-\beta)T_1}(s,a), & k=0.
    \end{array}
    \right.
\end{align}

Define the distribution $\overline{P}_{s,a}^{V_{k-2}}\in\mathbb{R}^{1\times S}$ as the distribution followed by \revision{sample} $s_{t+1}^{\prime}$ when $(s_{t},a_{t})=(s,a)$ at the $k$-th stage, which is given by
\begin{align}\label{eq:def-barP}
\overline{P}_{s,a}^{V_{k-2}}(s) = \left\{
\begin{array}{ll}
P_{s,a}^0(s),     &  {\rm if}~V_{k-2}(s)\le \tau_{k-1,s,a}^{\mathsf{L}}~{\rm and}~s \neq s_{\min},\\
\chi_1,     & {\rm if}~\tau_{k-1,s,a}^{\mathsf{L}} < V_{k-2}(s)\le \tau_{k-1,s,a}^{\mathsf{R}}~{\rm and}~s \neq s_{\min},\\
0, & {\rm if}~V_{k-2}(s)> \tau_{k-1,s,a}^{\mathsf{R}}\\
\chi_2, &{\rm if}~s = s_{\min}.
\end{array}
\right.
\end{align}
where $s_{\min} \coloneqq \arg\min_{s'\in\mathcal{S}} V_{k-2}(s')$.
\revision{
Here if multiple states $s'\in\mathcal{S}$ reach $\min_{s'} V_{k-2}(s')$, then we choose the one with the largest index as we have discussed in Remark \ref{rem:tie}.
}
If $\widehat{p}_{k-1,s,a}^{\mathsf{L}}=\widehat{p}_{k-1,s,a}^{\mathsf{R}}$, which implies that $\tau_{k-1,s,a}^{\mathsf{L}} = \tau_{k-1,s,a}^{\mathsf{R}}$, then the interval $(\tau_{k-1,s,a}^{\mathsf{L}},\tau_{k-1,s,a}^{\mathsf{R}}]$ is empty and we set 
\begin{align*}
\chi_1 = 0,\qquad \chi_2 = \sum_{u:V_{k-2}(u)>\tau_{k-1,s,a}^{\mathsf{R}}}P_{s,a}^{0}(u) +P_{s,a}^{0}(s_{\min}).
\end{align*}
Otherwise we set
\begin{align*}
\chi_1: &= \frac{\widehat{p}_{k-1,s,a}^{\mathsf{L}}-\sigma}{\widehat{p}_{k-1,s,a}^{\mathsf{L}}-\widehat{p}_{k-1,s,a}^{\mathsf{R}}}P_{s,a}^{0}(s),\\
\chi_2: &= \frac{\sigma-\widehat{p}_{k-1,s,a}^{\mathsf{R}}}{\widehat{p}_{k-1,s,a}^{\mathsf{L}}-\widehat{p}_{k-1,s,a}^{\mathsf{R}}}\sum_{u\neq s_{\min}:\tau_{k-1,s,a}^{\mathsf{L}}<V_{k-2}(u)\le \tau_{k-1,s,a}^{\mathsf{R}}}P_{s,a}^{0}(u)  + \sum_{u:V_{k-2}(u)>\tau_{k-1,s,a}^{\mathsf{R}}}P_{s,a}^{0}(u) +P_{s,a}^{0}(s_{\min}).
\end{align*}
Moreover, define $\overline{P}^{V_{k-2}}\in\mathbb{R}^{SA\times S}$ as the matrix collecting $\overline{P}_{s,a}^{V_{k-2}}\in\mathbb{R}^{1\times S}$ as rows.

Define $P^V \in \mathbb{R}^{SA\times S}$ as the matrix representing the probability transition kernel in the uncertainty set $\mathcal{U}^{\sigma}(P^0)$ that leads to the worst-case value for any vector $V\in\mathbb{R}^S$. 
We write $P_{s,a}^V=P^V(\cdot\mid s,a)$ for the $(s,a)$-th row,
as defined in \eqref{eq:def-Pprime}.

Additionally, we define the square probability transition matrix $P_{\pi}\in\mathbb{R}^{SA\times SA}$ induced by a deterministic policy $\pi$ over the
 state-action pairs as follows
 \begin{align}\label{eq:def-Ppi}
 P_{\pi}:=P\Pi_{\pi},
 \end{align}
 where $\Pi_{\pi}\in\{0,1\}^{S\times SA}$ is a projection matrix associated with the deterministic policy $\pi$:
 \begin{align}
 \Pi_{\pi}=\left[
 \begin{array}{cccc}
 e_{\pi(1)}^{\top} & & &\\
 & e_{\pi(2)}^{\top} & &\\
 & & \cdots &\\
 & & & e_{\pi(S)}^{\top}
 \end{array}
 \right],
 \end{align}
with $e_i\in\mathbb{R}^{A}$ denotes the $i$-th standard basis vector.
In addition, we denote $P_{\star}^{V^{\star,\sigma}}=P_{\pi^\star}^{V^{\star,\sigma}}$ for ease of notation, where $\pi^\star$ is the optimal policy with respect to the robust value function $V^{\star,\sigma}$.
For any matrix $P\in\mathbb{R}^{SA\times S}$,$P_{\pi}\in\mathbb{R}^{SA\times SA}$, value function $V\in\mathbb{R}^{S\times 1}$, and Q-function $Q\in\mathbb{R}^{SA\times 1}$, we define the variance $\mathsf{Var}_PV\in\mathbb{R}^{1\times SA}$ 
and $\mathsf{Var}_{P_{\pi}}Q\in\mathbb{R}^{1\times SA}$ 
as
\begin{align}
(\mathsf{Var}_{P}V)(s,a) &= \sum_{s'\in\mathcal{S}} P_{s,a}(s') V(s')^2-\big(\sum_{s'\in\mathcal{S}} P_{s,a}(s') V(s')\big)^2,\label{eq:def-varPV}\\
(\mathsf{Var}_{P_{\pi}}Q)(s,a) &= \sum_{(s',a')\in\mathcal{S}\times\mathcal{A}} P_{\pi}(s',a'|s,a) Q(s',a')^2\notag\\
&\quad -\Big(\sum_{(s',a')\in\mathcal{S}\times\mathcal{A}} P_{\pi}(s',a'|s,a) Q(s',a')\Big)^2.
\end{align}

Furthermore, we make use of the following short-hand notation: 
\begin{align}\label{eq:def-Pstar}
    P^{\star} = P^{V^{\star,\sigma}},\quad P^{\star}_{\pi} = P^{V^{\star,\sigma}}_{\pi}.
\end{align}  
Let $\left\|\cdot\right\|_{\infty}$ denote the infinity-norm of a vector,
$\left\|\cdot\right\|$ denote the operator norm of a matrix,  
and $\mathsf{1}$ denote an all-ones vector with appropriate length.
In expressions involving addition or subtraction of a scalar and a vector, the scalar is implicitly multiplied by $\mathsf{1}$.

\subsection{Key lemmas}
\label{subsec:key-lems}

\revision{
    The following lemmas connect the ideal resampling construction to its online implementation. We first verify that ideal resampling generates samples from a worst-case transition kernel, and then quantify how parameter estimation affects the expectation and variance of the resampled value.
}

\revision{
    The first lemma formalizes the correctness of the ideal construction.
The proof is deferred to Appendix \ref{app:proof-lem-ideal-resample}.
\begin{lem}\label{lem:ideal-resample}
    For a fixed value function $V$ and state-action pair $(s_t,a_t)\in\mathcal{S}\times\mathcal{A}$, let $s_{t+1}^{\prime}$ be generated by the ideal resampling procedure in Section \ref{subsec:resample}, using a nominal sample drawn from $P^0(\cdot\mid s_t,a_t)$.
    Then its distribution satisfies
    \begin{align}\label{eq:law-stprime}
\mathcal{L}(s_{t+1}^{\prime})\in
\arg\min_{\mathcal{P}\in\mathcal{U}^{\sigma}(P^0_{s_t,a_t})}\mathcal{P}V.
    \end{align}
\end{lem}
Throughout the subsequent analysis, we choose $P^V(\cdot\mid s_t,a_t)$
to be the distribution induced by this ideal resampling procedure.
The lemma therefore justifies using the resampled value as a one-sample estimate of the worst-case expected next-state value.
}

\revision{
    The implemented algorithm in Section \ref{subsec:alg} uses estimated
resampling parameters and therefore approximates the ideal construction.
The following lemma quantifies how the resulting discrepancy between
$\overline{P}^{V_k}$ and $P^{V_k}$ affects the expectation and variance
of the resampled value.
}
The proof is deferred to Appendix \ref{subsec:proof-lem-bound-alpha}.
\begin{lem}\label{lem:bound-alpha}
Assume that $\max_{s\in\mathcal{S}} V_{k}(s) - \min_{s\in\mathcal{S}} V_{k}(s) \le B$,
and the parameter $N_{k+1}$ in \eqref{eq:def-Np} satisfies
\begin{align*}
    N_{k+1} \ge \frac{\mu_{\mathsf{min}}\beta T_1}{2\log_2\frac{2}{(1-\gamma)\varepsilon_\tau}}\ge \frac{C\theta_p}{\max\{1-\gamma,\sigma\}^2},
\end{align*}
where $C$ is a sufficiently large universal constant.
For any $\delta\in(0,1)$, with probability at least $1-\delta$, 
\revision{the expectation and variance of the value function under the distribution \(\overline{P}^{V_{k}}\) deviate from those under \({P}^{V_{k}}\) by at most}
\begin{align*}
\left\|\left(\overline{P}^{V_{k}}-{P}^{V_{k}}\right)V_{k}\right\|_{\infty} &\lesssim \sqrt{\frac{\theta_p\max\{1-\gamma,\sigma\}}{N_{k+1}}}B + \varepsilon_\tau,\\
\left\|\mathsf{Var}_{\overline{P}^{V_{k}}}V_{k} -\mathsf{Var}_{P^{V_{k}}}V_{k}\right\|_{\infty} &\lesssim \sqrt{\frac{\theta_p\max\{1-\gamma,\sigma\}}{N_{k+1}}}B^2 + \varepsilon_\tau B,
\end{align*}
where 
\begin{align}
    \theta_p = \log\left(\frac{2SAK}{\delta}\log\left(\frac{1}{(1-\gamma)\varepsilon_\tau}\right)\right).
\end{align}
\end{lem}

\revision{Complementing the resampling error in Lemma \ref{lem:bound-alpha}, the following lemma provides an upper bound on \(\max_{s\in\mathcal{S}} V_{k}(s)-\min_{s\in\mathcal{S}} V_{k}(s)\), which serves as the quantity \(B\) in Lemma \ref{lem:bound-alpha}. The proof is deferred to Appendix~\ref{subsec:proof-lem-upper-bound-V}.}
\begin{lem}\label{lem:upper-bound-V}
For any $k\ge 1$,  with probability at least $1-\delta$,
\revision{the ranges of value functions and Q-functions satisfy}
    \begin{align}
    \max_{s\in\mathcal{S}} V^{\star,\sigma}(s) - \min_{s\in\mathcal{S}} V^{\star,\sigma}(s) \le  \max_{(s,a)\in\mathcal{S}\times\mathcal{A}} Q^{\star,\sigma}(s,a) - \min_{(s,a)\in\mathcal{S}\times\mathcal{A}} Q^{\star,\sigma}(s,a) \lesssim\frac{1}{\max\{\sigma,1-\gamma\}},\label{eq:lem-upper-bound-V-star}\\
        \max_{s\in\mathcal{S}} V_{k}(s) - \min_{s\in\mathcal{S}} V_{k}(s) \le  \max_{(s,a)\in\mathcal{S}\times\mathcal{A}} Q_{T_k}(s,a) - \min_{(s,a)\in\mathcal{S}\times\mathcal{A}} Q_{T_k}(s,a)\lesssim\frac{1}{\max\{\sigma,1-\gamma\}}.\label{eq:lem-upper-bound-V-k}\\
        \max_{(s,a)\in\mathcal{S}\times\mathcal{A}} Q_{(1-\beta)T_1}(s,a) - \min_{(s,a)\in\mathcal{S}\times\mathcal{A}} Q_{(1-\beta)T_1}(s,a) \lesssim\frac{1}{\max\{\sigma,1-\gamma\}}.\label{eq:lem-upper-bound-Q-T0}
    \end{align}
\end{lem}

Finally, the following lemma relates the stage boundaries $T_k$ to the
total sample budget $T$ and will be used to bound the subsequent error terms.
Its proof is deferred to Appendix~\ref{subsec:proof-lem-frac-k-Tk}.
\begin{lem}\label{lem:frac-k-Tk}
For $T_k$ defined in \eqref{eq:def-Tk} and $T$ satisfying \eqref{eq:thm-T}, the following properties hold.
\begin{align}
(1-\beta)T_1 &\ge \frac{1}{{\rm e}}T-1\ge \frac{32\log^2 T}{\mu_{\mathsf{min}}(1-\gamma)},\label{eq:lem-bound-T1}\\
\max_{1\le k\le K}\frac{k}{T_k}&\le \frac{K}{T} \le \frac{2\log\frac{C_1}{(1-\gamma)\varepsilon}}{(1-\gamma)T}.\label{eq:lem-frac-k-Tk}
\end{align}
\end{lem}

\subsection{Proof of Theorem \ref{thm:infinite-horizon}}

We are now positioned to describe how to establish Theorem \ref{thm:infinite-horizon}.
\revision{
    Lemma \ref{lem:ideal-resample} establishes the correctness of the
ideal resampling construction. Lemmas \ref{lem:bound-alpha} and
\ref{lem:upper-bound-V} then control the expectation and variance
errors introduced by its approximate implementation. Building on
these guarantees, we establish convergence in four steps. 
Step 1 decomposes the Q-function estimation error into propagated
estimation error, resampling bias, and sampling noise.
Step 2 uses Lemmas \ref{lem:bound-alpha} and \ref{lem:upper-bound-V} to control the resampling bias and the sampling noise within a single stage.
Step 3 bounds the cumulative sampling noise under changing worst-case kernels.
Step 4 combines these bounds to establish
the sample complexity guarantee.
}

Before diving into the proof details, we first express the Q-learning update rule \eqref{eq:alg-update} using the matrix notations defined in Section \ref{subsec:notations} as below.
\begin{align}\label{eq:alg-proof-update}
Q_{t+1} = (I-\Lambda_{t+1})Q_{t} + \Lambda_{t+1}(r + \gamma P_{t+1}V_{k-2}),\quad T_{k-1}\le t< T_k,
\end{align}
\revision{where $P_{t+1}$ is defined with respect to the resampled next state $s_{t+1}^{\prime}$ (cf. \eqref{eq:def-Pt}).}

\paragraph{Step 1: decomposition the Q-function estimation error.}
Starting from the update rule \eqref{eq:alg-proof-update}, we obtain the following elementary decomposition of Q-function estimation error for \(T_{k-1}<t\le T_k\):
\begin{align*}
Q_{t} - Q^{\star,\sigma} 
&= (I-\Lambda_t)\left(Q_{t-1}-  Q^{\star,\sigma}\right)+ \Lambda_t(r + \gamma P_tV_{k-2}) - \Lambda_t\left(r + \gamma P^{\star} V^{\star,\sigma}\right)\nonumber\\
&= (I-\Lambda_t)\left(Q_{t-1}-  Q^{\star,\sigma}\right)\nonumber\\
&\quad+ \gamma\Lambda_t\left( \left(P_t-\overline{P}^{V_{k-2}}\right)V_{k-2} + \left(\overline{P}^{V_{k-2}} - {P}^{V_{k-2}}\right)V_{k-2} + {P}^{V_{k-2}}V_{k-2}-P^{\star}V^{\star,\sigma}\right).
\end{align*}
Applying recursion on $t$ from $T_{k-1}+1$ to $T_k$ yields
\begin{align}\label{eq:proof-temp-12}
Q_{T_k} - Q^{\star,\sigma}&=\underbrace{\prod_{t=T_{k-1}+1}^{T_k}(I-\Lambda_t)\left(Q_{T_{k-1}}-  Q^{\star,\sigma}\right)+ \gamma\sum_{t=T_{k-1}+1}^{T_k}\Lambda_t^{T_k}\left({P}^{V_{k-2}}V_{k-2}-P^{\star}V^{\star,\sigma}\right)}_{\mathrm{error~propagation~term}}\notag\\
&\quad + \gamma\sum_{t=T_{k-1}+1}^{T_k}\Lambda_t^{T_k}\bigg( \underbrace{\left(P_t-\overline{P}^{V_{k-2}}\right)V_{k-2}}_{\mathrm{sampling~noise}} + \underbrace{\left(\overline{P}^{V_{k-2}} - {P}^{V_{k-2}}\right)V_{k-2}}_{\mathrm{resampling~bias}}\bigg).
\end{align}
\revision{
On the right-hand side of \eqref{eq:proof-temp-12}, the first term is the propagation term, which carries the estimation error from previous stages into the current one.
The second term is the sampling noise, which arises from approximating the expectation by a single sample at each iteration. The third term is the resampling bias, which stems from the distributional mismatch between the distribution of the surrogate state and the worst-case transition kernel. Notably, the appearance of resampling bias term is one of the main differences from the standard Q-learning analysis.
}

\revision{
The propagation term involves both the value-function error and the discrepancy between the historical worst-case distribution \(P^{V_{k-2}}\) and the optimal one \(P^{\star}\).
To enable recursion, we seek to bound it in terms of the Q-function estimation error. This is achieved by recalling that \(P^{\star}\) minimizes the value function \(V^{\star,\sigma}\), which yields
\begin{align}
{P}^{V_{k-2}}V_{k-2}-P^{\star}V^{\star,\sigma}
&\overset{\text{}}{\ge} {P}^{V_{k-2}}V_{k-2}-P^{V_{k-2}}V^{\star,\sigma}\nonumber\\
&\overset{\text{(i)}}{=} {P}^{V_{k-2}}_{\pi_{k-2}}Q_{T_{k-2}}-P^{V_{k-2}}_{\pi^{\star}}Q^{\star,\sigma}\nonumber\\
&\overset{\text{(ii)}}{\ge} {P}^{V_{k-2}}_{\pi^{\star}}Q_{T_{k-2}}-P^{V_{k-2}}_{\pi^{\star}}Q^{\star,\sigma}\nonumber\\
&={P}^{V_{k-2}}_{\pi^{\star}}\left(Q_{T_{k-2}}-Q^{\star,\sigma}\right),\label{eq:proof-temp-10}
\end{align}
where (i) follows from the definitions of \({P}^{V_{k-2}}_{\pi_{k-2}}\) and \(P^{V_{k-2}}_{\pi^{\star}}\) in \eqref{eq:def-Ppi}, and (ii) follows from \(Q_{T_{k-2}}(s,\pi_{k-2}(s))\ge Q_{T_{k-2}}(s,\pi^{\star}(s))\), by the definition of \(\pi_{k-2}\). Similarly, we bound it in the reverse direction:
\begin{align}
{P}^{V_{k-2}}V_{k-2}-P^{\star}V^{\star,\sigma}
&\overset{\text{(i)}}{\le} {P}^{\star}V_{k-2}-P^{\star}V^{\star,\sigma}\nonumber\\
&= {P}^{\star}_{\pi_{k-2}}Q_{T_{k-2}}-P^{\star}_{\pi^{\star}}Q^{\star,\sigma}\nonumber\\
&\overset{\text{(ii)}}{\le} {P}^{\star}_{\pi_{k-2}}Q_{T_{k-2}}-P^{\star}_{\pi_{k-2}}Q^{\star,\sigma}\nonumber\\
&={P}^{\star}_{\pi_{k-2}}\left(Q_{T_{k-2}}-Q^{\star,\sigma}\right),\label{eq:proof-temp-11}
\end{align}
where (i) and (ii) 
follow from the definition of \(P^{V_{k-2}}\) and from $Q^{\star,\sigma}(s,\pi^{\star}(s))\ge Q^{\star,\sigma}(s,\pi_{k-2}(s))$, respectively.
}

\paragraph{Step 2: controlling the resampling bias and sampling noise in a single stage.}
\revision{
The following lemma controls the sampling noise and the resampling bias in the $k$-th stage with high probability.
Its proof is deferred to Appendix \ref{subsec:proof-lem-bound-resample-sample}.
\begin{lem}\label{lem:bound-resample-sample}
    For any $\delta\in(0,1)$, with probability at least $1-\delta$, the sampling noise and the resampling bias terms in the right-hand-side of \eqref{eq:proof-temp-12} for $k\ge 2$ satisfy
    \begin{align}
    \Big|\gamma\sum_{t=T_{k-1}+1}^{T_k}\Lambda_t^{T_k}\left(P_t-\overline{P}^{V_{k-2}}\right)V_{k-2}\Big| &\lesssim \sqrt{\log\frac{SAT}{\delta}}\sqrt{\frac{\mathsf{Var}_{\overline{P}^{V_{k-2}}}V_{k-2}\log^2T}{\mu_{\mathsf{min}}(1-\gamma)T_k}} \notag\\
    &\qquad\qquad+ \frac{\log\frac{SAT}{\delta}\log^2 T}{\mu_{\mathsf{min}}(1-\gamma)\max\{1-\gamma,\sigma\}T_k},\\
     \Big\|\gamma\sum_{t=T_{k-1}+1}^{T_k}\Lambda_t^{T_k}\left(\overline{P}^{V_{k-2}} - {P}^{V_{k-2}}\right)V_{k-2}\Big\|_{\infty} &\lesssim \sqrt{\frac{\theta_p}{N_{k-1}\max\{1-\gamma,\sigma\}}} + \varepsilon_{\tau},
    \end{align}
    where the variance $\mathsf{Var}_{\overline{P}^{V_{k-2}}}V_{k-2}$ is defined in \eqref{eq:def-varPV}.
    For $k=1$, both of these two error terms are zero since $V{-1}=0$.
\end{lem}
}

\revision{
Inserting \eqref{eq:proof-temp-10} and \eqref{eq:proof-temp-11} into \eqref{eq:proof-temp-12}, and combining the result with Lemma \ref{lem:bound-resample-sample} and a recursion on \(t\), we obtain the following bound on the Q-function estimation error, which is decomposed into the initialization error, the propagation of sampling noise, including the variance term and the higher-order terms, and the propagation of the resampling bias across stages.
}
The proof is deferred to Section~\ref{subsec:proof-lem-decom}.
\begin{lem}\label{lem:decom}
For any $0\le k\le K$, denote $k_0=1$ for odd $k$ and $k_0 = 0$ for even $k$. The following inequalities hold:
\begin{small}
\begin{align}
&\quad Q_{\max\{T_k,(1-\beta)T_1\}} - Q^{\star,\sigma} \nonumber\\
&\gtrsim  -\underbrace{\sqrt{\log\frac{SAT}{\delta}}\sqrt{\frac{\log^2T}{\mu_{\mathsf{min}}(1-\gamma)}}\sum_{j=1}^{\lfloor k/2\rfloor}\gamma^{{\lfloor k/2\rfloor}-j}\prod_{i=j+1}^{\lfloor k/2\rfloor}P_{\pi^{\star}}^{V_{2i+k_0-2}}\sqrt{\frac{\mathsf{Var}_{\overline{P}^{V_{2j+k_0-2}}}V_{2j+k_0-2}}{T_{2j+k_0}}}}_{\mathrm{cumulative~sampling~noise(variance)}}\nonumber\\
&- \underbrace{\frac{\log\frac{SAT}{\delta}\log^2 T}{\mu_{\mathsf{min}}(1-\gamma)^2\max\{1-\gamma,\sigma\}T_k}}_{\mathrm{cumulative~sampling~noise(higher~order)}} - \underbrace{\frac{\gamma^{\lfloor k/2\rfloor}}{1-\gamma}}_{\mathrm{initialization~error}} - \underbrace{\sqrt{\frac{\theta_p\log T\log\frac{2}{(1-\gamma)\varepsilon_\tau}\log\frac{C_1}{(1-\gamma)\varepsilon}}{{\mu_{\mathsf{min}}(1-\gamma)^3\max\{1-\gamma,\sigma\} T}}} - \frac{\varepsilon_{\tau}}{1-\gamma}}_{\mathrm{cumulative~resampling~bias}},\label{eq:proof-thm1-decom-upper-lem}\\
&\quad Q_{\max\{T_k,(1-\beta)T_1\}} - Q^{\star,\sigma} \nonumber\\
&\lesssim  \underbrace{\sqrt{\log\frac{SAT}{\delta}}\sqrt{\frac{\log^2T}{\mu_{\mathsf{min}}(1-\gamma)}}\sum_{j=1}^{\lfloor k/2\rfloor}\gamma^{{\lfloor k/2\rfloor}-j}\prod_{i=j+1}^{\lfloor k/2\rfloor}P_{\pi_{2i+k_0-2}}^{\star}\sqrt{\frac{\mathsf{Var}_{\overline{P}^{V_{2j+k_0-2}}}V_{2j+k_0-2}}{T_{2j+k_0}}}}_{\mathrm{cumulative~sampling~noise(variance)}}\nonumber\\
&+ \underbrace{\frac{\log\frac{SAT}{\delta}\log^2 T}{\mu_{\mathsf{min}}(1-\gamma)^2\max\{1-\gamma,\sigma\}T_k}}_{\mathrm{cumulative~sampling~noise(higher~order)}}+ \underbrace{\frac{\gamma^{\lfloor k/2\rfloor}}{1-\gamma}}_{\mathrm{initialization~error}} + \underbrace{\sqrt{\frac{\theta_p\log T\log\frac{2}{(1-\gamma)\varepsilon_\tau}\log\frac{C_1}{(1-\gamma)\varepsilon}}{{\mu_{\mathsf{min}}(1-\gamma)^3\max\{1-\gamma,\sigma\} T}}} + \frac{\varepsilon_{\tau}}{1-\gamma}}_{\mathrm{cumulative~resampling~bias}}.\label{eq:proof-thm1-decom-lower-lem}
\end{align}
\end{small}
\end{lem}

\paragraph{Step 3: controlling the propagation of the sampling noise.}
At the beginning, we define the following notation
\begin{align}\label{eq:def-Delta}
    \Delta_j = \left\{
    \begin{array}{ll}
    \left\|Q_{T_j} - Q^{\star,\sigma}\right\|_{\infty},& j\ge 1,\\
    \max\left\{\left\|Q_{0} - Q^{\star,\sigma}\right\|_{\infty},\left\|Q_{(1-\beta)T_1} - Q^{\star,\sigma}\right\|_{\infty}\right\}, &j =0.
    \end{array}
    \right.
\end{align}
\revision{
In this step, we analyze the cumulative sampling noise terms in \eqref{eq:proof-thm1-decom-upper-lem} and \eqref{eq:proof-thm1-decom-lower-lem} separately. The key challenge is to bound the variance term, since the robust Bellman operator alters the error propagation structure. In particular, the transition kernel used for propagation differs from the one used for computing the variance, which distinguishes the analysis of this term from that of standard Q-learning.
}

\paragraph{Step 3.1: analysis of the cumulative sampling noise(variance) term in \eqref{eq:proof-thm1-decom-upper-lem}.}
\revision{
The distribution $\overline{P}^{V_{2j+k_0-2}}$ used for computing the variance depends on the estimated parameters $\tau_{2j+k_0-1,s,a}^{\mathsf{L}}$ and $\tau_{2j+k_0-1,s,a}^{\mathsf{R}}$, which makes it difficult to analyze directly. To handle this, we further decompose the first term in the right-hand-side of \eqref{eq:proof-thm1-decom-upper-lem} into the cumulative sum of variances under the distribution ${P}^{V_{2j+k_0-2}}$, plus the resampling error in terms of variance:
}
\begin{align}
&\quad\sum_{j=1}^{\lfloor k/2\rfloor}\gamma^{{\lfloor k/2\rfloor}-j}\prod_{i=j+1}^{\lfloor k/2\rfloor}P_{\pi^{\star}}^{V_{2i+k_0-2}}\sqrt{\frac{\mathsf{Var}_{\overline{P}^{V_{2j+k_0-2}}}V_{2j+k_0-2}}{T_{2j+k_0}}}\nonumber\\
&\le \sum_{j=1}^{\lfloor k/2\rfloor}\gamma^{{\lfloor k/2\rfloor}-j}\prod_{i=j+1}^{\lfloor k/2\rfloor}P_{\pi^{\star}}^{V_{2i+k_0-2}}\sqrt{\frac{\mathsf{Var}_{{P}^{V_{2j+k_0-2}}}V_{2j+k_0-2}}{T_{2j+k_0}}}\nonumber\\
&\quad+ \sum_{j=1}^{\lfloor k/2\rfloor}\frac{\gamma^{{\lfloor k/2\rfloor}-j}}{\sqrt{T_{2j+k_0}}}\sqrt{\left\|\mathsf{Var}_{\overline{P}^{V_{2j+k_0-2}}}V_{2j+k_0-2}-\mathsf{Var}_{{P}^{V_{2j+k_0-2}}}V_{2j+k_0-2}\right\|_\infty}.\label{eq:proof-decom-var-V-pik}
\end{align}
\revision{
Here, the error propagation matrix in the first term, given by $P_{\pi^{\star}}^{V_{2i+k_0-2}}$, still differs from the variance distribution ${P}^{V_{2j+k_0-2}}_{\pi_{2j+k_0-2}}$.
The latter involves the policy \(\pi_{2j+k_0-2}\) because the variance is taken with respect to the Q-function:
$$
\mathsf{Var}_{{P}^{V_{2j+k_0-2}}}V_{2j+k_0-2} = \mathsf{Var}_{{P}^{V_{2j+k_0-2}}_{\pi_{2j+k_0-2}}}Q_{T_{2j+k_0-2}}.
$$
This mismatch necessitates an additional error analysis when bounding the first term, in contrast to standard Q-learning. The resulting bound is stated in the following lemma, whose proof is deferred to Appendix~\ref{subsec:proof-lem-var-V-pik-1}.
}
\begin{lem}\label{lem:var-V-pik-1}
Based on definition \eqref{eq:def-Delta}, we have
\begin{align}\label{eq:lem-var-V-pik-1}
&\quad\sum_{j=1}^{\lfloor k/2\rfloor}\gamma^{{\lfloor k/2\rfloor}-j}\prod_{i=j+1}^{\lfloor k/2\rfloor}P_{\pi^{\star}}^{V_{2i+k_0-2}}\sqrt{\frac{\mathsf{Var}_{{P}^{V_{2j+k_0-2}}}V_{2j+k_0-2}}{T_{2j+k_0}}}\nonumber\\
&\lesssim \sqrt{\frac{\mu_{\mathsf{min}}\gamma^k}{\log^2T\log\frac{SAT}{\delta}}}+\sqrt{\log\frac{C_1}{(1-\gamma)\varepsilon}\frac{\sum_{j=2}^{\lfloor k/2\rfloor}\left(\frac{\gamma}{1-\beta}\right)^{k-k_0-2j}\Delta_{2j+k_0-2} + \Delta_k}{T\max\{1-\gamma,\sigma\}(1-\gamma)}}\nonumber\\
&\quad+ \sqrt{\frac{\log\frac{C_1}{(1-\gamma)\varepsilon}}{T\max\{1-\gamma,\sigma\}(1-\gamma)^2}}.
\end{align}
\end{lem}

\revision{
It remains to bound the second term in the right-hand side of \eqref{eq:proof-decom-var-V-pik}, i.e., the resampling error in terms of variance. By Lemmas \ref{lem:bound-alpha} and \ref{lem:upper-bound-V}, we have
}
\begin{align}\label{eq:proof-temp-21-1}
&\quad\left\|\mathsf{Var}_{\overline{P}^{V_{2j+k_0-2}}}V_{2j+k_0-2}-\mathsf{Var}_{{P}^{V_{2j+k_0-2}}}V_{2j+k_0-2}\right\|_{\infty}\nonumber\\
&\lesssim \sqrt{\frac{\theta_p}{N_{2j+k_0-1}\max\{1-\gamma,\sigma\}^3}} + \frac{\varepsilon_\tau}{\max\{1-\gamma,\sigma\}}.
\end{align}
\revision{Under the sample budget condition \eqref{eq:thm-T}, with $C_2$ sufficiently large, the rounding errors in the stage boundaries and sample counts can be absorbed to obtain $N_{2j+k_0-1} \ge \frac13\mu_{\mathsf{min}}\beta T_{2j+k_0-1}\log_2^{-1}\frac{2}{(1-\gamma)\varepsilon_{\tau}}$ from \eqref{eq:def-Np}. Using the definition of $\beta$ in \eqref{eq:def-beta}, we further simplify}
\begin{align}
\sqrt{\frac{\theta_p}{N_{2j+k_0-1}\max\{1-\gamma,\sigma\}^3}}
&\le \sqrt{\frac{3\theta_p\log_2\frac{2}{(1-\gamma)\varepsilon_{\tau}}}{\mu_{\mathsf{min}}\beta T_{2j+k_0-1}\max\{1-\gamma,\sigma\}^3}}\nonumber\\
&=\sqrt{\frac{12\theta_p\log_2\frac{2}{(1-\gamma)\varepsilon_{\tau}}\log T}{\mu_{\mathsf{min}} T_{2j+k_0-1}(1-\gamma)\max\{1-\gamma,\sigma\}^3}}\nonumber\\
&\overset{\text{(i)}}{\lesssim} \sqrt{\frac{\theta_p\log_2\frac{2}{(1-\gamma)\varepsilon_{\tau}}\log T}{\mu_{\mathsf{min}} T(1-\gamma)\max\{1-\gamma,\sigma\}^3}}\overset{\text{(ii)}}{\lesssim} 1,\label{eq:proof-temp-N}
\end{align}
where (i) follows from the lower bound $T_{2j+k_0-1}\ge T_1\gtrsim T$ in \eqref{eq:lem-bound-T1}, and (ii) follows from the lower bound on $T$ in \eqref{eq:thm-T}.
Substituting \eqref{eq:proof-temp-N} into \eqref{eq:proof-temp-21-1} and recalling the assumption $\varepsilon_\tau\lesssim (1-\gamma)\varepsilon$ in \eqref{eq:thm-condition}, we obtain
\begin{align}\label{eq:proof-var-V-pik-2}
\left\|\mathsf{Var}_{\overline{P}^{V_{2j+k_0-2}}}V_{2j+k_0-2}-\mathsf{Var}_{{P}^{V_{2j+k_0-2}}}V_{2j+k_0-2}\right\|_{\infty}
\lesssim 1.
\end{align}

Moreover, recalling that $T_{2j+k_0}\ge T_1 \ge {\rm e}^{-1}T$ in Lemma \ref{lem:frac-k-Tk}, we have
\begin{align}\label{eq:proof-temp-21-2}
\sum_{j=1}^{\lfloor k/2\rfloor}\gamma^{{\lfloor k/2\rfloor}-j}\frac{1}{\sqrt{T_{2j+k_0}}}
\lesssim \frac{1}{\sqrt{T}}\sum_{j=1}^{\lfloor k/2\rfloor}\gamma^{{\lfloor k/2\rfloor}-j}
\lesssim\frac{1}{\sqrt{T(1-\gamma)^2}}.
\end{align}
Combining \eqref{eq:proof-temp-21-1} and \eqref{eq:proof-temp-21-2}, the second term in \eqref{eq:proof-decom-var-V-pik} is bounded by
\begin{align}\label{eq:proof-temp-37}
&\quad \sum_{j=1}^{\lfloor k/2\rfloor}\frac{\gamma^{{\lfloor k/2\rfloor}-j}}{\sqrt{T_{2j+k_0}}}\sqrt{\left\|\mathsf{Var}_{\overline{P}^{V_{2j+k_0-2}}}V_{2j+k_0-2}-\mathsf{Var}_{{P}^{V_{2j+k_0-2}}}V_{2j+k_0-2}\right\|_\infty}\nonumber\\
&\lesssim\sum_{j=1}^{\lfloor k/2\rfloor}\frac{\gamma^{{\lfloor k/2\rfloor}-j}}{\sqrt{T_{2j+k_0}}}\lesssim\sqrt{\frac{1}{(1-\gamma)^2T}}.
\end{align}

Substituting \eqref{eq:lem-var-V-pik-1} and \eqref{eq:proof-temp-37} into \eqref{eq:proof-decom-var-V-pik}, we obtain
\begin{align}\label{eq:lem-var-V-pik}
&\quad\sum_{j=1}^{\lfloor k/2\rfloor}\gamma^{{\lfloor k/2\rfloor}-j}\prod_{i=j+1}^{\lfloor k/2\rfloor}P_{\pi^{\star}}^{V_{2i+k_0-2}}\sqrt{\frac{\mathsf{Var}_{\overline{P}^{V_{2j+k_0-2}}}V_{2j+k_0-2}}{T_{2j+k_0}}}\nonumber\\
&\lesssim \sqrt{\frac{\mu_{\mathsf{min}}\gamma^k}{\log^2T\log\frac{SAT}{\delta}}}+\sqrt{\log\frac{C_1}{(1-\gamma)\varepsilon}\frac{\sum_{j=2}^{\lfloor k/2\rfloor}\left(\frac{\gamma}{1-\beta}\right)^{k-k_0-2j}\Delta_{2j+k_0-2} + \Delta_k}{T\max\{1-\gamma,\sigma\}(1-\gamma)}}\nonumber\\
&\quad+ \sqrt{\frac{\log\frac{C_1}{(1-\gamma)\varepsilon}}{T\max\{1-\gamma,\sigma\}(1-\gamma)^2}}.
\end{align}

\paragraph{Step 3.2: analysis of the cumulative sampling noise(variance) term in \eqref{eq:proof-thm1-decom-lower-lem}.}
\revision{
Similarly, we decompose the first term in the right-hand-side of
\eqref{eq:proof-thm1-decom-lower-lem} into the cumulative sum of variances under the distribution ${P}^{\star}$, plus the resampling error in terms of variance:
}
\begin{align}\label{eq-lem-var-V-pistar-decomp}
&\quad \sum_{j=1}^{\lfloor k/2\rfloor}\gamma^{{\lfloor k/2\rfloor}-j}\prod_{i=j+1}^{\lfloor k/2\rfloor}P_{\pi_{2i+k_0-2}}^{\star}\sqrt{\frac{\mathsf{Var}_{\overline{P}^{V_{2j+k_0-2}}}V_{2j+k_0-2}}{T_{2j+k_0}}} \nonumber\\
&\le \sum_{j=1}^{\lfloor k/2\rfloor}\gamma^{{\lfloor k/2\rfloor}-j}\prod_{i=j+1}^{\lfloor k/2\rfloor}P_{\pi_{2i+k_0-2}}^{\star}\sqrt{\frac{\mathsf{Var}_{{P}^{\star}}V^{\star,\sigma}}{T_{2j+k_0}}} \nonumber\\
&\quad + \sum_{j=1}^{\lfloor k/2\rfloor}\gamma^{{\lfloor k/2\rfloor}-j}\prod_{i=j+1}^{\lfloor k/2\rfloor}P_{\pi_{2i+k_0-2}}^{\star}\frac{1}{\sqrt{T_{2j+k_0}}}\left(\sqrt{\left\|\mathsf{Var}_{\overline{P}^{V_{2j+k_0-2}}}V_{2j+k_0-2}-\mathsf{Var}_{{P}^{\star}}V^{\star,\sigma}\right\|_{\infty}}\right).
\end{align}
\revision{
In the first term, the error propagation matrix $P_{\pi_{2i+k_0-2}}^{\star}$ differs from the variance distribution ${P}^{\star}_{\pi^{\star}}$ only in the policy: both are the worst-case transition kernel for the optimal value function $V^{\star,\sigma}$, but they are associated with different policies, $\pi_{2i+k_0-2}$ and $\pi^{\star}$, respectively.
This structure allows us to bound the first term with a refined bound.
The second term compares the variances of different value functions under different distributions, so Lemma \ref{lem:bound-alpha} does not directly apply. The following two lemmas bound these terms separately.
Their proofs are deferred to Appendix \ref{subsec:proof-lem-var-V-pistar-1} and \ref{subsec:proof-lem-var-V-pistar-sub}, respectively.
}
\begin{lem}\label{lem:var-V-pistar-1}
Based on definition \eqref{eq:def-Delta}, we have
\begin{align*}
&\sum_{j=1}^{\lfloor k/2\rfloor}\gamma^{{\lfloor k/2\rfloor}-j}\prod_{i=j+1}^{\lfloor k/2\rfloor}P_{\pi_{2i+k_0-2}}^{\star}\sqrt{\frac{\mathsf{Var}_{{P}^{\star}}V^{\star,\sigma}}{T_{2j+k_0}}}\nonumber\\
&\lesssim\sqrt{\frac{\mu_{\mathsf{min}}\gamma^k}{\log^2T\log\frac{SAT}{\delta}}}
+\sqrt{\frac{\log\frac{C_1}{(1-\gamma)\varepsilon}\sum_{j=1}^{\lfloor k/2\rfloor}\left(\frac{\gamma}{1-\beta}\right)^{k-k_0-2j}\Delta_{2j+k_0-2}}{T(1-\gamma)\max\{1-\gamma,\sigma\}}}\nonumber\\
&\quad + \sqrt{\frac{\log\frac{C_1}{(1-\gamma)\varepsilon}}{T(1-\gamma)^2\max\{1-\gamma,\sigma\}}}.
\end{align*}
\end{lem}

\begin{lem}\label{lem:var-V-pistar-sub}
The following inequality holds.
\begin{small}
  \begin{align}\label{eq:lem-var-pistar-sub}
&\quad \sum_{j=1}^{\lfloor k/2\rfloor}\gamma^{{\lfloor k/2\rfloor}-j}\prod_{i=j+1}^{\lfloor k/2\rfloor}P_{\pi_{2i+k_0-2}}^{\star}\frac{1}{\sqrt{T_{2j+k_0}}}\sqrt{\left\|\mathsf{Var}_{\overline{P}^{V_{2j+k_0-2}}}V_{2j+k_0-2}-\mathsf{Var}_{{P}^{\star}}V^{\star,\sigma}\right\|_{\infty}} \nonumber\\
&\lesssim\sqrt{\frac{\log\frac{C_1}{(1-\gamma)\varepsilon}}{T(1-\gamma)\max\{1-\gamma,\sigma\}}}\left(\sqrt{\sum_{j=1}^{\lfloor k/2\rfloor}\left(\frac{\gamma}{1-\beta}\right)^{k-k_0-2j}\left\|{V}_{2j+k_0-2} - {V}^{\star,\sigma}\right\|_{\infty}} + \frac{1}{\sqrt{1-\gamma}}\right).
\end{align}  
\end{small}
\end{lem}

Substituting Lemmas \ref{lem:var-V-pistar-1} and \ref{lem:var-V-pistar-sub} into \eqref{eq-lem-var-V-pistar-decomp}, and recalling that $\|V_{2j+k_0-2}-V^{\star,\sigma}\|_{\infty}\le\|Q_{T_{2j+k_0-2}}-Q^{\star,\sigma}\|_{\infty}\le \Delta_{2j+k_0-2}$,
we obtain
\begin{align}\label{eq:lem-var-V-pistar}
&\quad\sum_{j=1}^{\lfloor k/2\rfloor}\gamma^{{\lfloor k/2\rfloor}-j}\prod_{i=j+1}^{\lfloor k/2\rfloor}P_{\pi_{2i+k_0-2}}^{\star}\sqrt{\frac{\mathsf{Var}_{\overline{P}^{V_{2j+k_0-2}}}V_{2j+k_0-2}}{T_{2j+k_0}}} \nonumber\\
&\lesssim\sqrt{\frac{\mu_{\mathsf{min}}\gamma^k}{\log^2T\log\frac{SAT}{\delta}}}
+\sqrt{\frac{\log\frac{C_1}{(1-\gamma)\varepsilon}\sum_{j=1}^{\lfloor k/2\rfloor}\left(\frac{\gamma}{1-\beta}\right)^{k-k_0-2j}\Delta_{2j+k_0-2}}{T(1-\gamma)\max\{1-\gamma,\sigma\}}}\nonumber\\
&\quad + \sqrt{\frac{\log\frac{C_1}{(1-\gamma)\varepsilon}}{T(1-\gamma)^2\max\{1-\gamma,\sigma\}}}.
\end{align}

\paragraph{Step 3.3: analysis of the cumulative sampling noise(higher-order) term.}
\revision{
We now bound the higher-order sampling noise term appearing in both \eqref{eq:proof-thm1-decom-upper-lem} and \eqref{eq:proof-thm1-decom-lower-lem}.
}
Recalling the lower bound on $T_1$ in \eqref{eq:lem-bound-T1}, we have
\begin{align}\label{eq:proof-temp-42}
\frac{\log\frac{SAT}{\delta}\log^2 T}{\mu_{\mathsf{min}}(1-\gamma)^2\max\{1-\gamma,\sigma\}T_k}
&\le \frac{\log\frac{SAT}{\delta}\log^2 T}{\mu_{\mathsf{min}}(1-\gamma)^2\max\{1-\gamma,\sigma\}T_1}\nonumber\\
&\lesssim \frac{\log\frac{SAT}{\delta}\log^2 T}{\mu_{\mathsf{min}}(1-\gamma)^2\max\{1-\gamma,\sigma\}T}\nonumber\\
&\lesssim \sqrt{\frac{\log\frac{SAT}{\delta}\log^2 T}{\mu_{\mathsf{min}}(1-\gamma)^2\max\{1-\gamma,\sigma\}T}},
\end{align}
where the last line uses 
$$
T\gtrsim \frac{\log^2 T\log\frac{SAT}{\delta}}{\mu_{\mathsf{min}}(1-\gamma)^{2}\max\{1-\gamma,\sigma\}}
$$
which follows from the lower bound on \(T\) in \eqref{eq:thm-T}.

\paragraph{Step 4: combining the bounds and establishing the sample complexity.}
Define the logarithmic factor
$$
G \coloneqq \log\frac{C_1}{(1-\gamma)\varepsilon}\log\frac{SAT}{\delta}\log^2T.
$$

Substituting \eqref{eq:lem-var-V-pik} and \eqref{eq:proof-temp-42} into \eqref{eq:proof-thm1-decom-upper-lem}, we obtain
\begin{align}
 Q_{\max\{T_k,(1-\beta)T_1\}} - Q^{\star,\sigma} 
&\gtrsim  
- \sqrt{\frac{G\sum_{j=2}^{\lfloor k/2\rfloor}\left(\frac{\gamma}{1-\beta}\right)^{k-k_0-2j}\Delta_{2j+k_0-2} + G\Delta_k}{T\mu_{\mathsf{min}}\max\{1-\gamma,\sigma\}(1-\gamma)^2}}\nonumber\\
&- \frac{\gamma^{\lfloor k/2\rfloor}}{1-\gamma} - \sqrt{\frac{G\log^2\frac{1}{(1-\gamma)\varepsilon_\tau}}{T{\mu_{\mathsf{min}}(1-\gamma)^3\max\{1-\gamma,\sigma\}}}} - \frac{\varepsilon_{\tau}}{1-\gamma},\label{eq:proof-thm1-upper-lem}
\end{align}
where we use the fact that 
$$
\theta_p =\log\left(\frac{2SAK}{\delta}\log\frac{1}{(1-\gamma)\varepsilon_{\tau}}\right)\le \log\left(\frac{2SAK}{\delta}\right)\log\left(\frac{1}{(1-\gamma)\varepsilon_{\tau}}\right),
$$
and 
$$
\frac{\gamma^{\lfloor k/2\rfloor}}{\sqrt{1-\gamma}}\le \frac{\gamma^{\lfloor k/2\rfloor}}{1-\gamma}.
$$

Similarly, substituting \eqref{eq:lem-var-V-pistar} and \eqref{eq:proof-temp-42} into \eqref{eq:proof-thm1-decom-lower-lem}, we obtain
\begin{align}
 Q_{\max\{T_k,(1-\beta)T_1\}} - Q^{\star,\sigma} 
&\lesssim  \sqrt{\frac{G\sum_{j=1}^{\lfloor k/2\rfloor}\left(\frac{\gamma}{1-\beta}\right)^{k-k_0-2j}\Delta_{2j+k_0-2}}{T\mu_{\mathsf{min}}(1-\gamma)^2\max\{1-\gamma,\sigma\}}}\nonumber\\
&+ \frac{\gamma^{\lfloor k/2\rfloor}}{1-\gamma} + \sqrt{\frac{G\log^2\frac{1}{(1-\gamma)\varepsilon_\tau}}{{T\mu_{\mathsf{min}}(1-\gamma)^3\max\{1-\gamma,\sigma\} }}} + \frac{\varepsilon_{\tau}}{1-\gamma}.\label{eq:proof-thm1-lower-lem}
\end{align}

Combining \eqref{eq:proof-thm1-upper-lem} and \eqref{eq:proof-thm1-lower-lem}, and recalling the definition of $\Delta_k$, we have for $k\ge 1$,
\begin{align}\label{eq:sequence}
\Delta_k 
&\lesssim  \frac{\gamma^{\lfloor k/2\rfloor}}{1-\gamma} + \sqrt{\frac{G\sum_{j=1}^{\lfloor k/2\rfloor}\left(\frac{\gamma}{1-\beta}\right)^{k-k_0-2j}\Delta_{2j+k_0-2} + G\Delta_k}{T\mu_{\mathsf{min}}(1-\gamma)^2\max\{1-\gamma,\sigma\}}} \nonumber\\
&\quad+\sqrt{\frac{G\log^2\frac{1}{(1-\gamma)\varepsilon_{\tau}}}{T\mu_{\mathsf{min}}(1-\gamma)^3\max\{1-\gamma,\sigma\}}} + \frac{\varepsilon_{\tau}}{1-\gamma}.
\end{align}
For $k=0$, we have $\Delta_0\le \frac{1}{1-\gamma}$
 and \eqref{eq:sequence} holds trivially.
 
The following lemma completes the proof, with its proof deferred to Appendix \ref{subsec:proof-lem-sequence}.
\begin{lem}\label{lem:sequence}
Based on \eqref{eq:sequence}, suppose that
$$
T\ge \frac{4C^2G\log^2\frac{1}{(1-\gamma)\varepsilon_{\tau}}}{\mu_{\mathsf{min}}\varepsilon^2(1-\gamma)^3\max\{1-\gamma,\sigma\}},
$$
where $C= (4C_1)^{1/3}$ is a sufficiently large universal constant.
Then, for $1\le k\le K$,
\begin{align*}
\Delta_k \le\frac{C\gamma^{\lfloor k/2\rfloor}}{1-\gamma} + C\sqrt{\frac{G\log^2\frac{1}{(1-\gamma)\varepsilon_{\tau}}}{T\mu_{\mathsf{min}}(1-\gamma)^3\max\{1-\gamma,\sigma\}}} + \frac{C\varepsilon_{\tau}}{1-\gamma}.
\end{align*}
\end{lem}

Applying Lemma \ref{lem:sequence}, we obtain
\begin{align*}
\left\|Q_T - Q^{\star,\sigma}\right\|_{\infty} 
&= \Delta_K \le\frac{C\gamma^{\lfloor K/2\rfloor}}{1-\gamma} + C\sqrt{\frac{G\log^2\frac{1}{(1-\gamma)\varepsilon_{\tau}}}{T\mu_{\mathsf{min}}(1-\gamma)^3\max\{1-\gamma,\sigma\}}} + \frac{C\varepsilon_{\tau}}{1-\gamma}\\
&\le \frac{\sqrt{2}C\varepsilon}{C_1}+\frac{\varepsilon}{2} + \frac{\varepsilon}{4} \le \varepsilon
\end{align*}
where the second line follows from  $C/C_1 = (4C_1)^{1/3}/C_1 \le 1/(4\sqrt{2})$ as long as \(C_1\ge 32\), together with the lower bound on \(T\) in \eqref{eq:thm-T} and the assumption \(\varepsilon_{\tau}\le c(1-\gamma)\varepsilon\) for \(c\le1/(4C)\).

\section{Discussions}
\label{sec:conclusion}

In this paper, we present a model-free algorithm designed for RMDPs,
and provide a sharp analysis 
for its sample complexity under uncertainty sets defined by the Total Variation (TV) distance, assuming access of a single sample trajectory.
Our algorithm achieves an optimal space complexity of $O(SA)$, which is significantly more efficient compared to the $O(S^2A)$ space requirement of existing model-based approaches.
Compared with other memory-efficient algorithms, our algorithm operates in a more realistic asynchronous setting.
Furthermore, we demonstrate that the algorithm reliably produces an 
$\varepsilon$-approximate Q-function estimate using $\widetilde{O}\left(\frac{1}{\mu_{\mathsf{min}}\varepsilon^2(1-\gamma)^3\max\{1-\gamma,\sigma\}}+\frac{t_{\mathsf{mix}}}{\mu_{\mathsf{min}}(1-\gamma)}\right)$ samples.
This bound is nearly minimax optimal with an  additional factor of $\tilde{O}\left(\frac{1}{1-\gamma}\right)$,
and significantly outperforms sample complexity of existing asynchronous Q-learning algorithms for RMDPs.
 Finally, we extend both the algorithm and its analysis to the synchronous setting, achieving a sample complexity of
$\widetilde{O}\left(\frac{SA}{\varepsilon^2(1-\gamma)^3\max\{1-\gamma,\sigma\}}\right)$.
These results highlight the effectiveness and efficiency of our model-free method in addressing RMDPs.

Our work paves the way for numerous future research directions.
\revision{
One direction for future research is to develop model-free algorithms with minimax-optimal sample complexity for robust MDPs. For standard MDPs, \citet{li2024q} establish that vanilla Q-learning incurs an additional factor of \((1-\gamma)^{-1}\) relative to the minimax lower bound. This limitation does not rule out improvements through algorithmic modifications: variance reduction improves the effective-horizon dependence of asynchronous Q-learning to \((1-\gamma)^{-3}\), matching the minimax dependence on this parameter \citep{li2021sample}. Our robust Q-learning upper bound contains the same additional factor. It is therefore of interest to investigate whether variance reduction can close this gap in the robust setting.
}
Another direction is to explore alternative distance measures, such as \(\chi^2\)-divergence. While we have established a model-free algorithm and its sample complexity for RMDPs under the TV distance, \revision{our approach relies heavily on the specific structure of the TV uncertainty set and can not be extended trivially to more general uncertainty sets. Extending our resampling construction and its guarantees to other uncertainty sets is left for future work.}

\newpage

\appendix
\section{preliminary Lemmas}

Before diving into the proof details, we present some preliminary lemmas, which are frequently used throughout the proof.

We begin from the following lemma which comes from Lemma 8 in \citet{li2021sample}.
Its proof is deferred to Appendix~\ref{subsec:proof-lem-markovian}.

\begin{lem}\label{lem:markovian}
Consider the above-mentioned Markov chain. For any $0\le \delta\le 1$, $0\le h< k\le K$, and for all $y\in\mathcal{S}\times\mathcal{A}$, we have
\begin{align*}
\mathsf{P}_{(s_{T_{h}},a_{T_{h}})=y}\left(\exists x\in\mathcal{S}\times \mathcal{A}:\left|\sum_{i=T_{h}}^{T_k-1}\mathds{1}\{s_{i+1}=x\}-(T_k-T_{h})\mu_{\pi_\mathsf{b}}(x)\right|\ge \frac12(T_k-T_{h})\mu_{\pi_\mathsf{b}}(x)\right)\le \delta.
\end{align*}
\end{lem}

Next we introduce a lemma about the properties of $\Lambda_t^{T_k}$ defined in \eqref{eq:def-LamdatT}, the proof of which is deferred to Appendix~\ref{subsec:proof-lem-eta}.
\begin{lem}\label{lem:eta}
 For any integer $k\ge 1$ and $T$ satisfying \eqref{eq:thm-T}, the following properties hold with probability at least $1-\delta$:
 \begin{align}
 \Big\|\Lambda_t^{T_k}\Big\|_{\infty} &\le  \Big\|\prod_{i=t+1}^{T_k}(1-\Lambda_i)\Big\|_{\infty}\lesssim \frac{1}{T^2},\quad 0\le t\le (1-\beta)T_{k},\label{eq:lem-bound-eta-1}\\
 \Big\|\Lambda_t^{T_k} \Big\|_{\infty}&\lesssim \frac{\log^2 T}{(1-\gamma)T_k}, \quad (1-\beta)T_k< t\le T_k,\label{eq:lem-bound-eta-2}\\
 \sum_{t=0}^{T_k}\Lambda_t^{T_k} &= I.\label{eq:sum-eta-1}
 \end{align}
\end{lem}

The following lemma shows some properties about $\beta$ and $\gamma$, the proof of which is deferred to Appendix~\ref{subsec:proof-lem-gamma-beta}.
\begin{lem}\label{lem:gamma-beta}
For $T\ge C$ where $C$ is a sufficiently large universal constant, we have
\begin{align*}
\frac{1}{1-\frac{\gamma}{1-\beta}} \lesssim \frac{1}{1-\gamma},\quad  
\frac{1}{1-\frac{\gamma^2}{1-\beta}}\le \frac{1}{1-\frac{\gamma}{\sqrt{1-\beta}}}\lesssim \frac{1}{1-\gamma}.
\end{align*}
\end{lem}

The following lemma provides an upper bound for the difference between the worst-case expectations based on two value-functions, the proof of which is deferred to Appendix \ref{subsec:proof-lem-infinity-norm}.
\begin{lem}\label{lem:infinity-norm}
For any two vectors $V$ and $V'\in\mathbb{R}^{S}$, define $P^{V}V = \inf_{\mathcal{P}\in\mathcal{U}^{\sigma}(P^0)} \mathcal{P} V$ and $P^{V'}V' = \inf_{\mathcal{P}\in\mathcal{U}^{\sigma}(P^0)} \mathcal{P} V'$.
We have
\begin{align*}
    \left\|P^{V}V - P^{V'}V'\right\|_{\infty} \le \left\|V - V'\right\|_{\infty}.
\end{align*}
\end{lem}

\section{Proof of Lemma \ref{lem:bound-resample-sample}}
\label{subsec:proof-lem-bound-resample-sample}
Denote 
\begin{align}\label{eq:def-Lt}
L_t(s,a) = \sum_{j=1}^t\mathds{1}_{(s_{j-1},a_{j-1})=(s,a)}
\end{align}
as the number that the trajectory visits $(s,a)$ before the time \revision{step} $t$.

We begin with the analysis of the sampling noise.
Since $P_t$ and $\overline{P}^{V_{k-2}}$ both have unit row sum at the visited state-action pair, and $\Lambda_t^{T_k}$ is zero on all other rows, subtracting a constant from $V_{k-2}$ leaves the weighted difference unchanged. Thus,
\begin{align}\label{eq:proof-lem-sample-1}
\Big|\gamma\sum_{t=T_{k-1}+1}^{T_k}\Lambda_t^{T_k} (P_t-\overline{P}^{V_{k-2}})V_{k-2}\Big| 
&=\Big|\gamma\sum_{t=T_{k-1}+1}^{T_k}\Lambda_t^{T_k} (P_t-\overline{P}^{V_{k-2}})\left(V_{k-2}-\min_{s\in\mathcal{S}} V_{k-2}(s)\right)\Big|.
\end{align}
To apply Bernstein's inequality, we first bound the sum of squared weights and the largest weight. For each $(s,a)\in\mathcal{S}\times\mathcal{A}$, we have
\begin{align}
   \sum_{t=T_{k-1}+1}^{T_k}(\Lambda_t^{T_k}(s,a))^2 
   &=\sum_{t=T_{k-1}+1}^{T_k}\prod_{j=t+1}^{T_k}\left(I-\Lambda_j(s,a)\right)^2\Lambda_t^2(s,a) \nonumber\\
   &\overset{\text{(i)}}{=}\sum_{n=L_{T_{k-1}}+1}^{L_{T_k}}\left(1-\eta_k\right)^{2(L_{T_k}-n)}\eta_k^2\le \frac{\eta_k^2}{1-(1-\eta_k)^2} \le \eta_k\nonumber\\
   & \overset{\text{(ii)}}{=} \frac{8\log T}{8\log T + \mu_{\mathsf{min}}\beta T_k}\le \frac{8\log T}{\mu_{\mathsf{min}}\beta T_k}\overset{\text{(iii)}}{\lesssim} \frac{\log^2T}{\mu_{\mathsf{min}}(1-\gamma)T_k},\label{eq:proof-temp-47}
\end{align} 
where (i) uses the definition of $L_t$ in \eqref{eq:def-Lt}, (ii)
arises from the definition of $\eta_k$ in \eqref{eq:thm-condition}, and (iii) follows from the definition of $\beta$ in \eqref{eq:def-beta}.
Moreover, we have
\begin{align}
   \max_{T_{k-1}< t\le T_k}\left\|\Lambda_t^{T_k} \right\|
   &= \eta_{k} = \frac{8\log T}{8\log T + \mu_{\mathsf{min}}\beta T_k}\le \frac{8\log T}{\mu_{\mathsf{min}}\beta T_k}\lesssim \frac{\log^2T}{\mu_{\mathsf{min}}(1-\gamma)T_k}.\label{eq:proof-temp-48}
   \end{align}

The centering in \eqref{eq:proof-lem-sample-1} allows us to control the sample magnitude by the span of $V_{k-2}$. By Lemma \ref{lem:upper-bound-V},
$$
\left\|V_{k-2}-\min_{s\in\mathcal{S}} V_{k-2}(s)\right\|_{\infty}\lesssim \frac{1}{\max\{\sigma,1-\gamma\}}.
$$
Condition on the history at the beginning of stage $k$, so that the value function and the resampling parameters used in this stage are fixed. For each state-action pair $(s,a)$, index the resampled next states by successive visits to $(s,a)$. For any fixed number $n$ of visits, these samples are independent, and the geometric weights in their weighted sum are deterministic. Applying Bernstein's inequality with failure probability $\delta/(SAT)$ gives the following coordinatewise bound for each fixed choice of $k$, $(s,a)$, and $1\le n\le T_k-T_{k-1}$:
\begin{align*}
&\quad\Big|\gamma\sum_{t=T_{k-1}+1}^{T_k}\Lambda_t^{T_k} (P_t-\overline{P}^{V_{k-2}})V_{k-2}\Big| \nonumber\\
&\overset{}{\lesssim} \sqrt{\log\frac{SAT}{\delta}}\sqrt{\frac{\log^2T}{\mu_{\mathsf{min}}(1-\gamma)T_k}}\sqrt{\mathsf{Var}_{\overline{P}^{V_{k-2}}}V_{k-2}} + \frac{\log\frac{SAT}{\delta}\log^2 T}{\mu_{\mathsf{min}}(1-\gamma)T_k}\frac{1}{\max\{1-\gamma,\sigma\}}\nonumber\\
&\lesssim\sqrt{\log\frac{SAT}{\delta}}\sqrt{\frac{\log^2T}{\mu_{\mathsf{min}}(1-\gamma)T_k}}\sqrt{\mathsf{Var}_{\overline{P}^{V_{k-2}}}V_{k-2}} + \frac{\log\frac{SAT}{\delta}\log^2 T}{\mu_{\mathsf{min}}(1-\gamma)\max\{1-\gamma,\sigma\}T_k}.
\end{align*}
We first apply a union bound over all fixed visit counts $1\le n\le T_k-T_{k-1}$, all state-action pairs, and all stages. Since $\sum_{k=1}^K(T_k-T_{k-1})=T$, there are at most $SAT$ such choices. Thus, with probability at least $1-\delta$, the bounds hold simultaneously for all these choices. On this event, we may substitute the actual random visit count $n=L_{T_k}(s,a)-L_{T_{k-1}}(s,a)$ without conditioning on its realized value; if this count is zero, the sampling-noise term is zero. Consequently, we obtain
\begin{align}\label{eq:proof-lem-sample-2}
&\quad\Big|\gamma\sum_{t=T_{k-1}+1}^{T_k}\Lambda_t^{T_k} (P_t-\overline{P}^{V_{k-2}})V_{k-2}\Big| \nonumber\\
&\lesssim\sqrt{\log\frac{SAT}{\delta}}\sqrt{\frac{\log^2T}{\mu_{\mathsf{min}}(1-\gamma)T_k}}\sqrt{\mathsf{Var}_{\overline{P}^{V_{k-2}}}V_{k-2}} + \frac{\log\frac{SAT}{\delta}\log^2 T}{\mu_{\mathsf{min}}(1-\gamma)\max\{1-\gamma,\sigma\}T_k}.
\end{align}

We next control the resampling bias. Since the update weights are nonnegative and sum to at most $I$, the weighted bias is bounded by the single-sample expectation discrepancy:
 \begin{align*}
&\Big\|\gamma\sum_{t=T_{k-1}+1}^{T_k}\Lambda_t^{T_k}\left(\overline{P}^{V_{k-2}} - {P}^{V_{k-2}}\right)V_{k-2}\Big\|_{\infty}
\le \left\|\left(\overline{P}^{V_{k-2}} - {P}^{V_{k-2}}\right)V_{k-2}\right\|_{\infty}.
 \end{align*}
Applying the expectation bound in Lemma \ref{lem:bound-alpha} together with the span bound in Lemma \ref{lem:upper-bound-V} gives
\begin{align*}
\Big\|\gamma\sum_{t=T_{k-1}+1}^{T_k}\Lambda_t^{T_k}\left(\overline{P}^{V_{k-2}} - {P}^{V_{k-2}}\right)V_{k-2}\Big\|_{\infty}
&\lesssim \sqrt{\frac{\theta_p\max\{1-\gamma,\sigma\}}{N_{k-1}}}\frac{1}{\max\{\sigma,1-\gamma\}} + \varepsilon_\tau\nonumber\\
&\lesssim \sqrt{\frac{\theta_p}{N_{k-1}\max\{1-\gamma,\sigma\}}} + \varepsilon_{\tau}.
\end{align*}

\section{Proof of Lemma \ref{lem:decom}}
\label{subsec:proof-lem-decom}
We first derive a one-stage inequality and then iterate it separately over the even and odd stages, since the update in stage $k$ uses $V_{k-2}$. The two directions of the error bound follow from the corresponding Bellman comparisons in \eqref{eq:proof-temp-10} and \eqref{eq:proof-temp-11}.

\paragraph{Proof of \eqref{eq:proof-thm1-decom-upper-lem}.}
Substituting \eqref{eq:proof-temp-10} into \eqref{eq:proof-temp-12} yields
 \begin{align}\label{eq:proof-thm1-decom-upper}
   Q_{T_k} - Q^{\star,\sigma} 
&\ge\prod_{t=T_{k-1}+1}^{T_k}(I-\Lambda_t)\left(Q_{T_{k-1}}-  Q^{\star,\sigma}\right)+ \gamma\sum_{t=T_{k-1}+1}^{T_k}\Lambda_t^{T_k}{P}^{V_{k-2}}_{\pi^{\star}}\left(Q_{T_{k-2}}-Q^{\star,\sigma}\right)\notag\\
&\quad + \gamma\sum_{t=T_{k-1}+1}^{T_k}\Lambda_t^{T_k}\bigg( \left(P_t-\overline{P}^{V_{k-2}}\right)V_{k-2} + \left(\overline{P}^{V_{k-2}} - {P}^{V_{k-2}}\right)V_{k-2}\bigg).
\end{align}
For the first term, according to \eqref{eq:lem-bound-eta-1} in Lemma \ref{lem:eta}, we have
\begin{align*}
\prod_{t=T_{k-1}+1}^{T_k}(I-\Lambda_t) \lesssim \frac{1}{T^2},
\end{align*}
and therefore
\begin{align}\label{eq:proof-lem-decom-1}
\Big\|\prod_{t=T_{k-1}+1}^{T_k}(I-\Lambda_t)\left(Q_{T_{k-1}}-  Q^{\star,\sigma}\right)\Big\|_{\infty} \lesssim \frac{1}{T^2(1-\gamma)}.
\end{align}
Define
\begin{align}\label{eq:def-tildelambad}
\widetilde{\Lambda}_k:=\sum_{t=T_{k-1}+1}^{T_k}\Lambda_t^{T_k} \le \sum_{t=0}^{T_k}\Lambda_t^{T_k} = 1.
\end{align}
Combining \eqref{eq:proof-lem-decom-1} and the bounds for single-stage resample bias and sampling noise in Lemma \ref{lem:bound-resample-sample} into \eqref{eq:proof-thm1-decom-upper} yields
\begin{align}
   Q_{T_k} - Q^{\star,\sigma}
   &\ge \gamma\widetilde{\Lambda}_k{P}^{V_{k-2}}_{\pi^{\star}}\left(Q_{T_{k-2}} - Q^{\star,\sigma}\right)-C\sqrt{\log\frac{SAT}{\delta}}\sqrt{\mathsf{Var}_{\overline{P}^{V_{k-2}}}V_{k-2}\frac{\log^2T}{\mu_{\mathsf{min}}(1-\gamma)T_k}}\nonumber\\
   &\quad - \frac{C\log\frac{SAT}{\delta}\log^2 T}{\mu_{\mathsf{min}}(1-\gamma)\max\{1-\gamma,\sigma\}T_k}- C\sqrt{\frac{\theta_p}{N_{k-1}\max\{1-\gamma,\sigma\}}}-C\varepsilon_{\tau},\label{eq:proof-recursive-ge}
\end{align}
where $C$ is a universal constant, and we have used the fact that
\begin{align*}
\frac{1}{T^2(1-\gamma)}\lesssim \frac{\log\frac{SAT}{\delta}\log^2 T}{\mu_{\mathsf{min}}(1-\gamma)\max\{1-\gamma,\sigma\}T_k}.
\end{align*}
The recursion in \eqref{eq:proof-recursive-ge} connects stages $k$ and $k-2$. We therefore set $k_0=0$ for even $k$ and $k_0=1$ for odd $k$, and expand the recursion along stages:
\begin{align}
Q_{T_k} - Q^{\star,\sigma} 
&\gtrsim  \gamma^{\lfloor k/2\rfloor}\prod_{j=1}^{\lfloor k/2\rfloor}\left(\widetilde{\Lambda}_{2j+k_0}{P}^{V_{2j+k_0-2}}_{\pi^{\star}}\right)\left(Q_{T_{k_0}} - Q^{\star,\sigma}\right)\nonumber\\
&\quad -C\sqrt{\log\frac{SAT}{\delta}}\sqrt{\frac{\log^2T}{\mu_{\mathsf{min}}(1-\gamma)}}\sum_{j=1}^{\lfloor k/2\rfloor}\gamma^{\lfloor k/2\rfloor-j}\prod_{i=j+1}^{\lfloor k/2\rfloor}P_{\pi^{\star}}^{V_{2i+k_0-2}}\sqrt{\frac{\mathsf{Var}_{\overline{P}^{V_{2j+k_0-2}}}V_{2j+k_0-2}}{T_{2j+k_0}}}\nonumber\\
&\quad - C\frac{\log\frac{SAT}{\delta}\log^2 T}{\mu_{\mathsf{min}}(1-\gamma)\max\{1-\gamma,\sigma\}}\sum_{j=1}^{\lfloor k/2\rfloor}\frac{\gamma^{\lfloor k/2\rfloor-j}}{T_{2j+k_0}}\nonumber\\
&\quad- C\sqrt{\frac{\theta_p}{\max\{1-\gamma,\sigma\}}}\sum_{j=1}^{\lfloor k/2\rfloor}\frac{\gamma^{{\lfloor k/2\rfloor}-j}}{\sqrt{N_{2j+k_0-1}}} -C\sum_{j=1}^{\lfloor k/2\rfloor}\gamma^{{\lfloor k/2\rfloor}-j}\varepsilon_{\tau},\label{eq:proof-temp-13}
\end{align}
Here we use $0\le\widetilde{\Lambda}_{2i+k_0}\le I$ and the nonnegativity of the transition matrices. The accumulated error terms are nonnegative before their leading minus signs, so replacing their propagation matrices $\widetilde{\Lambda}_{2i+k_0}P_{\pi^\star}^{V_{2i+k_0-2}}$ by $P_{\pi^\star}^{V_{2i+k_0-2}}$ preserves the lower bound.
According to Lemma \ref{lem:gamma-beta}, we further have 
\begin{align}
\left\|Q_{(1-\beta)T_{k_0+1}} - Q^{\star,\sigma}\right\|_{\infty}
&\le\frac{1}{1-\gamma},\quad \left\|Q_{0} - Q^{\star,\sigma}\right\|_{\infty}
\le\frac{1}{1-\gamma},\quad \left\|Q_{T_1} - Q^{\star,\sigma}\right\|_{\infty}
\le\frac{2}{1-\gamma},\notag\\
\sum_{j=1}^{\lfloor k/2\rfloor}\gamma^{{\lfloor k/2\rfloor}-j}&\lesssim\frac{1}{1-\gamma},\label{eq:proof-temp-14}\\
\sum_{j=1}^{\lfloor k/2\rfloor}\frac{\gamma^{{\lfloor k/2\rfloor}-j}}{T_{2j+k_0}}
&\lesssim\sum_{j=1}^{\lfloor k/2\rfloor}\frac{\gamma^{{\lfloor k/2\rfloor}-j}}{(1-\beta)^{k-2j-k_0}T_k}\lesssim\frac{1}{T_k(1-\frac{\gamma}{(1-\beta)^2})}\lesssim\frac{1}{T_k(1-\gamma)}.\label{eq:proof-temp-15}
\end{align}
It remains to bound the accumulated resampling bias, which involves the sample counts $N_{2j+k_0-1}$: 
\begin{align}\label{eq:proof-decom-2}
\sum_{j=1}^{\lfloor k/2\rfloor}\frac{\gamma^{{\lfloor k/2\rfloor}-j}}{\sqrt{N_{2j+k_0-1}}}
&\lesssim\sqrt{\frac{2\log_2\frac{2}{(1-\gamma)\varepsilon_\tau}}{{\mu_{\mathsf{min}}(1-\beta)\beta T_k}}}\sum_{j=1}^{\lfloor k/2\rfloor}\frac{\gamma^{{\lfloor k/2\rfloor}-j}}{\sqrt{(1-\beta)^{k-2j-k_0}}}\nonumber\\
&\le\sqrt{\frac{\log_2\frac{2}{(1-\gamma)\varepsilon_\tau}}{{\mu_{\mathsf{min}}(1-\beta)\beta T_k}}}\sqrt{k\sum_{j=1}^{\lfloor k/2\rfloor}\frac{\gamma^{k-k_0-2j}}{(1-\beta)^{k-2j-k_0}}}\nonumber\\
&\le\sqrt{\frac{k\log_2\frac{2}{(1-\gamma)\varepsilon_\tau}}{{\mu_{\mathsf{min}}(1-\beta)\beta T_k}}\frac{1}{1-\frac{\gamma}{1-\beta}}}
\lesssim\sqrt{\frac{k\log\frac{2}{(1-\gamma)\varepsilon_\tau}}{{\mu_{\mathsf{min}}(1-\beta)(1-\gamma)\beta T_k}}}\nonumber\\
&\lesssim\sqrt{\frac{K\log T\log\frac{2}{(1-\gamma)\varepsilon_\tau}}{{\mu_{\mathsf{min}}(1-\gamma)^2 T}}}
\overset{\text{(i)}}{\lesssim}\sqrt{\frac{\log T\log\frac{2}{(1-\gamma)\varepsilon_\tau}\log\frac{C_1}{(1-\gamma)\varepsilon}}{{\mu_{\mathsf{min}}(1-\gamma)^3 T}}},
\end{align}
where (i) uses the choice of $K$ in \eqref{eq:thm-condition}.
Inserting \eqref{eq:proof-temp-14}, \eqref{eq:proof-temp-15}, and \eqref{eq:proof-decom-2} into \eqref{eq:proof-temp-13}, we have for $k\ge 1$,
\begin{align*}
&\quad Q_{T_k} - Q^{\star,\sigma}\nonumber\\
&\gtrsim  -\sqrt{\log\frac{SAT}{\delta}}\sqrt{\frac{\log^2T}{\mu_{\mathsf{min}}(1-\gamma)}}\sum_{j=1}^{\lfloor k/2\rfloor}\gamma^{{\lfloor k/2\rfloor}-j}\prod_{i=j+1}^{\lfloor k/2\rfloor}P_{\pi^{\star}}^{V_{2i+k_0-2}}\sqrt{\frac{\mathsf{Var}_{\overline{P}^{V_{2j+k_0-2}}}V_{2j+k_0-2}}{T_{2j+k_0}}}\nonumber\\
&\quad - \frac{\gamma^{\lfloor k/2\rfloor}}{1-\gamma} - \frac{\log\frac{SAT}{\delta}\log^2 T}{\mu_{\mathsf{min}}(1-\gamma)^2\max\{1-\gamma,\sigma\}T_k} - \sqrt{\frac{\theta_p\log T\log\frac{2}{(1-\gamma)\varepsilon_\tau}\log\frac{C_1}{(1-\gamma)\varepsilon}}{{\mu_{\mathsf{min}}(1-\gamma)^3\max\{1-\gamma,\sigma\} T}}}-\frac{\varepsilon_{\tau}}{1-\gamma}.
\end{align*}

\paragraph{Proof of \eqref{eq:proof-thm1-decom-lower-lem}.}
Substituting \eqref{eq:proof-temp-11} into \eqref{eq:proof-temp-12} yields
\begin{align}\label{eq:proof-thm1-decom-lower}
Q_{T_k} - Q^{\star,\sigma} 
&\le \prod_{t=T_{k-1}+1}^{T_k}(I-\Lambda_t)\left(Q_{T_{k-1}}-  Q^{\star,\sigma}\right)+ \gamma\sum_{t=T_{k-1}+1}^{T_k}\Lambda_t^{T_k}{P}^{\star}_{\pi_{k-2}}\left(Q_{T_{k-2}}-Q^{\star,\sigma}\right)\notag\\
&\quad + \gamma\sum_{t=T_{k-1}+1}^{T_k}\Lambda_t^{T_k}\bigg( \left(P_t-\overline{P}^{V_{k-2}}\right)V_{k-2} + \left(\overline{P}^{V_{k-2}} - {P}^{V_{k-2}}\right)V_{k-2}\bigg).
\end{align}
By similar argument as the proof of \eqref{eq:proof-recursive-ge}, we have for $k\ge 2$,
\begin{align*}
Q_{T_k} - Q^{\star,\sigma}
&\le \gamma\widetilde{\Lambda}_k {P}^{\star}_{\pi_{k-2}}\left(Q_{T_{k-2}} - Q^{\star,\sigma}\right) + C\sqrt{\log\frac{SAT}{\delta}}\sqrt{\frac{\log^2T}{\mu_{\mathsf{min}}(1-\gamma)T_k}}\sqrt{\mathsf{Var}_{\overline{P}^{V_{k-2}}}V_{k-2}}\nonumber\\
&+ \frac{C\log\frac{SAT}{\delta}\log^2 T}{\mu_{\mathsf{min}}(1-\gamma)\max\{1-\gamma,\sigma\}T_k}+ C\sqrt{\frac{\theta_p}{N_{k-1}\max\{1-\gamma,\sigma\}}} + C\varepsilon_{\tau},
\end{align*}
where $C$ is a large enough universal constant and $\widetilde{\Lambda}_k$ is defined in \eqref{eq:def-tildelambad}.
We now iterate this upper bound over stages by using \eqref{eq:proof-temp-14} and \eqref{eq:proof-temp-15}. The scalar error terms are bounded as above, while the variance terms propagate through $P^\star_{\pi_{2i+k_0-2}}$: 
\begin{align*}
Q_{T_k} - Q^{\star,\sigma} 
&\lesssim  \sqrt{\log\frac{SAT}{\delta}}\sqrt{\frac{\log^2T}{\mu_{\mathsf{min}}(1-\gamma)}}\sum_{j=1}^{\lfloor k/2\rfloor}\gamma^{{\lfloor k/2\rfloor}-j}\prod_{i=j+1}^{\lfloor k/2\rfloor}P_{\pi_{2i+k_0-2}}^{\star}\sqrt{\frac{\mathsf{Var}_{\overline{P}^{V_{2j+k_0-2}}}V_{2j+k_0-2}}{T_{2j+k_0}}}\nonumber\\
&\quad + \frac{\gamma^{\lfloor k/2\rfloor}}{1-\gamma} + \frac{\log\frac{SAT}{\delta}\log^2 T}{\mu_{\mathsf{min}}(1-\gamma)^2\max\{1-\gamma,\sigma\}T_k}\nonumber\\
&\quad + \sqrt{\frac{\theta_p\log T\log\frac{2}{(1-\gamma)\varepsilon_\tau}\log\frac{C_1}{(1-\gamma)\varepsilon}}{{\mu_{\mathsf{min}}(1-\gamma)^3\max\{1-\gamma,\sigma\} T}}} + \frac{\varepsilon_{\tau}}{1-\gamma}.
\end{align*}
For $k=0$ and $k=1$, considering that for any $t\ge0$, $\|Q_t-Q^{\star,\sigma}\|_{\infty}\le\frac{1}{1-\gamma}$, inequalities \eqref{eq:proof-thm1-decom-lower-lem} and \eqref{eq:proof-thm1-decom-upper-lem} hold obviously.

\section{Proof of Lemma \ref{lem:var-V-pik-1}}
\label{subsec:proof-lem-var-V-pik-1}

We first use Cauchy--Schwarz and Jensen's inequalities to reduce the weighted sum of standard deviations to a weighted sum of variances:
\begin{align}\label{eq:proof-temp-35}
&\quad \sum_{j=1}^{\lfloor k/2\rfloor}\gamma^{{\lfloor k/2\rfloor}-j}\prod_{i=j+1}^{\lfloor k/2\rfloor}P_{\pi^{\star}}^{V_{2i+k_0-2}}\sqrt{\frac{\mathsf{Var}_{{P}^{V_{2j+k_0-2}}}V_{2j+k_0-2}}{T_{2j+k_0}}}\nonumber\\
&\le \sqrt{k}\sqrt{\sum_{j=1}^{\lfloor k/2\rfloor}\frac{\gamma^{k-k_0-2j}}{T_{2j+k_0}}\prod_{i=j+1}^{\lfloor k/2\rfloor}P_{\pi^{\star}}^{V_{2i+k_0-2}}\mathsf{Var}_{{P}^{V_{2j+k_0-2}}}V_{2j+k_0-2}}.
\end{align}

To relate these variances across stages, define the residual of the ideal robust update:
\begin{align}\label{eq:def-xi}
\xi_j \coloneqq Q_{T_j} - r - \gamma P^{V_{j-2}}_{\pi_{j-2}} Q_{T_{j-2}},\quad 2\le j\le K.
\end{align}
According to robust Bellman consistency equation in \eqref{eq:robust-Bellman}, we have
\begin{align*}
\xi_j &= Q_{T_j} - r - \gamma P^{V_{j-2}}_{\pi_{j-2}} Q_{T_{j-2}} - Q^{\star,\sigma} + r + \gamma P^{\star}_{\pi^{\star}} Q^{\star,\sigma}\notag\\
&=Q_{T_j} - Q^{\star,\sigma} - \gamma\left(P^{V_{j-2}}_{\pi_{j-2}} Q_{T_{j-2}}-P^{\star}_{\pi^{\star}} Q^{\star,\sigma}\right).
\end{align*}
Applying Lemma \ref{lem:infinity-norm} to the worst-case expectations and using the nonexpansiveness of statewise maximization in the infinity norm gives
\begin{align}\label{eq:bound-xi}
\left\|\xi_j\right\|_{\infty}\le \left\|Q_{T_j} - Q^{\star,\sigma}\right\|_{\infty} + \gamma\left\|Q_{T_{j-2}} - Q^{\star,\sigma}\right\|_{\infty}\le \Delta_j + \Delta_{j-2},\quad 2\le j\le K.
\end{align}
We also center the value and Q-functions by the minimum Q-value at the corresponding stage. This preserves the variance and allows the centered functions to be bounded by their spans:
\begin{align}
    \widetilde{V}_{2j+k_0-2} &\coloneqq V_{2j+k_0-2} - \min_{s,a} Q_{T_{2j+k_0-2}}(s,a),\label{eq:def-tildeV}\\
    \widetilde{Q}_{2j+k_0-2}&\coloneqq Q_{T_{2j+k_0-2}} - \min_{s,a} Q_{T_{2j+k_0-2}}(s,a),\quad 1\le j\le \lfloor K/2\rfloor+1.\label{eq:def-tildeQ}
\end{align}
Based on these definitions, we express variance term in the left-hand-side of \eqref{eq:lem-var-V-pik-1} as
\begin{align}\label{eq:proof-temp-30}
&\quad \mathsf{Var}_{{P}^{V_{2j+k_0-2}}}V_{2j+k_0-2} 
= \mathsf{Var}_{{P}^{V_{2j+k_0-2}}}\widetilde{V}_{2j+k_0-2}\nonumber\\
&=P^{V_{2j+k_0-2}}_{\pi_{2j+k_0-2}}\left(\widetilde{Q}_{2j+k_0-2}\circ\widetilde{Q}_{2j+k_0-2}\right) - P^{V_{2j+k_0-2}}_{\pi_{2j+k_0-2}}\widetilde{Q}_{2j+k_0-2}\circ P^{V_{2j+k_0-2}}_{\pi_{2j+k_0-2}}\widetilde{Q}_{2j+k_0-2}.
\end{align}
Combining \eqref{eq:def-xi} and \eqref{eq:def-tildeQ}, we have
$$
P^{V_{2j+k_0-2}}_{\pi_{2j+k_0-2}}\widetilde{Q}_{2j+k_0-2} =\frac{1}{\gamma}\left(Q_{T_{2j+k_0}}-r-\xi_{2j+k_0}-\gamma\min Q_{T_{2j+k_0-2}}\right)
$$
and therefore the square of the expectation term satisfies
\begin{align}
\quad &P^{V_{2j+k_0-2}}_{\pi_{2j+k_0-2}}\widetilde{Q}_{2j+k_0-2}\circ P^{V_{2j+k_0-2}}_{\pi_{2j+k_0-2}}\widetilde{Q}_{2j+k_0-2}\nonumber\\
&=\frac{1}{\gamma^2}\left(Q_{T_{2j+k_0}}-r-\xi_{2j+k_0}-\gamma\min Q_{T_{2j+k_0-2}}\right)\circ \left(Q_{T_{2j+k_0}}-r-\xi_{2j+k_0}-\gamma\min Q_{T_{2j+k_0-2}}\right)\nonumber\\
&=\frac{1}{\gamma^2}\left(\widetilde{Q}_{2j+k_0}-r-\xi_{2j+k_0}-\gamma\min Q_{T_{2j+k_0-2}} + \min Q_{T_{2j+k_0}}\right)^2\nonumber\\
&\ge \frac{1}{\gamma^2}\widetilde{Q}_{2j+k_0}\circ \widetilde{Q}_{2j+k_0} -\frac{2}{\gamma^2}\left(r + \xi_{2j+k_0} + \gamma\min Q_{T_{2j+k_0-2}} - \min Q_{T_{2j+k_0}}\right)\circ\widetilde{Q}_{2j+k_0}.\label{eq:proof-temp-46}
\end{align}
Notice that
\begin{align*}
{Q}_{T_{2j+k_0}}
&\ge (I-\Lambda_{T_{2j+k_0}}) {Q}_{T_{2j+k_0}-1}  + \gamma\Lambda_{T_{2j+k_0}}\mathsf{1}\min{Q}_{T_{2j+k_0-2}}\nonumber\\
&\ge \prod_{t=T_{2j+k_0-1}+1}^{T_{2j+k_0}}(I-\Lambda_t){Q}_{T_{2j+k_0-1}} + \gamma\sum_{t=T_{2j+k_0-1}+1}^{T_{2j+k_0}} \prod_{i=t+1}^{T_{2j+k_0}}(I-\Lambda_i)\Lambda_t\mathsf{1}\min{Q}_{T_{2j+k_0-2}}\nonumber\\
&\ge \gamma\sum_{t=T_{2j+k_0-1}+1}^{T_{2j+k_0}} \prod_{i=t+1}^{T_{2j+k_0}}(I-\Lambda_i)\Lambda_t\mathsf{1}\min{Q}_{T_{2j+k_0-2}}.
\end{align*}
Recalling that $\prod_{t=T_{2j+k_0-1}+1}^{T_{2j+k_0}}(I-\Lambda_t)+\sum_{t=T_{2j+k_0-1}+1}^{T_{2j+k_0}} \prod_{i=t+1}^{T_{2j+k_0}}(I-\Lambda_i)\Lambda_t=I$, we have
\begin{align}\label{eq:proof-revision-temp-1}
\gamma\min {Q}_{T_{2j+k_0-2}} - \min {Q}_{T_{2j+k_0}}
&\le\gamma\bigg\|\prod_{t=T_{2j+k_0-1}+1}^{T_{2j+k_0}}(I-\Lambda_t)\bigg\|_{\infty}\min{Q}_{T_{2j+k_0-2}}
\overset{\text{(i)}}{\lesssim} \frac{1}{T^2(1-\gamma)}\lesssim 1,
\end{align}
where (i) uses \eqref{eq:lem-bound-eta-1} in Lemma \ref{lem:eta}.
Inserting \eqref{eq:proof-revision-temp-1} into \eqref{eq:proof-temp-46}, we have
\begin{align*}
\quad &P^{V_{2j+k_0-2}}_{\pi_{2j+k_0-2}}\widetilde{Q}_{2j+k_0-2}\circ P^{V_{2j+k_0-2}}_{\pi_{2j+k_0-2}}\widetilde{Q}_{2j+k_0-2}\nonumber\\
&\ge \frac{1}{\gamma^2}\widetilde{Q}_{2j+k_0}\circ \widetilde{Q}_{2j+k_0} -\frac{2}{\gamma^2}\left(r + \xi_{2j+k_0} + O(1)\right)\circ\widetilde{Q}_{2j+k_0},
\end{align*}
Recalling the bound for $\xi_{2j+k_0}$ in \eqref{eq:bound-xi}, the assumption that $\gamma\ge1/2$, and the bound for $\widetilde{Q}_{2j+k_0}$ in \eqref{eq:lem-upper-bound-V-k}, we further have
\begin{align}\label{eq:proof-temp-31}
&\quad P^{V_{2j+k_0-2}}_{\pi_{2j+k_0-2}}\widetilde{Q}_{2j+k_0-2}\circ P^{V_{2j+k_0-2}}_{\pi_{2j+k_0-2}}\widetilde{Q}_{2j+k_0-2}\nonumber\\
&\ge\frac{1}{\gamma^2}\widetilde{Q}_{2j+k_0}\circ \widetilde{Q}_{2j+k_0} -\frac{C(1+\Delta_{2j+k_0} + \Delta_{2j+k_0-2})}{\max\{1-\gamma,\sigma\}},
\end{align}
where $C$ is a large enough universal constant. 

We next bound the second moment in \eqref{eq:proof-temp-30}. Its transition matrix uses the greedy policy $\pi_{2j+k_0-2}$, whereas the accumulated variance propagates under $\pi^\star$. To align these matrices, we write
\begin{align}\label{eq:proof-temp-33}
P^{V_{2j+k_0-2}}_{\pi_{2j+k_0-2}}\left(\widetilde{Q}_{2j+k_0-2}\circ\widetilde{Q}_{2j+k_0-2}\right)&=P^{V_{2j+k_0-2}}_{\pi^{\star}}\left(\widetilde{Q}_{2j+k_0-2}\circ\widetilde{Q}_{2j+k_0-2}\right)\nonumber\\
&
+\left(P^{V_{2j+k_0-2}}_{\pi_{2j+k_0-2}}-P^{V_{2j+k_0-2}}_{\pi^{\star}}\right)\left(\widetilde{Q}_{2j+k_0-2}\circ\widetilde{Q}_{2j+k_0-2}\right).
\end{align}
Notice that the difference term is bounded by
\begin{align}
&\quad \left\|\left(P^{V_{2j+k_0-2}}_{\pi_{2j+k_0-2}}-P^{V_{2j+k_0-2}}_{\pi^{\star}}\right)\left(\widetilde{Q}_{2j+k_0-2}\circ\widetilde{Q}_{2j+k_0-2}\right)\right\|_{\infty}\nonumber\\
&=\left\|P^{V_{2j+k_0-2}}\left(\Pi_{\pi_{2j+k_0-2}}-\Pi_{\pi^{\star}}\right)\left(\widetilde{Q}_{2j+k_0-2}\circ\widetilde{Q}_{2j+k_0-2}\right)\right\|_{\infty}\nonumber\\
&=\left\|P^{V_{2j+k_0-2}}\left(\left(\Pi_{\pi_{2j+k_0-2}}\widetilde{Q}_{2j+k_0-2}-\Pi_{\pi^{\star}}\widetilde{Q}_{2j+k_0-2}\right)\circ\left(\Pi_{\pi_{2j+k_0-2}}\widetilde{Q}_{2j+k_0-2}+\Pi_{\pi^{\star}}\widetilde{Q}_{2j+k_0-2}\right)\right)\right\|_{\infty}\nonumber\\
&\le\left\|\Pi_{\pi_{2j+k_0-2}}\widetilde{Q}_{2j+k_0-2}+\Pi_{\pi^{\star}}\widetilde{Q}_{2j+k_0-2}\right\|_\infty\left\|\Pi_{\pi_{2j+k_0-2}}\widetilde{Q}_{2j+k_0-2}-\Pi_{\pi^{\star}}\widetilde{Q}_{2j+k_0-2}\right\|_{\infty}\nonumber\\
&\text{\text{(i)}}{\lesssim}\frac{1}{\max\{1-\gamma,\sigma\}}\left\|\Pi_{\pi_{2j+k_0-2}}{Q}_{T_{2j+k_0-2}}-\Pi_{\pi^{\star}}{Q}_{T_{2j+k_0-2}}\right\|_{\infty}\nonumber\\
&\lesssim\frac{1}{\max\{1-\gamma,\sigma\}}\left(\left\|\Pi_{\pi_{2j+k_0-2}}{Q}_{T_{2j+k_0-2}}-\Pi_{\pi^{\star}}Q^{\star,\sigma}\right\|_{\infty} + \left\|\Pi_{\pi^{\star}}Q^{\star,\sigma}-\Pi_{\pi^{\star}}{Q}_{T_{2j+k_0-2}}\right\|_{\infty}\right)\nonumber\\
&\text{\text{(ii)}}{\lesssim}\frac{\Delta_{2j+k_0-2}}{\max\{1-\gamma,\sigma\}},\label{eq:proof-temp-43}
\end{align}
where (i) uses the bound for $\widetilde{Q}_{2j+k_0-2}$ in \eqref{eq:lem-upper-bound-V-k}. For (ii), the greedy-policy definitions imply that $\Pi_{\pi_{2j+k_0-2}}Q_{T_{2j+k_0-2}}$ and $\Pi_{\pi^{\star}}Q^{\star,\sigma}$ are the corresponding statewise maxima. Their difference is therefore bounded by $\Delta_{2j+k_0-2}$, by the nonexpansiveness of maximization. 
Inserting \eqref{eq:proof-temp-43} into \eqref{eq:proof-temp-33}, we have
\begin{align}\label{eq:proof-temp-32}
P^{V_{2j+k_0-2}}_{\pi_{2j+k_0-2}}\left(\widetilde{Q}_{2j+k_0-2}\circ\widetilde{Q}_{2j+k_0-2}\right)
\le P^{V_{2j+k_0-2}}_{\pi^{\star}}\left(\widetilde{Q}_{2j+k_0-2}\circ\widetilde{Q}_{2j+k_0-2}\right)
+\frac{C\Delta_{2j+k_0-2}}{\max\{1-\gamma,\sigma\}}.
\end{align}

Inserting \eqref{eq:proof-temp-31} and \eqref{eq:proof-temp-32} into \eqref{eq:proof-temp-30}, we have
\begin{align*}
&\quad \mathsf{Var}_{{P}^{V_{2j+k_0-2}}}V_{2j+k_0-2} \nonumber\\
&\le P^{V_{2j+k_0-2}}_{\pi^{\star}}\left(\widetilde{Q}_{2j+k_0-2}\circ\widetilde{Q}_{2j+k_0-2}\right) - \frac{1}{\gamma^2}\widetilde{Q}_{2j+k_0}\circ \widetilde{Q}_{2j+k_0} +\frac{2C(1+\Delta_{2j+k_0} + \Delta_{2j+k_0-2})}{\max\{1-\gamma,\sigma\}}.
\end{align*}

We now multiply this bound by the propagation matrices and sum over $j$. The second-moment terms cancel between consecutive stages, apart from the boundary terms and terms caused by the changing denominators $T_{2j+k_0}$. Since these denominators increase with $j$, the remaining interior terms are nonpositive and can be discarded:
\begin{align}
\quad &\sum_{j=1}^{\lfloor k/2\rfloor}\frac{\gamma^{k-k_0-2j}}{T_{2j+k_0}}\prod_{i=j+1}^{\lfloor k/2\rfloor}P_{\pi^{\star}}^{V_{2i+k_0-2}}\mathsf{Var}_{{P}^{V_{2j+k_0-2}}}V_{2j+k_0-2}\nonumber\\
&\le \sum_{j=1}^{\lfloor k/2\rfloor}\frac{\gamma^{k-k_0-2j}}{T_{2j+k_0}}\prod_{i=j+1}^{\lfloor k/2\rfloor}P_{\pi^{\star}}^{V_{2i+k_0-2}}P^{V_{2j+k_0-2}}_{\pi^{\star}}\left(\widetilde{Q}_{2j+k_0-2}\circ\widetilde{Q}_{2j+k_0-2}\right) \nonumber\\
&\quad - \sum_{j=1}^{\lfloor k/2\rfloor}\frac{\gamma^{k-k_0-2j-2}}{T_{2j+k_0}}\prod_{i=j+1}^{\lfloor k/2\rfloor}P_{\pi^{\star}}^{V_{2i+k_0-2}}\left(\widetilde{Q}_{2j+k_0}\circ \widetilde{Q}_{2j+k_0}\right)
+\sum_{j=1}^{\lfloor k/2\rfloor}\frac{\gamma^{k-k_0-2j}}{T_{2j+k_0}}\frac{2C(1+\Delta_{2j+k_0} + \Delta_{2j+k_0-2})}{\max\{1-\gamma,\sigma\}}\nonumber\\
&\le\frac{\gamma^{k-k_0-2}}{T_{k_0+2}}\prod_{i=1}^{\lfloor k/2\rfloor}P_{\pi^{\star}}^{V_{2i+k_0-2}}\left(\widetilde{Q}_{k_0}\circ\widetilde{Q}_{k_0}\right)
+2C\sum_{j=1}^{\lfloor k/2\rfloor}\left(\frac{\gamma}{1-\beta}\right)^{k-k_0-2j}\frac{1+\Delta_{2j+k_0} + \Delta_{2j+k_0-2}}{T_k\max\{1-\gamma,\sigma\}}\nonumber\\
&\overset{\text{(i)}}{\lesssim} \frac{\gamma^k}{\max\{1-\gamma,\sigma\}^2T_1} + \sum_{j=1}^{\lfloor k/2\rfloor}\left(\frac{\gamma}{1-\beta}\right)^{k-k_0-2j}\frac{1+\Delta_{2j+k_0} + \Delta_{2j+k_0-2}}{T_k\max\{1-\gamma,\sigma\}}\nonumber\\
&\lesssim \frac{\gamma^k}{\max\{1-\gamma,\sigma\}^2T_1} + \frac{\sum_{j=1}^{\lfloor k/2\rfloor}\left(\frac{\gamma}{1-\beta}\right)^{k-k_0-2j}}{T_k\max\{1-\gamma,\sigma\}}  + \frac{\Delta_{k_0}\left(\frac{\gamma}{1-\beta}\right)^{k-k_0-2}}{T_k\max\{1-\gamma,\sigma\}}\nonumber\\
&\quad+\sum_{j=2}^{\lfloor k/2\rfloor}\left(\frac{\gamma}{1-\beta}\right)^{k-k_0-2j}\frac{\Delta_{2j+k_0-2}}{T_k\max\{1-\gamma,\sigma\}}+\frac{\Delta_{k}}{T_k\max\{1-\gamma,\sigma\}}\nonumber\\
&\overset{\text{(ii)}}{\lesssim}\frac{\mu_{\mathsf{min}}(1-\gamma)\gamma^k}{\log\frac{C_1}{(1-\gamma)\varepsilon}\log^2 T\log\frac{SAT}{\delta}}+\frac{1}{T_k\max\{1-\gamma,\sigma\}(1-\gamma)}+\frac{\left(\frac{\gamma}{1-\beta}\right)^{k-k_0-2}}{T_k\max\{1-\gamma,\sigma\}(1-\gamma)}\nonumber\\
&\quad +\frac{\sum_{j=2}^{\lfloor k/2\rfloor}\left(\frac{\gamma}{1-\beta}\right)^{k-k_0-2j}\Delta_{2j+k_0-2} + \Delta_k}{T_k\max\{1-\gamma,\sigma\}}\nonumber\\
&\lesssim
\frac{\mu_{\mathsf{min}}(1-\gamma)\gamma^k}{\log\frac{C_1}{(1-\gamma)\varepsilon}\log^2 T\log\frac{SAT}{\delta}} + \frac{\sum_{j=2}^{\lfloor k/2\rfloor}\left(\frac{\gamma}{1-\beta}\right)^{k-k_0-2j}\Delta_{2j+k_0-2} + \Delta_{k}}{T_k\max\{1-\gamma,\sigma\}}\nonumber\\
&\quad + \frac{1}{T_k\max\{1-\gamma,\sigma\}(1-\gamma)},
\label{eq:proof-temp-34}
\end{align}
where (i) comes from $\widetilde{Q}_{k_0} \le 1/\max\{1-\gamma,\sigma\}$, and (ii) uses the fact that $\Delta_{k_0}\le 1/(1-\gamma)$, Lemma \ref{lem:gamma-beta}, and the lower bound of $T_1\ge {\rm e}^{-1}T$ in \eqref{eq:thm-T}.

Substituting \eqref{eq:proof-temp-34} into \eqref{eq:proof-temp-35} yields
\begin{align}\label{eq:proof-temp-36}
&\quad \sum_{j=1}^{\lfloor k/2\rfloor}\gamma^{{\lfloor k/2\rfloor}-j}\prod_{i=j+1}^{\lfloor k/2\rfloor}P_{\pi^{\star}}^{V_{2i+k_0-2}}\sqrt{\frac{\mathsf{Var}_{{P}^{V_{2j+k_0-2}}}V_{2j+k_0-2}}{T_{2j+k_0}}}\nonumber\\
&\lesssim \sqrt{\frac{k\mu_{\mathsf{min}}(1-\gamma)\gamma^k}{\log\frac{C_1}{(1-\gamma)\varepsilon}\log^2 T\log\frac{SAT}{\delta}}}+\sqrt{\frac{k\sum_{j=2}^{\lfloor k/2\rfloor}\left(\frac{\gamma}{1-\beta}\right)^{k-k_0-2j}\Delta_{2j+k_0-2} + k\Delta_k}{T_k\max\{1-\gamma,\sigma\}}}\nonumber\\
&\quad  + \sqrt{\frac{k}{T_k\max\{1-\gamma,\sigma\}(1-\gamma)}}\nonumber\\
&\lesssim \sqrt{\frac{\mu_{\mathsf{min}}\gamma^k}{\log^2 T\log\frac{SAT}{\delta}}}+\sqrt{\log\frac{C_1}{(1-\gamma)\varepsilon}\frac{\sum_{j=2}^{\lfloor k/2\rfloor}\left(\frac{\gamma}{1-\beta}\right)^{k-k_0-2j}\Delta_{2j+k_0-2}+\Delta_k}{T\max\{1-\gamma,\sigma\}(1-\gamma)}}\nonumber\\
&\quad + \sqrt{\frac{\log\frac{C_1}{(1-\gamma)\varepsilon}}{T\max\{1-\gamma,\sigma\}(1-\gamma)^2}},
\end{align}
where the last line uses \eqref{eq:lem-frac-k-Tk} in Lemma \ref{lem:frac-k-Tk}.


\section{Proof of Lemma \ref{lem:var-V-pistar-1}}
\label{subsec:proof-lem-var-V-pistar-1}
As in the proof of Lemma \ref{lem:var-V-pik-1}, we first reduce the sum of standard deviations to a weighted variance sum using Cauchy--Schwarz and Jensen's inequalities:

\begin{align}\label{eq:proof-temp-24}
\sum_{j=1}^{\lfloor k/2\rfloor}\gamma^{{\lfloor k/2\rfloor}-j}\prod_{i=j+1}^{\lfloor k/2\rfloor}P_{\pi_{2i+k_0-2}}^{\star}\sqrt{\frac{\mathsf{Var}_{{P}^{\star}}V^{\star,\sigma}}{T_{2j+k_0}}} \le \sqrt{\frac{k}{2}}\sqrt{\sum_{j=1}^{\lfloor k/2\rfloor}\gamma^{k-k_0-2j}\prod_{i=j+1}^{\lfloor k/2\rfloor}P_{\pi_{2i+k_0-2}}^{\star}\frac{\mathsf{Var}_{{P}^{\star}}V^{\star,\sigma}}{T_{2j+k_0}}}.
\end{align}
Define $\widetilde{V}^{\star,\sigma}$ (resp. $\widetilde{Q}^{\star,\sigma}$) as the optimal value function (resp. optimal Q-function) with the minimal value of $Q^{\star,\sigma}(s,a)$ removed, i.e.,
\begin{align*}
\widetilde{V}^{\star,\sigma} &\coloneqq V^{\star,\sigma} - \min_{s,a} Q^{\star,\sigma}(s,a),\notag\\
\widetilde{Q}^{\star,\sigma} &\coloneqq Q^{\star,\sigma} - \min_{s,a} Q^{\star,\sigma}(s,a)
\end{align*}
We make the following decomposition.
\begin{align}
\mathsf{Var}_{{P}^{\star}}V^{\star,\sigma} = \mathsf{Var}_{{P}^{\star}}\widetilde{V}^{\star,\sigma}
= {P}^{\star}\left(\widetilde{V}^{\star,\sigma}\circ \widetilde{V}^{\star,\sigma}\right) - {P}^{\star}\widetilde{V}^{\star,\sigma}\circ {P}^{\star}\widetilde{V}^{\star,\sigma}\nonumber\\
={P}^{\star}_{\pi^{\star}}\left(\widetilde{Q}^{\star,\sigma}\circ \widetilde{Q}^{\star,\sigma}\right) - {P}^{\star}_{\pi^{\star}}\widetilde{Q}^{\star,\sigma}\circ {P}^{\star}_{\pi^{\star}}\widetilde{Q}^{\star,\sigma}.\label{eq:proof-temp-23}
\end{align}
According to the robust Bellman consistency equation \eqref{eq:robust-Bellman}, we have
\begin{align*}
{P}^{\star}_{\pi^{\star}}\widetilde{Q}^{\star,\sigma}\circ {P}^{\star}_{\pi^{\star}}\widetilde{Q}^{\star,\sigma} &= \frac{1}{\gamma^2}\left(Q^{\star,\sigma} - r - \gamma\min Q^{\star,\sigma}\right)\circ\left(Q^{\star,\sigma} - r - \gamma\min Q^{\star,\sigma}\right)\nonumber\\
&\ge \frac{1}{\gamma^2}\widetilde{Q}^{\star,\sigma} \circ \widetilde{Q}^{\star,\sigma} -\frac{2}{\gamma^2}(r-(1-\gamma)\min Q^{\star,\sigma})\circ \widetilde{Q}^{\star,\sigma}\nonumber\\
&\ge \frac{1}{\gamma^2}\widetilde{Q}^{\star,\sigma} \circ \widetilde{Q}^{\star,\sigma} -\frac{2}{\gamma^2}r\circ \widetilde{Q}^{\star,\sigma}.
\end{align*}
Substituting this bound into \eqref{eq:proof-temp-23}, we then replace $\pi^\star$ in the second-moment term by $\pi_{2j+k_0-2}$ to match the propagation matrices: 
\begin{align*}
\mathsf{Var}_{{P}^{\star}}V^{\star,\sigma}
&\le {P}^{\star}_{\pi^{\star}}\left(\widetilde{Q}^{\star,\sigma}\circ \widetilde{Q}^{\star,\sigma}\right) - \frac{1}{\gamma^2}\widetilde{Q}^{\star,\sigma} \circ \widetilde{Q}^{\star,\sigma} + \frac{2}{\gamma^2}r\circ \widetilde{Q}^{\star,\sigma}\nonumber\\
&={P}^{\star}_{\pi_{2j+k_0-2}}\left(\widetilde{Q}^{\star,\sigma}\circ \widetilde{Q}^{\star,\sigma}\right) - \frac{1}{\gamma^2}\widetilde{Q}^{\star,\sigma} \circ \widetilde{Q}^{\star,\sigma} + \frac{2}{\gamma^2}r\circ \widetilde{Q}^{\star,\sigma}\nonumber\\
&\quad +\left({P}^{\star}_{\pi^{\star}} - {P}^{\star}_{\pi_{2j+k_0-2}}\right)\left(\widetilde{Q}^{\star,\sigma}\circ \widetilde{Q}^{\star,\sigma}\right).
\end{align*}
Notice that $\gamma \ge 1/2$, we have
\begin{align}\label{eq:proof-revision-temp-3}
\frac{1}{\gamma^2}r\circ \widetilde{Q}^{\star,\sigma}
&\le 4\widetilde{Q}^{\star,\sigma}\lesssim \frac{1}{\max\{1-\gamma,\sigma\}}.
\end{align}
Moreover, we have
\begin{align*}
&\quad \left\|\left({P}^{\star}_{\pi^{\star}} - {P}^{\star}_{\pi_{2j+k_0-2}}\right)\left(\widetilde{Q}^{\star,\sigma}\circ \widetilde{Q}^{\star,\sigma}\right)\right\|_{\infty}\nonumber\\
&=\left\|{P}^{\star}\left(\Pi_{\pi^{\star}} - \Pi_{\pi_{2j+k_0-2}}\right)\left(\widetilde{Q}^{\star,\sigma}\circ \widetilde{Q}^{\star,\sigma}\right)\right\|_{\infty}\nonumber\\
&\le \left\|\Pi_{\pi^{\star}}\widetilde{Q}^{\star,\sigma} + \Pi_{\pi_{2j+k_0-2}}\widetilde{Q}^{\star,\sigma}\right\|_{\infty}\left\|\Pi_{\pi^{\star}}\widetilde{Q}^{\star,\sigma} - \Pi_{\pi_{2j+k_0-2}}\widetilde{Q}^{\star,\sigma}\right\|_{\infty}\nonumber\\
&\lesssim \frac{1}{\max\{1-\gamma,\sigma\}}\left\|\Pi_{\pi^{\star}}{Q}^{\star,\sigma} - \Pi_{\pi_{2j+k_0-2}}{Q}^{\star,\sigma}\right\|_{\infty}\nonumber\\
&\lesssim \frac{1}{\max\{1-\gamma,\sigma\}}\bigg(\left\|\Pi_{\pi^{\star}}{Q}^{\star,\sigma} - \Pi_{\pi_{2j+k_0-2}}{Q}_{\max\{T_{2j+k_0-2},(1-\beta)T_1\}}\right\|_{\infty}\nonumber\\
&\quad +\left\|\Pi_{\pi_{2j+k_0-2}}{Q}_{\max\{T_{2j+k_0-2},(1-\beta)T_1\}} - \Pi_{\pi_{2j+k_0-2}}{Q}^{\star,\sigma}\right\|_{\infty}\bigg)\nonumber\\
&\overset{\text{(i)}}{\lesssim} \frac{\Delta_{2j+k_0-2}}{\max\{1-\gamma,\sigma\}},
\end{align*}
where (i) uses the greedy-policy definitions in \eqref{eq:def-pik}. In particular, $\Pi_{\pi_{2j+k_0-2}}Q_{\max\{T_{2j+k_0-2},(1-\beta)T_1\}}$ and $\Pi_{\pi^{\star}}Q^{\star,\sigma}$ are the corresponding statewise maxima. The nonexpansiveness of maximization and the definition of $\Delta_{2j+k_0-2}$ thus give the first bound below; the second follows because $\Pi_{\pi_{2j+k_0-2}}$ selects one action per state:
\begin{align*}
&\left\|\Pi_{\pi^{\star}}{Q}^{\star,\sigma} - \Pi_{\pi_{2j+k_0-2}}{Q}_{\max\{T_{2j+k_0-2},(1-\beta)T_1\}}\right\|_{\infty}\le\Delta_{2j+k_0-2},\nonumber\\ &\left\|\Pi_{\pi_{2j+k_0-2}}{Q}_{\max\{T_{2j+k_0-2},(1-\beta)T_1\}} - \Pi_{\pi_{2j+k_0-2}}{Q}^{\star,\sigma}\right\|_{\infty}\le\Delta_{2j+k_0-2}.
\end{align*}
Thus we have
\begin{align*}
\mathsf{Var}_{{P}^{\star}}V^{\star,\sigma}
&\le {P}^{\star}_{\pi_{2j+k_0-2}}\left(\widetilde{Q}^{\star,\sigma}\circ \widetilde{Q}^{\star,\sigma}\right) - \frac{1}{\gamma^2}\widetilde{Q}^{\star,\sigma} \circ \widetilde{Q}^{\star,\sigma} + \frac{C}{\max\{1-\gamma,\sigma\}}+\frac{C\Delta_{2j+k_0-2}}{\max\{1-\gamma,\sigma\}}\nonumber\\
&\le{P}^{\star}_{\pi_{2j+k_0-2}}\left(\widetilde{Q}^{\star,\sigma}\circ \widetilde{Q}^{\star,\sigma}\right) - \frac{1}{\gamma^2}\widetilde{Q}^{\star,\sigma} \circ \widetilde{Q}^{\star,\sigma} + \frac{C\Delta_{2j+k_0-2}+C}{\max\{1-\gamma,\sigma\}}.
\end{align*}
We further have
\begin{align}
&\quad \sum_{j=1}^{\lfloor k/2\rfloor}\gamma^{k-k_0-2j}\prod_{i=j+1}^{\lfloor k/2\rfloor}P_{\pi_{2i+k_0-2}}^{\star}\frac{\mathsf{Var}_{{P}^{\star}}V^{\star,\sigma}}{T_{2j+k_0}} \nonumber\\
&\le \sum_{j=1}^{\lfloor k/2\rfloor}\frac{\gamma^{k-k_0-2j}}{T_{2j+k_0}}\Big(\prod_{i=j+1}^{\lfloor k/2\rfloor}P_{\pi_{2i+k_0-2}}^{\star}\Big) \left({P}^{\star}_{\pi_{2j+k_0-2}}\left(\widetilde{Q}^{\star,\sigma}\circ \widetilde{Q}^{\star,\sigma}\right) - \frac{1}{\gamma^2}\widetilde{Q}^{\star,\sigma} \circ \widetilde{Q}^{\star,\sigma}\right)\nonumber\\
&\quad +  \sum_{j=1}^{\lfloor k/2\rfloor}\frac{\gamma^{k-k_0-2j}}{T_{2j+k_0}}\prod_{i=j+1}^{\lfloor k/2\rfloor}P_{\pi_{2i+k_0-2}}^{\star}\frac{C\Delta_{2j+k_0-2}+C}{\max\{1-\gamma,\sigma\}}.\label{eq:proof-temp-27}
\end{align}
We bound the two terms in \eqref{eq:proof-temp-27} separately. For the first, the second-moment terms cancel after reindexing. The interior terms are nonpositive because $T_{2j+k_0}$ increases with $j$, leaving only the initial boundary term:
\begin{align}
&\quad\sum_{j=1}^{\lfloor k/2\rfloor}\frac{\gamma^{k-k_0-2j}}{T_{2j+k_0}}\prod_{i=j+1}^{\lfloor k/2\rfloor}P_{\pi_{2i+k_0-2}}^{\star} \left({P}^{\star}_{\pi_{2j+k_0-2}}\left(\widetilde{Q}^{\star,\sigma}\circ \widetilde{Q}^{\star,\sigma}\right) - \frac{1}{\gamma^2}\widetilde{Q}^{\star,\sigma} \circ \widetilde{Q}^{\star,\sigma}\right)\nonumber\\
&=\sum_{j=1}^{\lfloor k/2\rfloor}\frac{\gamma^{k-k_0-2j}}{T_{2j+k_0}}\Big(\prod_{i=j}^{\lfloor k/2\rfloor}P_{\pi_{2i+k_0-2}}^{\star}\Big) \left(\widetilde{Q}^{\star,\sigma}\circ \widetilde{Q}^{\star,\sigma}\right) - \sum_{j=1}^{\lfloor k/2\rfloor}\frac{\gamma^{k-k_0-2j-2}}{T_{2j+k_0}}\Big(\prod_{i=j+1}^{\lfloor k/2\rfloor}P_{\pi_{2i+k_0-2}}^{\star}\Big)\widetilde{Q}^{\star,\sigma} \circ \widetilde{Q}^{\star,\sigma}\nonumber\\
&\le\frac{\gamma^{k-k_0-2}}{T_{2+k_0}}\Big(\prod_{i=1}^{\lfloor k/2\rfloor}P_{\pi_{2i+k_0-2}}^{\star}\Big)\left(\widetilde{Q}^{\star,\sigma}\circ \widetilde{Q}^{\star,\sigma}\right) \nonumber\\
&\quad + \sum_{j=2}^{\lfloor k/2\rfloor}\gamma^{k-k_0-2j}\left(\frac{1}{T_{2j+k_0}}-\frac{1}{T_{2j+k_0-2}}\right)\Big(\prod_{i=j}^{\lfloor k/2\rfloor}P_{\pi_{2i+k_0-2}}^{\star}\Big)\widetilde{Q}^{\star,\sigma} \circ \widetilde{Q}^{\star,\sigma}\nonumber\\
&\le \frac{\gamma^{k-k_0-2}}{T_{2+k_0}}\Big(\prod_{i=1}^{\lfloor k/2\rfloor}P_{\pi_{2i+k_0-2}}^{\star}\Big)\left(\widetilde{Q}^{\star,\sigma}\circ \widetilde{Q}^{\star,\sigma}\right)\lesssim \frac{\gamma^{k}}{T\max\{1-\gamma,\sigma\}^2}\nonumber\\
&\lesssim \frac{\mu_{\mathsf{min}}(1-\gamma)\gamma^{k}}{\log\frac{C_1}{\varepsilon(1-\gamma)}\log^2T\log\frac{SAT}{\delta}},\label{eq:proof-temp-25}
\end{align}
where the last line comes from the assumption that $T\gtrsim (1-\gamma)^{-1}\max\{1-\gamma,\sigma\}^{-2}\mu_{\mathsf{min}}^{-1}\log\frac{C_1}{\varepsilon(1-\gamma)}\log^2T\log\frac{SAT}{\delta}$.
For the second term, the propagation matrices act on scalar multiples of $\mathsf{1}$ and therefore leave them unchanged. Using the stage schedule and summing the resulting geometric weights gives
\begin{align}
&\quad\sum_{j=1}^{\lfloor k/2\rfloor}\frac{\gamma^{k-k_0-2j}}{T_{2j+k_0}}\prod_{i=j+1}^{\lfloor k/2\rfloor}P_{\pi_{2i+k_0-2}}^{\star}\frac{\Delta_{2j+k_0-2}+1}{\max\{1-\gamma,\sigma\}} \nonumber\\
&\lesssim 
\sum_{j=1}^{\lfloor k/2\rfloor}\left(\frac{\gamma}{1-\beta}\right)^{k-k_0-2j}\frac{1}{T_k}\frac{\Delta_{2j+k_0-2}+1}{\max\{1-\gamma,\sigma\}} \nonumber\\
&\lesssim \frac{\sum_{j=1}^{\lfloor k/2\rfloor}\left(\frac{\gamma}{1-\beta}\right)^{k-k_0-2j}\Delta_{2j+k_0-2}}{T_k\max\{1-\gamma,\sigma\}} + \frac{1}{T_k(1-\gamma)\max\{1-\gamma,\sigma\}},\label{eq:proof-temp-26}
\end{align}
where the last line uses Lemma \ref{lem:gamma-beta}.

Inserting \eqref{eq:proof-temp-25} and \eqref{eq:proof-temp-26} into \eqref{eq:proof-temp-27}, we have
\begin{align}
&\quad \sum_{j=1}^{\lfloor k/2\rfloor}\gamma^{k-k_0-2j}\prod_{i=j+1}^{\lfloor k/2\rfloor}P_{\pi_{2i+k_0-2}}^{\star}\frac{\mathsf{Var}_{{P}^{\star}}V^{\star,\sigma}}{T_{2j+k_0}} \nonumber\\
&\lesssim \frac{\mu_{\mathsf{min}}(1-\gamma)\gamma^{k}}{\log\frac{C_1}{\varepsilon(1-\gamma)}\log^2T\log\frac{SAT}{\delta}}
+\frac{\sum_{j=1}^{\lfloor k/2\rfloor}\left(\frac{\gamma}{1-\beta}\right)^{k-k_0-2j}\Delta_{2j+k_0-2}}{T_k\max\{1-\gamma,\sigma\}}\nonumber\\
&\quad+ \frac{1}{T_k(1-\gamma)\max\{1-\gamma,\sigma\}}.\label{eq:proof-temp-28}
\end{align}

Inserting \eqref{eq:proof-temp-28} into \eqref{eq:proof-temp-24}, we have
\begin{align}\label{eq:proof-temp-29}
&\quad\sum_{j=1}^{\lfloor k/2\rfloor}\gamma^{{\lfloor k/2\rfloor}-j}\prod_{i=j+1}^{\lfloor k/2\rfloor}P_{\pi_{2i+k_0-2}}^{\star}\sqrt{\frac{\mathsf{Var}_{{P}^{\star}}V^{\star,\sigma}}{T_{2j+k_0}}} \nonumber\\
&\lesssim \gamma^{k/2}\sqrt{\frac{k\mu_{\mathsf{min}}(1-\gamma)}{\log\frac{C_1}{\varepsilon(1-\gamma)}\log^2T\log\frac{SAT}{\delta}}}
+\sqrt{\frac{k\sum_{j=1}^{\lfloor k/2\rfloor}\left(\frac{\gamma}{1-\beta}\right)^{k-k_0-2j}\Delta_{2j+k_0-2}}{T_k\max\{1-\gamma,\sigma\}}} \nonumber\\
&\quad+ \sqrt{\frac{k}{T_k(1-\gamma)\max\{1-\gamma,\sigma\}}}\nonumber\\
&\lesssim\gamma^{k/2}\sqrt{\frac{\mu_{\mathsf{min}}}{\log^2T\log\frac{SAT}{\delta}}}
+\sqrt{\frac{\log\frac{C_1}{(1-\gamma)\varepsilon}\sum_{j=1}^{\lfloor k/2\rfloor}\left(\frac{\gamma}{1-\beta}\right)^{k-k_0-2j}\Delta_{2j+k_0-2}}{T(1-\gamma)\max\{1-\gamma,\sigma\}}}\nonumber\\
&\quad+ \sqrt{\frac{\log\frac{C_1}{(1-\gamma)\varepsilon}}{T(1-\gamma)^2\max\{1-\gamma,\sigma\}}},
\end{align}
where the last line uses \eqref{eq:lem-frac-k-Tk} in Lemma \ref{lem:frac-k-Tk}.

\section{Proof of Lemma \ref{lem:var-V-pistar-sub}}
\label{subsec:proof-lem-var-V-pistar-sub}

We separate the variance discrepancy into the error caused by estimated resampling parameters and the difference between the ideal variances associated with $V_{2j+k_0-2}$ and $V^{\star,\sigma}$:
\begin{align}\label{eq:proof-temp-20}
&\quad \left\|\mathsf{Var}_{\overline{P}^{V_{2j+k_0-2}}}V_{2j+k_0-2}-\mathsf{Var}_{{P}^{\star}}V^{\star,\sigma}\right\|_{\infty}\nonumber\\
&\le \left\|\mathsf{Var}_{\overline{P}^{V_{2j+k_0-2}}}V_{2j+k_0-2}-\mathsf{Var}_{{P}^{V_{2j+k_0-2}}}V_{2j+k_0-2}\right\|_{\infty} + \left\|\mathsf{Var}_{{P}^{V_{2j+k_0-2}}}V_{2j+k_0-2}-\mathsf{Var}_{{P}^{\star}}V^{\star,\sigma}\right\|_{\infty}.
\end{align}

The first term is controlled by the variance bound in Lemma \ref{lem:bound-alpha} and the span bound in Lemma \ref{lem:upper-bound-V}:
\begin{align*}
\left\|\mathsf{Var}_{\overline{P}^{V_{2j+k_0-2}}}V_{2j+k_0-2}-\mathsf{Var}_{{P}^{V_{2j+k_0-2}}}V_{2j+k_0-2}\right\|_{\infty}&\lesssim \sqrt{\frac{\theta_p}{N_{2j+k_0-1}\max\{1-\gamma,\sigma\}^3}} + \frac{\varepsilon_\tau}{\max\{1-\gamma,\sigma\}}\nonumber\\
&\lesssim \sqrt{\frac{\theta_p}{N_{2j+k_0-1}\max\{1-\gamma,\sigma\}^3}} + \frac{(1-\gamma)\varepsilon}{\max\{1-\gamma,\sigma\}},
\end{align*}
as long as $\varepsilon_{\tau}\le (1-\gamma)\varepsilon$.
Furthermore, recalling that $N_{2j+k_0-1} \asymp\mu_{\mathsf{min}}\beta T_{2j+k_0-1}\log_2^{-1}\frac{2}{(1-\gamma)\varepsilon_{\tau}}$, we have
\begin{align}\label{eq:proof-temp-21}
\sqrt{\frac{\theta_p}{N_{2j+k_0-1}\max\{1-\gamma,\sigma\}^3}}
&\lesssim \sqrt{\frac{\theta_p\log T\log\frac{1}{(1-\gamma)\varepsilon_{\tau}}}{T_{2j+k_0-1}\mu_{\mathsf{min}}(1-\gamma)\max\{1-\gamma,\sigma\}^3}}\nonumber\\
&\overset{\text{(i)}}{\lesssim} \sqrt{\frac{\theta_p\log T\log\frac{1}{(1-\gamma)\varepsilon_{\tau}}}{T\mu_{\mathsf{min}}(1-\gamma)\max\{1-\gamma,\sigma\}^3}}\nonumber\\
&\overset{\text{(ii)}}{\lesssim} \sqrt{\frac{1}{\max\{1-\gamma,\sigma\}^2}},
\end{align}
where (i) comes from the fact that $T_{2j+k_0-1}\ge T_1\ge  {\rm e}^{-1}T$, and (ii) comes from the fact that $T\gtrsim \theta_p\log T\log \frac{1}{(1-\gamma)\varepsilon_{\tau}}\mu_{\mathsf{min}}^{-1}(1-\gamma)^{-1}\max\{1-\gamma,\sigma\}^{-1}$.
To control the second term, fix $(s,a)$ and center each value function by its expectation under the corresponding worst-case kernel:
\begin{align}
\overline{V}_{j} &\coloneqq {V}_{j}-P^{V_{j}}_{s,a}V_{j},\qquad 
\overline{V}^{\star,\sigma} \coloneqq {V}^{\star,\sigma}-P^{\star}_{s,a}V^{\star,\sigma}.
\end{align}
Adding and subtracting the squared centered value under $P^\star_{s,a}$ separates the effects of changing the centered value function and changing the transition kernel:
\begin{align}
&\quad \left|\mathsf{Var}_{{P}^{V_{2j+k_0-2}}_{s,a}}V_{2j+k_0-2}-\mathsf{Var}_{{P}^{\star}_{s,a}}V^{\star,\sigma}\right| \nonumber\\
&\le \left|{P}^{{\star}}_{s,a}\left(\overline{V}_{2j+k_0-2}\circ \overline{V}_{2j+k_0-2} - \overline{V}^{\star,\sigma}\circ \overline{V}^{\star,\sigma}\right)\right| + \left|\left({P}^{{V}_{2j+k_0-2}}_{s,a}-{P}^{{\star}}_{s,a}\right)\left(\overline{V}_{2j+k_0-2}\circ \overline{V}_{2j+k_0-2}\right)\right|\nonumber\\
&\le \left\|\overline{V}_{2j+k_0-2} - \overline{V}^{\star,\sigma}\right\|_{\infty}\left\|\overline{V}_{2j+k_0-2} + \overline{V}^{\star,\sigma}\right\|_{\infty} + \left\|{P}^{{V}_{2j+k_0-2}}_{s,a}-{P}^{{\star}}_{s,a}\right\|_1\left\|\overline{V}_{2j+k_0-2}\right\|_{\infty}^2.\label{eq:proof-temp-19}
\end{align}
According to Lemma \ref{lem:upper-bound-V}, we have
\begin{align*}
\left\|\overline{V}_{2j+k_0-2} + \overline{V}^{\star,\sigma}\right\|_{\infty} &\le \max V_{2j+k_0-2} -\min V_{2j+k_0-2} + \max V^{\star,\sigma} - \min V^{\star,\sigma} \lesssim \frac{1}{\max\{1-\gamma,\sigma\}},\nonumber\\
\left\|{P}^{{V}_{2j+k_0-2}}_{s,a}-{P}^{\star}_{s,a}\right\|_1 &\le \left\|{P}^{{V}_{2j+k_0-2}}_{s,a}-{P}^{0}_{s,a}\right\|_1 + \left\|{P}^{0}_{s,a}-{P}^{\star}_{s,a}\right\|_1\le 4\sigma,\nonumber\\ 
\left\|\overline{V}_{2j+k_0-2}\right\|_{\infty}^2 &\le \left(\max V_{2j+k_0-2} - \min V_{2j+k_0-2}\right)^2 \lesssim \frac{1}{\max\{1-\gamma,\sigma\}^2}.
\end{align*}
Furthermore, we have
\begin{align*}
\left\|\overline{V}_{2j+k_0-2} - \overline{V}^{\star,\sigma}\right\|_{\infty}
& = \left\|\left(I - \mathsf{1}P_{s,a}^{\star}\right)\left({V}_{2j+k_0-2} - {V}^{\star,\sigma}\right) + (P_{s,a}^{\star}-P_{s,a}^{V_{2j+k_0-2}})V_{2j+k_0-2}\right\|_{\infty}\nonumber\\
& \le 2\left\|{V}_{2j+k_0-2} - {V}^{\star,\sigma}\right\|_{\infty} + 2\sigma\left(\max {V}_{2j+k_0-2}-\min {V}_{2j+k_0-2}\right)\nonumber\\
& \le 2\left\|{V}_{2j+k_0-2} - {V}^{\star,\sigma}\right\|_{\infty} + C,
\end{align*}
where the last line uses Lemma \ref{lem:upper-bound-V}.
Combining these into \eqref{eq:proof-temp-19} yields
\begin{align*}
&\quad \Big\|\mathsf{Var}_{{P}^{V_{2j+k_0-2}}_{s,a}}V_{2j+k_0-2}-\mathsf{Var}_{{P}^{\star}_{s,a}}V^{\star,\sigma}\Big\|_{\infty}\nonumber\\
&\lesssim \frac{1}{\max\{1-\gamma,\sigma\}}\left(\left\|{V}_{2j+k_0-2} - {V}^{\star,\sigma}\right\|_{\infty} + 2\right) + \frac{1}{\max\{1-\gamma,\sigma\}}.
\end{align*}

The preceding bounds hold uniformly over $(s,a)$. Combining them with \eqref{eq:proof-temp-21} and the decomposition \eqref{eq:proof-temp-20} gives
\begin{align*}
\left\|\mathsf{Var}_{\overline{P}^{V_{2j+k_0-2}}}V_{2j+k_0-2}-\mathsf{Var}_{{P}^{\star}}V^{\star,\sigma}\right\|_{\infty} 
&\lesssim  \frac{\left\|{V}_{2j+k_0-2} - {V}^{\star,\sigma}\right\|_{\infty} + 1}{\max\{1-\gamma,\sigma\}}.
\end{align*}
Finally, the variance discrepancy is bounded by a scalar uniformly over all state-action pairs, so the propagation matrices leave this bound unchanged. Applying Cauchy--Schwarz and the bounds on the stage schedule yields
\begin{align}
&\quad \sum_{j=1}^{\lfloor k/2\rfloor}\gamma^{{\lfloor k/2\rfloor}-j}\prod_{i=j+1}^{\lfloor k/2\rfloor}P_{\pi_{2i+k_0-2}}^{\star}\frac{1}{\sqrt{T_{2j+k_0}}}\sqrt{\left\|\mathsf{Var}_{\overline{P}^{V_{2j+k_0-2}}}V_{2j+k_0-2}-\mathsf{Var}_{{P}^{\star}}V^{\star,\sigma}\right\|_{\infty}} \nonumber\\
&\lesssim
\sum_{j=1}^{\lfloor k/2\rfloor}\gamma^{{\lfloor k/2\rfloor}-j}\sqrt{\frac{\left\|{V}_{2j+k_0-2} - {V}^{\star,\sigma}\right\|_{\infty} + 1}{T_{2j+k_0}\max\{1-\gamma,\sigma\}}}.\label{eq:proof-temp-22}\\
&\le\sqrt{k}\sqrt{\sum_{j=1}^{\lfloor k/2\rfloor}\frac{\gamma^{k-k_0-2j}\left(\left\|{V}_{2j+k_0-2} - {V}^{\star,\sigma}\right\|_{\infty} + 1\right)}{T_{2j+k_0}\max\{1-\gamma,\sigma\}}}\nonumber\\
&\le\sqrt{\frac{k\sum_{j=1}^{\lfloor k/2\rfloor}\left(\frac{\gamma}{1-\beta}\right)^{k-k_0-2j}\left\|{V}_{2j+k_0-2} - {V}^{\star,\sigma}\right\|_{\infty}}{T_k\max\{1-\gamma,\sigma\}}} + \sqrt{\frac{k\sum_{j=1}^{\lfloor k/2\rfloor}\left(\frac{\gamma}{1-\beta}\right)^{k-k_0-2j}}{T_k\max\{1-\gamma,\sigma\}}}\nonumber\\
&\overset{\text{(i)}}{\lesssim}\sqrt{\frac{k\sum_{j=1}^{\lfloor k/2\rfloor}\left(\frac{\gamma}{1-\beta}\right)^{k-k_0-2j}\left\|{V}_{2j+k_0-2} - {V}^{\star,\sigma}\right\|_{\infty}}{T_k\max\{1-\gamma,\sigma\}}} + \sqrt{\frac{k}{T_k(1-\gamma)\max\{1-\gamma,\sigma\}}}\nonumber\\
&\overset{\text{(ii)}}{\lesssim}\sqrt{\frac{\log\frac{C_1}{(1-\gamma)\varepsilon}\sum_{j=1}^{\lfloor k/2\rfloor}\left(\frac{\gamma}{1-\beta}\right)^{k-k_0-2j}\left\|{V}_{2j+k_0-2} - {V}^{\star,\sigma}\right\|_{\infty}}{T(1-\gamma)\max\{1-\gamma,\sigma\}}} + \sqrt{\frac{\log\frac{C_1}{(1-\gamma)\varepsilon}}{T(1-\gamma)^2\max\{1-\gamma,\sigma\}}},\label{eq:proof-Np-3}
\end{align}
where (i) uses Lemma \ref{lem:gamma-beta}, and (ii) uses Lemma \ref{lem:frac-k-Tk}.

\section{Proof of Lemma \ref{lem:sequence}}
\label{subsec:proof-lem-sequence}

We further simplify \eqref{eq:sequence} as 
\begin{align*}
\Delta_k 
&\lesssim  \frac{\gamma^{\lfloor k/2\rfloor}}{1-\gamma} + \sqrt{\frac{G\sum_{j=1}^{\lfloor k/2\rfloor}\left(\frac{\gamma}{1-\beta}\right)^{k-k_0-2j}\Delta_{2j+k_0-2}}{T\mu_{\mathsf{min}}(1-\gamma)^2\max\{1-\gamma,\sigma\}}} + \sqrt{\frac{G\Delta_k}{T\mu_{\mathsf{min}}(1-\gamma)^2\max\{1-\gamma,\sigma\}}} \nonumber\\
&\quad+\sqrt{\frac{G\log_2^2\frac{1}{(1-\gamma)\varepsilon_{\tau}}}{T\mu_{\mathsf{min}}(1-\gamma)^3\max\{1-\gamma,\sigma\}}} + \frac{\varepsilon_{\tau}}{1-\gamma}\nonumber\\
&\lesssim  \frac{\gamma^{\lfloor k/2\rfloor}}{1-\gamma} + \sqrt{\frac{G\sum_{j=1}^{\lfloor k/2\rfloor}\left(\frac{\gamma}{1-\beta}\right)^{k-k_0-2j}\Delta_{2j+k_0-2}}{T\mu_{\mathsf{min}}(1-\gamma)^2\max\{1-\gamma,\sigma\}}}\nonumber\\
&\quad +\sqrt{\frac{G\log_2^2\frac{1}{(1-\gamma)\varepsilon_{\tau}}}{T\mu_{\mathsf{min}}(1-\gamma)^3\max\{1-\gamma,\sigma\}}} + \frac{\varepsilon_{\tau}}{1-\gamma},
\end{align*}
where the last line uses $\Delta_k\le \frac{1}{1-\gamma}$.
Define
$$
\overline{\Delta}_k:=\max_{r\in\{2k,2k+1\},r\le K}\Delta_{r},\qquad 0\le k\le \lfloor K/2\rfloor.
$$ 
The above inequality can be further simplified as 
\begin{align*}
\overline{\Delta}_k 
&\lesssim  \frac{\gamma^{k}}{1-\gamma} + \sqrt{\frac{G\sum_{j=1}^{k}\left(\frac{\gamma}{1-\beta}\right)^{2k-2j}\overline{\Delta}_{j-1}}{T\mu_{\mathsf{min}}(1-\gamma)^2\max\{1-\gamma,\sigma\}}}+\sqrt{\frac{G\log_2^2\frac{1}{(1-\gamma)\varepsilon_{\tau}}}{T\mu_{\mathsf{min}}(1-\gamma)^3\max\{1-\gamma,\sigma\}}} + \frac{\varepsilon_{\tau}}{1-\gamma},
\end{align*}
Define a new sequence $\widetilde{\Delta}_k$ as below.
\begin{align*}
\widetilde{\Delta}_0 &= \max\{\Delta_0,\Delta_1\} \le\frac{1}{1-\gamma},\nonumber\\
\widetilde{\Delta}_k &= \frac{C\gamma^{k}}{1-\gamma} + \sqrt{(1-\gamma)\rho}\sqrt{\sum_{j=1}^{k}\left(\frac{\gamma}{1-\beta}\right)^{2k-2j}\widetilde{\Delta}_{j-1}} + \sqrt{\rho} + \frac{C\varepsilon_{\tau}}{1-\gamma},\quad 1\le k\le \lfloor K/2\rfloor,
\end{align*}
where
\begin{align*}
\rho 
&= \frac{CG\log_2^2\frac{1}{(1-\gamma)\varepsilon_\tau}}{T\mu_{\mathsf{min}}(1-\gamma)^3\max\{1-\gamma,\sigma\}}.
\end{align*}
It is easy to check that $\Delta_k\le \overline{\Delta}_{\lfloor k/2\rfloor}\le\widetilde{\Delta}_{\lfloor k/2\rfloor}$ for $0\le k\le K$.
Thus it suffices to bound $\widetilde{\Delta}_k$.
Now we are ready to prove the following inequality by mathematical induction.
\begin{align}\label{eq:proof-sequence-induction}
\widetilde{\Delta}_k\le \frac{C'\gamma^k}{1-\gamma} + C'\sqrt{\rho} + \frac{C'\varepsilon_{\tau}}{1-\gamma},
\end{align}
where $C'$ is a large enough universal constant.
Under assumption \eqref{eq:thm-T} and \eqref{eq:thm-condition}, we have
\begin{align*}
 C'\sqrt{\rho}
 &\le \frac{\varepsilon}{2},\quad 
\varepsilon_{\tau}
\le \frac{(1-\gamma)\varepsilon}{4C'}.
\end{align*}

Towards proving \eqref{eq:proof-sequence-induction},
it is obvious that \eqref{eq:proof-sequence-induction} holds for $k=0$ with the constant $C'\ge 1$.
Assume that \eqref{eq:proof-sequence-induction} holds for $j\le k-1$.
Then it suffices to prove that
\begin{align}\label{eq:proof-temp-39}
\widetilde{\Delta}_k &= \frac{C\gamma^k}{1-\gamma} + \sqrt{(1-\gamma)\rho}\sqrt{\sum_{j=1}^{k}\left(\frac{\gamma}{1-\beta}\right)^{2k-2j}\widetilde{\Delta}_{j-1}} + \sqrt{\rho} + \frac{C\varepsilon_{\tau}}{1-\gamma}\nonumber\\
&\le \frac{C'\gamma^k}{1-\gamma} + C'\sqrt{\rho} + \frac{C'\varepsilon_{\tau}}{1-\gamma}.
\end{align}
For convenience, define 
$$
\zeta: = \left(\frac{\gamma}{1-\beta}\right)^2 \le \gamma
$$
as long as $\log T\ge 1$.
To prove \eqref{eq:proof-temp-39}, notice that
\begin{align*}
&\quad \frac{C\gamma^k}{1-\gamma} + \sqrt{(1-\gamma)\rho}\sqrt{\sum_{j=1}^{k}\zeta^{k-j}\widetilde{\Delta}_{j-1}} + \sqrt{\rho} +\frac{C\varepsilon_{\tau}}{1-\gamma} \nonumber\\
&\le \frac{C\gamma^k}{1-\gamma} + \sqrt{(1-\gamma)\rho}\sqrt{\sum_{j=1}^{k}\zeta^{k-j}\frac{C'\gamma^{j-1}}{1-\gamma}} + \sqrt{(1-\gamma)\rho}\sqrt{\sum_{j=1}^{k}\zeta^{k-j}C'\left(\sqrt{\rho}+\frac{\varepsilon_{\tau}}{1-\gamma}\right)} + \sqrt{\rho} +\frac{C\varepsilon_{\tau}}{1-\gamma}\nonumber\\
&\le\frac{C\gamma^k}{1-\gamma} + \sqrt{\frac{C'\rho\gamma^{k}}{\gamma-\zeta}} + \sqrt{\frac{C'(1-\gamma)\rho}{1-\zeta}\left(\sqrt{\rho}+\frac{\varepsilon_{\tau}}{1-\gamma}\right)} + \sqrt{\rho}+\frac{C\varepsilon_{\tau}}{1-\gamma},
\end{align*}
where the last line comes from
\begin{align*}
\sum_{j=1}^{k}\zeta^{k-j}\gamma^{j-1} = \frac{\zeta\gamma^{k-1}-\zeta^k}{\gamma-\zeta} + \gamma^{k-1}\le \frac{\zeta\gamma^{k-1}}{\gamma-\zeta}+\gamma^{k-1}=\frac{\gamma^{k}}{\gamma-\zeta}.
\end{align*}

Furthermore, we have
\begin{align}\label{eq:proof-temp-40}
\frac{1-\gamma}{1-\zeta} 
&= \frac{1-\gamma}{1-\frac{\gamma^2}{(1-\beta)^2}} = \frac{(1-\beta)^2(1-\gamma)}{(1-\beta)^2 - \gamma^2}\nonumber\\
& \le\frac{1-\gamma}{1-2\beta - \gamma^2}= \frac{1-\gamma}{(1-\gamma)(1+\gamma-\frac{1}{2\log T})}
= \frac{2\log T}{2(1+\gamma)\log T-1}\le 2,
\end{align}
as long as $\log T\ge 1$, 
and 
\begin{align*}
&\quad \sqrt{\frac{C'(1-\gamma)\rho}{1-\zeta}\left(\sqrt{\rho}+\frac{\varepsilon_{\tau}}{1-\gamma}\right)} + \sqrt{\rho} + \frac{C\varepsilon_{\tau}}{1-\gamma}\nonumber\\
&\le\sqrt{2C'\rho\left(\sqrt{\rho}+\frac{\varepsilon_{\tau}}{1-\gamma}\right)} + \sqrt{\rho} + \frac{C\varepsilon_{\tau}}{1-\gamma}\\
&\le \sqrt{2C'}\left(\sqrt{\rho}+\frac{\varepsilon_{\tau}}{1-\gamma}\right)^{3/2} + \max\{C,1\}\left(\sqrt{\rho} + \frac{\varepsilon_{\tau}}{1-\gamma}\right)\\
&\le \left(\sqrt{2C'}+\max\{C,1\}\right)\left(\sqrt{\rho} + \frac{\varepsilon_{\tau}}{1-\gamma}\right),
\end{align*}
where the last line uses $\sqrt{\rho}+\frac{\varepsilon_{\tau}}{1-\gamma}\le 1$ for $C'\ge 1$.
The inequality $\sqrt{2C'}+\max\{C,1\}\le C'$ holds when $C'\ge 4C+6$.
In this case, we have
\begin{align*}
\sqrt{\frac{C'(1-\gamma)\rho}{1-\zeta}\left(\sqrt{\rho}+\frac{\varepsilon_{\tau}}{1-\gamma}\right)} + \sqrt{\rho} + \frac{C\varepsilon_{\tau}}{1-\gamma}
&\le C'\left(\sqrt{\rho} + \frac{\varepsilon_{\tau}}{1-\gamma}\right).
\end{align*}

In addition, we have
\begin{align*}
\frac{C\gamma^k}{1-\gamma} \le \frac{C'\gamma^k}{2(1-\gamma)},
\end{align*}
as long as $C'\ge 2C$ and
\begin{align*}
\frac{C'\rho\gamma^k}{\gamma-\zeta}\frac{4(1-\gamma)^2}{C'^2\gamma^{2k}} 
&\overset{\text{(i)}}{\le} \frac{\varepsilon^2(1-\gamma)^2}{C'^3\gamma^k(\gamma-\zeta)} \overset{\text{(ii)}}{\le} 4\varepsilon^2\frac{1-\gamma}{C'^3\gamma^{k}}\\
&\le 4\varepsilon^2\frac{1-\gamma}{C'^3\gamma^{\lfloor K/2\rfloor}} \overset{(iii)}{\le} 4C_1\varepsilon^2\frac{1-\gamma}{C'^3\varepsilon(1-\gamma)} = \frac{4C_1\varepsilon}{C'^3}\le 1
\end{align*}
as long as $C'^3\ge 4C_1$, where (i) uses $C'\sqrt{\rho}\le \varepsilon/2$, (ii) uses 
\begin{align}\label{eq:proof-revision-temp-4}
    \gamma - \zeta &= \gamma - \frac{\gamma^2}{(1-\beta)^2} = \gamma \frac{1-2\beta+\beta^2 - \gamma}{(1-\beta)^2} \ge \gamma(1-2\beta- \gamma) \notag\\
    &= \gamma (1-\gamma)\left(1 - \frac{1}{2\log T}\right)\ge \frac{1-\gamma}{4}
\end{align} 
for $\log T\ge 1$ and $\gamma\ge 1/2$,
and (iii) uses 
\begin{align*}
    \gamma^{\lfloor K/2\rfloor} \ge \gamma^{K/2} = \gamma^{\frac{1}{\log\frac{1}{\gamma}}\log\frac{C_1}{\varepsilon(1-\gamma)}} = \frac{(1-\gamma)\varepsilon}{C_1}.
\end{align*}

Based on \eqref{eq:proof-sequence-induction}, we have
\begin{align*}
\Delta_k\le\widetilde{\Delta}_{\lfloor k/2\rfloor} \le\frac{C'\gamma^{\lfloor k/2\rfloor}}{1-\gamma} + C'\sqrt{\rho} + \frac{C'\varepsilon_{\tau}}{1-\gamma}.
\end{align*}
Replacing $C'$ with $C$, we complete the proof.

\section{Proof of key lemmas}
In this section, we shall provide the proof of lemmas presented in Section \ref{subsec:key-lems}.

\subsection{Proof of Lemma \ref{lem:ideal-resample}}
\label{app:proof-lem-ideal-resample}

The existence of $
\tau_{s,a}^{\star}$ defined in \eqref{eq:def-tau} can be immediately established by noting that both \(F_{\mathsf{lower}}(\tau) \coloneqq \sum_{u:V(u)>\tau}P_{s,a}^{0}(u)\) and \(F_{\mathsf{upper}}(\tau) \coloneqq \sum_{u:V(u)\ge \tau}P_{s,a}^{0}(u)\) are monotonically non-increasing step functions, and $F_{\mathsf{lower}}(\max_{s\in\mathcal{S}}V(s))=0$, $F_{\mathsf{upper}}(\min_{s\in\mathcal{S}}V(s))=1$, and \(\sigma\in(0,1]\).

We prove \eqref{eq:law-stprime} in two steps. First, we show that the distribution below solves \eqref{eq:def-Pprime}. We then verify that the ideal resampling procedure generates this distribution.
\begin{align}\label{eq:resample}
P_{s_t,a_t}^V(s)&= \left[\frac{\sum_{u:V(u)\ge V(s)}P_{s_t,a_t}^{0}(u)-\sigma}{\sum_{u:V(u)= V(s)}P_{s_t,a_t}^{0}(u)}\right]_{[0,1]}P_{s_t,a_t}^0(s) + \sigma \delta_{s_{\min}}(s), 
\end{align}
where \([\cdot]_{[0,1]}\coloneqq \min\{\max\{\cdot,0\},1\}\) and \(\delta_{s_{\min}}(s)\) is the Dirac function satisfying \(\delta_{s_{\min}}(s)=1\) if \(s=s_{\min}\) and \(0\) otherwise.

We first consider the case in which the nominal probability mass above the minimum value is less than $\sigma$:
\begin{align}\label{eq:proof-resample-temp-0}
\sum_{u:V(u)>V(s_{\min})} P^0_{s_t,a_t}(u) < \sigma.
\end{align}
Then
$$
\sum_{u:V(u)= V(s_{\min})}P^0_{s_t,a_t}(u) \ge 1- \sigma,\qquad\text{and}\quad  \sum_{u:V(u)\ge V(s)}P^0_{s_t,a_t}(u) < \sigma,\quad \forall V(s)\neq V(s_{\min}).
$$
When $V(s)=V(s_{\min})$, the factor in \eqref{eq:resample} satisfies
$$
\frac{\sum_{u:V(u)\ge V(s)}P_{s_{t},a_{t}}^{0}(u)-\sigma}{\sum_{u:V(u)= V(s)}P_{s_{t},a_{t}}^{0}(u)} = \frac{1-\sigma}{\sum_{u:V(u)= V(s_{\min})}P_{s_{t},a_{t}}^{0}(u)} \in [0,1].
$$
while for \(V(s)\neq V(s_{\min})\), the factor in \eqref{eq:resample} is zero.
Therefore the distribution in \eqref{eq:resample} reduces to
\begin{align*}
P^V(s|s_t,a_t) &= 0,\qquad V(s)\neq V(s_{\min});\notag\\
P^V(s|s_t,a_t) &= \frac{1-\sigma}{\sum_{u:V(u)= V(s_{\min})}P_{s_{t},a_{t}}^{0}(u)}P_{s_{t},a_{t}}^0(s) + \sigma\delta_{s_{\min}}(s),\qquad V(s)= V(s_{\min}).
\end{align*}
We verify that this \(P^V(s|s_t,a_t)\) is a valid distribution:
\begin{align*}
\sum_{s\in\mathcal{S}}P^V(s|s_t,a_t) = \frac{1-\sigma}{\sum_{u:V(u)= V(s_{\min})}P_{s_{t},a_{t}}^{0}(u)}\sum_{s:V(s)= V(s_{\min})}P_{s_{t},a_{t}}^0(s) + \sigma = 1-\sigma + \sigma =1.
\end{align*}
Moreover, its total variation distance from \(P^0_{s_t,a_t}\) is bounded by
\begin{align*}
&\quad \sum_{s\in\mathcal{S}} | P^V(s|s_t,a_t) - P^0_{s_t,a_t}(s)| \notag\\
&\le \sum_{u:V(u)> V(s_{\min})}P_{s_{t},a_{t}}^{0}(u) + \left(1-\frac{1-\sigma}{\sum_{u:V(u)= V(s_{\min})}P_{s_{t},a_{t}}^{0}(u)}\right) \sum_{s:V(s)= V(s_{\min})}P_{s_{t},a_{t}}^0(s)+ \sigma\notag\\
&=\sum_{u:V(u)> V(s_{\min})}P_{s_{t},a_{t}}^{0}(u) + \sum_{u:V(u)= V(s_{\min})}P_{s_{t},a_{t}}^{0}(u) - 1+\sigma+\sigma = 2\sigma,
\end{align*}
and therefore \(P^V(\cdot|s_t,a_t)\) lies in the uncertainty set \(\mathcal{U}^{\sigma}(P^0_{s_t,a_t})\).
Finally, its expected value function is
\begin{align*}
\sum_{s\in\mathcal{S}}P^V(s|s_t,a_t) V(s) = V(s_{\min}).
\end{align*}
Since \(\sum_s\mathcal{P}(s)V(s)\ge V(s_{\min})\) for any distribution \(\mathcal{P}\), this expectation is minimal, and hence \eqref{eq:resample} is one of the solutions to \eqref{eq:def-Pprime}.

We next consider the complementary case:
\begin{align}\label{eq:proof-resample-temp-1}
\sum_{u:V(u)>V(s_{\min})} P^0_{s_t,a_t}(u) \ge \sigma, \qquad \iff\qquad \sum_{u:V(u)=V(s_{\min})} P^0_{s_t,a_t}(u) \le 1-\sigma,
\end{align}
where the equivalence follows from \(\sum_{u:V(u)>V(s_{\min})}P^0_{s_t,a_t}(u)+\sum_{u:V(u)=V(s_{\min})}P^0_{s_t,a_t}(u)=1\). 
For any $\mathcal{P}_{s_t,a_t}\in\mathcal{U}^{\sigma}(P_{s_t,a_t}^0)$, denote its difference from $P^0_{s_t,a_t}$ by $D\in\mathbb{R}^{1\times S}$, i.e.,
$$
D\coloneqq \mathcal{P}_{s_t,a_t} - P_{s_t,a_t}^0.
$$
The optimization problem in \eqref{eq:def-Pprime} is equivalent to $P^{V}(\cdot|s_t,a_t) = P_{s_t,a_t}^0 + D^{\star}$, where \(D^{\star}\) is a solution to the following optimization problem:
\begin{align}\label{eq:opt-D}
    &D^{\star}\in \arg\min \sum_{s\in\mathcal{S}}D(s)(V(s)-V(s_{\min})),\notag\\
    &\text{s.t.}\quad \sum_{s\in\mathcal{S}}|D(s)|\le 2\sigma,\quad -P_{s_t,a_t}^0(s)\le D(s)\le1-P_{s_t,a_t}^0(s),\quad \sum_{s\in\mathcal{S}} D(s)=0.
\end{align}
For ease of notation, we write \(D_{+}\coloneqq \max\{D,0\}\) and \(D_{-}\coloneqq \max\{-D,0\}\), which denote the positive and negative parts of \(D\), respectively.
Then the constraints in \eqref{eq:opt-D} are equivalent to
\begin{align}
  &\sum_{s\in\mathcal{S}}D_{+}(s)\le \sigma, \quad D_{+}(s)\le1-P_{s_t,a_t}^0(s),\notag\\
  & D_{-}(s)\le P_{s_t,a_t}^0(s), \quad \sum_{s\in\mathcal{S}} D_{+}(s)=\sum_{s\in\mathcal{S}} D_{-}(s).  
\end{align}
Consequently,
\begin{align*}
\sum_{s\in\mathcal{S}} D_{+}(s)(V(s)-V(s_{\min})) &\ge 0,\notag\\
\sum_{s\in\mathcal{S}} D_{-}(s)(V(s)-V(s_{\min})) &\le \sum_{u:V(u)>\tau_{s_t,a_t}^{\star}} P^0_{s_t,a_t}(u) (V(u)-V(s_{\min}))\notag\\
&\quad + \Big(\sigma-\sum_{u:V(u)>\tau_{s_t,a_t}^{\star}} P^0_{s_t,a_t}(u)\Big)(\tau_{s_t,a_t}^{\star}-V(s_{\min})),
\end{align*}
Here $\tau_{s_t,a_t}^{\star}$ is defined in \eqref{eq:def-tau}. The first inequality follows from the nonnegativity of $D_+$ and $V-V(s_{\min})$. For the second inequality, the constraints $D_-\le P^0_{s_t,a_t}$ and $\sum_s D_-(s)\le \sigma$ imply that the largest possible contribution is obtained by removing probability mass from the largest values of $V$, with partial removal at $\tau_{s_t,a_t}^{\star}$. The condition \eqref{eq:proof-resample-temp-1} ensures that at least $\sigma$ probability mass is available above the minimum value. Both bounds are attained by the following choice of $D_+^{\star}$ and $D_-^{\star}$:
\begin{align}
D_{+}^{\star}(s) = \left\{
    \begin{array}{ll}
        \sigma, &s=s_{\min},\\
        0,&s\neq s_{\min};
    \end{array}
\right.
\end{align}
and
\begin{align}
D_{-}^{\star}(s) = \left\{
    \begin{array}{ll}
        P^0_{s_t,a_t}(s), &\text{if}~V(s)>\tau_{s_t,a_t}^{\star},\\
        \frac{\sigma-\sum_{u:V(u)>\tau_{s_t,a_t}^{\star}} P^0_{s_t,a_t}(u)}{\sum_{u:V(u)=\tau_{s_t,a_t}^{\star}} P^0_{s_t,a_t}(u)}P^0_{s_t,a_t}(s),&\text{if}~V(s)=\tau_{s_t,a_t}^{\star},\\
        0, &\text{if}~V(s)<\tau_{s_t,a_t}^{\star}.
    \end{array}
\right.
\end{align}
By the definition of $\tau_{s_t,a_t}^{\star}$, $D_-^{\star}$ removes a total mass of $\sigma$ and satisfies $D_-^{\star}\le P^0_{s_t,a_t}$. Moreover, \eqref{eq:proof-resample-temp-1} implies that $P^0_{s_t,a_t}(s_{\min})+\sigma\le 1$. Hence $D_+^{\star}-D_-^{\star}$ is feasible for \eqref{eq:opt-D} and attains the lower bound on its objective.
The corresponding worst-case transition kernel $P^V(\cdot|s_t,a_t)$ is given by
\begin{align}\label{eq:proof-resample-temp-2}
P^V(s|s_t,a_t) &= P^0_{s_t,a_t}(s) +D_{+}^{\star}(s)- D_{-}^{\star}(s) \notag\\
&= \sigma\delta_{s_{\min}}(s) + \left\{
    \begin{array}{ll}
        0, &\text{if}~V(s)>\tau_{s_t,a_t}^{\star},\\
        \frac{\sum_{u:V(u)\ge \tau_{s_t,a_t}^{\star}} P^0_{s_t,a_t}(u)-\sigma}{\sum_{u:V(u)=\tau_{s_t,a_t}^{\star}} P^0_{s_t,a_t}(u)}P^0_{s_t,a_t}(s),&\text{if}~V(s)=\tau_{s_t,a_t}^{\star},\\
        P^0_{s_t,a_t}(s), &\text{if}~V(s)<\tau_{s_t,a_t}^{\star}.
    \end{array}
\right.
\end{align}
Since both expressions share the term \(\sigma\delta_{s_{\min}}\), it suffices to prove that the first term in the right-hand-side of \eqref{eq:resample} coincides with the second term in the right-hand-side of \eqref{eq:proof-resample-temp-2}.
To this end, note that
\begin{align*}
&\sum_{u:V(u)\ge V(s)}P_{s_{t},a_{t}}^{0}(u)\le \sum_{u:V(u)> \tau_{s_t,a_t}^{\star}}P_{s_{t},a_{t}}^{0}(u) <\sigma,\qquad \text{if}~V(s)>\tau_{s_t,a_t}^{\star};\notag\\
&\sum_{u:V(u)\ge V(s)}P_{s_{t},a_{t}}^{0}(u) - \sum_{u:V(u)= V(s)}P_{s_{t},a_{t}}^{0}(u) = \sum_{u:V(u)> V(s)}P_{s_{t},a_{t}}^{0}(u)\notag\\
&\qquad\qquad\qquad\qquad\ge \sum_{u:V(u)\ge \tau_{s_t,a_t}^{\star}}P_{s_{t},a_{t}}^{0}(u)\ge\sigma,\qquad \text{if}~V(s)<\tau_{s_t,a_t}^{\star};
\end{align*}
where we have used the definition of \(\tau_{s_t,a_t}^{\star}\) in \eqref{eq:def-tau}. Therefore the first term on the right-hand side of \eqref{eq:resample} simplifies to
\begin{align}
\left[\frac{\sum_{u:V(u)\ge V(s)}P_{s_{t},a_{t}}^{0}(u)-\sigma}{\sum_{u:V(u)= V(s)}P_{s_{t},a_{t}}^{0}(u)}\right]_{[0,1]}P_{s_{t},a_{t}}^0(s) = \left\{
    \begin{array}{ll}
        0, &\text{if}~V(s)>\tau_{s_t,a_t}^{\star},\\
        \frac{\sum_{u:V(u)\ge\tau_{s_t,a_t}^{\star}} P^0_{s_t,a_t}(u)-\sigma}{\sum_{u:V(u)=\tau_{s_t,a_t}^{\star}} P^0_{s_t,a_t}(u)}P^0_{s_t,a_t}(s),&\text{if}~V(s)=\tau_{s_t,a_t}^{\star},\\
        P^0_{s_t,a_t}(s), &\text{if}~V(s)<\tau_{s_t,a_t}^{\star},
    \end{array}
\right.
\end{align}
which coincides with the second term on the right-hand-side of \eqref{eq:proof-resample-temp-2}.
This proves that \eqref{eq:resample} is a solution to \eqref{eq:def-Pprime}.

It remains to verify that the ideal resampling procedure generates \eqref{eq:resample}. We consider the following five cases, separating $s_{\min}$ because it receives the redirected probability mass.
\begin{itemize}
\item
For $V(s)>\tau_{s_t,a_t}^{\star}$, by the resampling procedure, we have
$$
\mathsf{P}(s_{t+1}^{\prime}=s) = 0,\qquad \text{if}~V(s)>\tau_{s_t,a_t}^{\star},
$$
since $s_{t+1}^{\prime}$ is never set to such states. Moreover, 
we have $\sum_{u:V(u)\ge V(s)} P^0_{s_t,a_t}(u) \le \sum_{u:V(u)>\tau_{s_t,a_t}^{\star}} P^0_{s_t,a_t}(u)<\sigma$, and $s\neq s_{\min}$ as $\tau_{s_t,a_t}^{\star} \ge V(s_{\min})$. Therefore by \eqref{eq:resample}, 
$$
P^V(s|s_t,a_t) = 0,\qquad \text{if}~V(s)>\tau_{s_t,a_t}^{\star}.
$$
\item
For $V(s)<\tau_{s_t,a_t}^{\star}$ and $s\neq s_{\min}$, $s_{t+1}^{\prime}$ is set to $s$ whenever $s_{t+1}=s$, so
$$
\mathsf{P}(s_{t+1}^{\prime}=s) = P_{s_{t},a_{t}}^{0}(s),\qquad \text{if}~V(s)<\tau_{s_t,a_t}^{\star},\quad s\neq s_{\min}.
$$
On the other hand, we have $\sum_{u:V(u)\ge V(s)} P^0_{s_t,a_t}(u)-\sum_{u:V(u)= V(s)} P^0_{s_t,a_t}(u)=\sum_{u:V(u)> V(s)} P^0_{s_t,a_t}(u) \ge \sum_{u:V(u)\ge\tau_{s_t,a_t}^{\star}} P^0_{s_t,a_t}(u)\ge\sigma$. Therefore by \eqref{eq:resample}, 
$$
P^V(s|s_t,a_t) = P_{s_{t},a_{t}}^{0}(s),\qquad \text{if}~V(s)<\tau_{s_t,a_t}^{\star},\quad s\neq s_{\min}.
$$
\item
For $V(s)=\tau_{s_t,a_t}^{\star}$ and $s\neq s_{\min}$, $s_{t+1}^{\prime}$ is set to $s$ with probability $1-p_{s_t,a_t}^{\star}$ whenever $s_{t+1}=s$, and therefore
$$
\mathsf{P}(s_{t+1}^{\prime}=s)=(1-p_{s_t,a_t}^{\star})P_{s_{t},a_{t}}^{0}(s)=\frac{\sum_{u:V(u)\ge\tau_{s_t,a_t}^{\star}} P^0_{s_t,a_t}(u)-\sigma}{\sum_{u:V(u)=\tau_{s_t,a_t}^{\star}} P^0_{s_t,a_t}(u)}P^0_{s_t,a_t}(s),\qquad \text{if}~V(s)=\tau_{s_t,a_t}^{\star},\quad s\neq s_{\min}.
$$
On the other hand, by \eqref{eq:resample},
$$
P^V(s|s_t,a_t)  = \frac{\sum_{u:V(u)\ge\tau_{s_t,a_t}^{\star}} P^0_{s_t,a_t}(u)-\sigma}{\sum_{u:V(u)=\tau_{s_t,a_t}^{\star}} P^0_{s_t,a_t}(u)}P^0_{s_t,a_t}(s),\qquad \text{if}~V(s)=\tau_{s_t,a_t}^{\star},\quad s\neq s_{\min}.
$$
\item
For $s=s_{\min}$ and \eqref{eq:proof-resample-temp-0} holds, we have $\tau_{s_t,a_t}^{\star}=V(s_{\min})$ and $s_{t+1}^{\prime}=s_{\min}$ when $V(s_{t+1})>\tau_{s_t,a_t}^{\star}$, or with probability $p_{s_t,a_t}^{\star}$ when $V(s_{t+1})=\tau_{s_t,a_t}^{\star}$ but $s_{t+1}\neq s_{\min}$, or when $s_{t+1}=s_{\min}$.
Therefore
\begin{align*}
\mathsf{P}(s_{t+1}^{\prime}=s_{\min})&= \sum_{u:V(u)>\tau_{s_t,a_t}^{\star}} P^0_{s_t,a_t}(u)  + p_{s_t,a_t}^{\star} \sum_{u:V(u)=\tau_{s_t,a_t}^{\star}}P_{s_{t},a_{t}}^{0}(u) + (1-p_{s_t,a_t}^{\star})P_{s_{t},a_{t}}^{0}(s_{\min})\notag\\
&=\sum_{u:V(u)>\tau_{s_t,a_t}^{\star}} P^0_{s_t,a_t}(u) +  \sigma - \sum_{u:V(u)>\tau_{s_t,a_t}^{\star}} P^0_{s_t,a_t}(u) + (1-p_{s_t,a_t}^{\star})P_{s_{t},a_{t}}^{0}(s_{\min})\notag\\
&= \frac{\sum_{u:V(u)\ge V(s_{\min})} P^0_{s_t,a_t}(u)-\sigma}{\sum_{u:V(u)=V(s_{\min})}P^0_{s_t,a_t}(u)}P_{s_{t},a_{t}}^{0}(s_{\min}) + \sigma,
\end{align*}
where we have used $\tau_{s_t,a_t}^{\star}=V(s_{\min})$ in the last equation.
On the other hand,
$
P^V(s_{\min}|s_t,a_t)
$
defined in \eqref{eq:resample} coincides with this.
\item For $s=s_{\min}$ and \eqref{eq:proof-resample-temp-0} does not hold, this implies that \eqref{eq:proof-resample-temp-1} holds and $\tau_{s_t,a_t}^{\star} > V(s_{\min})$.
Then $s_{t+1}^{\prime}=s_{\min}$ when $V(s_{t+1})>\tau_{s_t,a_t}^{\star}$, or with probability $p_{s_t,a_t}^{\star}$ when $V(s_{t+1})=\tau_{s_t,a_t}^{\star}$, or when $s_{t+1}=s_{\min}$.
We have
\begin{align*}
\mathsf{P}(s_{t+1}^{\prime}=s_{\min})&= \sum_{u:V(u)>\tau_{s_t,a_t}^{\star}} P^0_{s_t,a_t}(u)  + p_{s_t,a_t}^{\star} \sum_{u:V(u)=\tau_{s_t,a_t}^{\star}}P_{s_{t},a_{t}}^{0}(u) + P^0_{s_t,a_t}(s_{\min})\notag\\
&=\sigma + P^0_{s_t,a_t}(s_{\min}).
\end{align*}
By \eqref{eq:resample}, this coincides with $P^V(s_{\min}|s_t,a_t)$ since
$$
\left[\frac{\sum_{u:V(u)\ge V(s_{\min})} P^0_{s_t,a_t}(u)-\sigma}{\sum_{u:V(u)=V(s_{\min})}P^0_{s_t,a_t}(u)}\right]_{[0,1]} = \left[\frac{1-\sigma}{\sum_{u:V(u)=V(s_{\min})}P^0_{s_t,a_t}(u)}\right]_{[0,1]} = 1.
$$
\end{itemize}
These cases establish \eqref{eq:law-stprime}, completing the proof.

\subsection{Proof of Lemma \ref{lem:bound-alpha}}
\label{subsec:proof-lem-bound-alpha}

We prove that for all $(s,a)\in\mathcal{S}\times\mathcal{A}$,
\begin{align}
  \left|\overline{P}_{s,a}^{V_{k}}V_{k}-{P}_{s,a}^{V_{k}}V_{k}\right| &\lesssim \sqrt{\frac{\theta_p\max\{1-\gamma,\sigma\}}{N_{k+1}}}B+\varepsilon_\tau,\label{eq:lem-bound-alpha-1}\\
\left|\mathsf{Var}_{\overline{P}_{s,a}^{V_{k}}}V_{k} -\mathsf{Var}_{P_{s,a}^{V_{k}}}V_{k}\right| &\lesssim \sqrt{\frac{\theta_p\max\{1-\gamma,\sigma\}}{N_{k+1}}}B^2 + \varepsilon_\tau B, \label{eq:lem-bound-alpha-2} 
\end{align}
which implies Lemma \ref{lem:bound-alpha} obviously.
It suffices to establish \eqref{eq:lem-bound-alpha-1} and \eqref{eq:lem-bound-alpha-2}.
We first control the width of the estimated threshold interval. By Lemma \ref{lem:markovian}, with probability at least $1-\delta$, the number of visits to each $(s,a)$ during stage $k+1$ satisfies
\begin{align*}
L_{T_{k+1}}(s,a)-L_{T_{k}}(s,a) \ge \frac12\mu_{\mathsf{min}}(T_{k+1}-T_{k}) \ge
N_{k+1}\log_2\frac{2}{(1-\gamma)\varepsilon_{\tau}}.
\end{align*}
Thus we have $\tau_2-\tau_1\le \varepsilon_{\tau}$ for any $k,s,a$.

Fix $k$ and $(s,a)$. For notational simplicity, write the estimated parameters $\widehat{p}^{\mathsf{L}}_{k+1,s,a}$, $\widehat{p}^{\mathsf{R}}_{k+1,s,a}$, $\tau^{\mathsf{L}}_{k+1,s,a}$, $\tau^{\mathsf{R}}_{k+1,s,a}$, and $p_{k+1,s,a}$ as $\widehat{p}_1$, $\widehat{p}_2$, $\tau_1$, $\tau_2$, and $p$, respectively. Similarly, write the ideal resampling parameters associated with $V_k$, defined in \eqref{eq:def-tau} and \eqref{eq:def-psa}, as $\tau^{\star}$ and $p^{\star}$.
In addition, define 
\begin{align}\label{eq:def-pstar}
p_1^{\star} &\coloneqq \sum_{u:V_k(u)>\tau_1} P^0(u|s,a),\qquad
p_2^{\star} \coloneqq \sum_{u:V_k(u)>\tau_2} P^0(u|s,a).
\end{align}
Denote
$$
V_{\min}\coloneqq \min_{s\in\mathcal{S}}V_k(s).
$$
By the ideal resampling procedure constructed in Section \ref{subsec:resample}, which has been proved to generate worst-case transition next states in Lemma \ref{lem:ideal-resample}, we have
\begin{subequations}
\begin{align}
  {P}_{s,a}^{V_{k}}V_{k} &= \sum_{u:V_k(u)<\tau^{\star}}P^0(u|s,a)V_k(u) + \sum_{u:V_k(u)=\tau^{\star}}P^0(u|s,a)((1-p^{\star})\tau^{\star} +p^{\star}V_{\min})\notag\\
&\qquad+ \sum_{u:V_k(u)>\tau^{\star}}P^0(u|s,a)V_{\min}.\label{eq:proof-PV}
\end{align}
By resampling procedure in Algorithm \ref{alg:subroutine-resample}, we have
\begin{align}
\overline{P}_{s,a}^{V_{k}}V_{k} &= \sum_{u:V_k(u)\le\tau_1}P^0(u|s,a)V_k(u) + \sum_{u:\tau_1<V_k(u)\le\tau_2}P^0(u|s,a)((1-p)V_k(u)+pV_{\min})\notag\\
&\qquad + \sum_{u:V_k(u)>\tau_2}P^0(u|s,a)V_{\min}\label{eq:proof-barPV}
\end{align}
\end{subequations}

We may replace $\tau_2$ by $\min\{\tau_2,\max_{s\in\mathcal{S}}V_k(s)\}$, and replace $\tau_1$ by $\max\{\tau_1,\min_{s\in\mathcal{S}}V_k(s)+\varepsilon_0\}$ without changing the resampling distribution, where $\varepsilon_0$ is an arbitrarily small positive constant.

We first prove the expectation bound \eqref{eq:lem-bound-alpha-1}. The comparison depends on whether the ideal threshold $\tau^{\star}$ lies inside, below, or above the estimated interval $(\tau_1,\tau_2]$. We consider these three cases separately.


\textbf{Case 1: $\tau^{\star}\in(\tau_1,\tau_2]$}, implying that  $\widehat{p}_1-\widehat{p}_2\neq 0$ and 
$$
p_2^{\star}\le \sum_{u:V_k(u)>\tau^{\star}} P^0(u|s,a) < \sigma.
$$ 
In this case, we can further decompose ${P}_{s,a}^{V_{k}}V_{k}$ as
\begin{align}\label{eq:proof-PV-2}
{P}_{s,a}^{V_{k}}V_{k} &= \sum_{u:V_k(u)\le\tau_1}P^0(u|s,a)V_k(u) + \sum_{u:\tau_1<V_k(u)<\tau^{\star}}P^0(u|s,a)V_k(u) \notag\\
&\qquad+ \sum_{u:V_k(u)=\tau^{\star}}P^0(u|s,a)((1-p^{\star})V_k(u) +p^{\star}V_{\min})+ \sum_{u:\tau^{\star}<V_k(u)\le \tau_2}P^0(u|s,a)V_{\min} \notag\\
&\qquad+ \sum_{u:V_k(u)> \tau_2}P^0(u|s,a)V_{\min}\nonumber\\
&= \sum_{u:V_k(u)\le\tau_1}P^0(u|s,a)V_k(u) + \sum_{u:\tau_1<V_k(u)<\tau^{\star}}P^0(u|s,a)V_k(u) + \sum_{u:V_k(u)=\tau^{\star}}P^0(u|s,a)((1-p^{\star})V_k(u))\notag\\
&\qquad+ \Big(\sum_{u:V_k(u)=\tau^{\star}}P^0(u|s,a)p^{\star}+\sum_{u:\tau^{\star}<V_k(u)\le \tau_2}P^0(u|s,a)\Big)V_{\min} + \sum_{u:V_k(u)> \tau_2}P^0(u|s,a)V_{\min}\notag\\
&= \sum_{u:V_k(u)\le\tau_1}P^0(u|s,a)V_k(u) + \sum_{u:\tau_1<V_k(u)<\tau^{\star}}P^0(u|s,a)V_k(u) \notag\\
&\qquad+ \sum_{u:V_k(u)=\tau^{\star}}P^0(u|s,a)((1-p^{\star})V_k(u))+ (\sigma - p_2^{\star})V_{\min} + \sum_{u:V_k(u)> \tau_2}P^0(u|s,a)V_{\min}
\end{align}
where the last line uses the fact that
\begin{align*}
  \sum_{u:V_k(u)=\tau^{\star}}P^0(u|s,a)p^{\star}+\sum_{u:\tau^{\star}<V_k(u)\le \tau_2}P^0(u|s,a) &= \sigma - \sum_{u:V_k(u)>\tau^{\star}}P^0(u|s,a)+\sum_{u:\tau^{\star}<V_k(u)\le \tau_2}P^0(u|s,a)\notag\\
  &= \sigma - \sum_{u:V_k(u)>\tau_2}P^0(u|s,a) = \sigma - p_2^{\star}.
\end{align*}
Comparing \eqref{eq:proof-barPV} and \eqref{eq:proof-PV-2}, we have
\begin{align*}
\overline{P}_{s,a}^{V_{k}}V_{k}-{P}_{s,a}^{V_{k}}V_{k}&=\sum_{u:\tau_1<V_k(u)\le\tau_2}P^0(u|s,a)((1-p)V_k(u)+pV_{\min})-
\sum_{u:\tau_1<V_k(u)<\tau^{\star}}P^0(u|s,a)V_k(u) \notag\\
&\qquad - \sum_{u:V_k(u)=\tau^{\star}}P^0(u|s,a)((1-p^{\star})V_k(u))- (\sigma - p_2^{\star})V_{\min}\notag\\
&\overset{\text{(i)}}{=}(1-p)\sum_{u:\tau_1<V_k(u)\le\tau_2}P^0(u|s,a) \tau' + p\sum_{u:\tau_1<V_k(u)\le\tau_2}P^0(u|s,a)V_{\min}- \notag\\
&\qquad\Big(\sum_{u:\tau_1<V_k(u)<\tau^{\star}}P^0(u|s,a) + \sum_{u:V_k(u)=\tau^{\star}}P^0(u|s,a)(1-p^{\star})\Big)\tau'' - (\sigma - p_2^{\star})V_{\min}\notag\\
&\overset{\text{(ii)}}{=}(1-p)(p_1^{\star}-p_2^{\star})\tau' + p(p_1^{\star}-p_2^{\star})V_{\min}- (p_1^{\star}-\sigma)\tau'' - (\sigma - p_2^{\star})V_{\min},
\end{align*}
where (i) introduces $\tau'\in(\tau_1,\tau_2]$ and $\tau''\in(\tau_1,\tau^{\star}]$:
\begin{align}
\tau'&\coloneqq \frac{\sum_{u:\tau_1<V_k(u)\le\tau_2}P^0(u|s,a)V_k(u)}{\sum_{u:\tau_1<V_k(u)\le\tau_2}P^0(u|s,a)}\in(\tau_1,\tau_2],\notag\\
\tau''&\coloneqq\left\{
  \begin{array}{ll} 
    \frac{\sum_{u:\tau_1<V_k(u)<\tau^{\star}}P^0(u|s,a)V_k(u) + \sum_{u:V_k(u)=\tau^{\star}}P^0(u|s,a)(1-p^{\star})V_k(u)}{\sum_{u:\tau_1<V_k(u)<\tau^{\star}}P^0(u|s,a) + \sum_{u:V_k(u)=\tau^{\star}}P^0(u|s,a)(1-p^{\star})}\in(\tau_1,\tau^{\star}],& \text{if}~p_1^{\star}>\sigma. \\
    \tau^{\star},&\text{otherwise.}
  \end{array}
  \right.
\end{align}
and (ii) uses
\begin{align}
  &\sum_{u:\tau_1<V_k(u)\le\tau_2}P^0(u|s,a)=p_1^{\star} - p_2^{\star},\label{eq:proof-temp-revision-1}\\
&\sum_{u:\tau_1<V_k(u)<\tau^{\star}}P^0(u|s,a) + \sum_{u:V_k(u)=\tau^{\star}}P^0(u|s,a)(1-p^{\star}) \notag\\
&\quad= \sum_{u:\tau_1<V_k(u)<\tau^{\star}}P^0(u|s,a) + \sum_{u:V_k(u)\ge\tau^{\star}}P^0(u|s,a) - \sigma \notag\\
&\quad = p_1^{\star}-\sigma.\notag
\end{align}
Inserting the expression of $p$
$$
p = \frac{\sigma-\widehat{p}_2}{\widehat{p}_1-\widehat{p}_2},
$$ 
we have
\begin{align}\label{eq:proof-temp-4}
&\quad \left|\overline{P}_{s,a}^{V_{k}}V_{k}-{P}_{s,a}^{V_{k}}V_{k}\right| \nonumber\\
&\le \max_{\tau_1< \tau'\le\tau_2,\tau_1< \tau''\le\tau^{\star}}\left|\frac{\widehat{p}_1-\sigma}{\widehat{p}_1-\widehat{p}_2}\tau' + \frac{\sigma - \widehat{p}_2}{\widehat{p}_1-\widehat{p}_2}V_{\min} -\frac{{p}_1^{\star}-\sigma}{p_1^{\star}-p_2^{\star}}\tau'' - \frac{\sigma - {p}_2^{\star}}{{p}_1^{\star}-{p}_2^{\star}}V_{\min}\right|(p_1^{\star}-p_2^{\star}) \nonumber\\
&\le  \left|\frac{\widehat{p}_1-\sigma}{\widehat{p}_1-\widehat{p}_2} -\frac{{p}_1^{\star}-\sigma}{p_1^{\star}-p_2^{\star}}\right|\left(\tau_2-V_{\min}\right)(p_1^{\star}-p_2^{\star}) + \varepsilon_{\tau}\nonumber\\
&\overset{\text{(i)}}{\le} \left|\frac{\widehat{p}_1-\sigma}{\widehat{p}_1-\widehat{p}_2} -\frac{{p}_1^{\star}-\sigma}{p_1^{\star}-p_2^{\star}}\right|(p_1^{\star}-p_2^{\star})B + \varepsilon_{\tau},
\end{align}
where (i) uses the fact that $\max_s V_k(s) - \min_s V_k(s) \le B$.
We further have
\begin{align}\label{eq:proof-temp-0}
\left(\frac{\widehat{p}_1-\sigma}{\widehat{p}_1-\widehat{p}_2} -\frac{{p}_1^{\star}-\sigma}{p_1^{\star}-p_2^{\star}}\right)(p_1^{\star}-p_2^{\star})
&=\left(\frac{-\sigma p_1^{\star}-\widehat{p}_1p_2^{\star}+\sigma p_2^{\star} +\widehat{p}_1\sigma+\widehat{p}_2p_1^{\star}-\widehat{p}_2\sigma}{\widehat{p}_1-\widehat{p}_2}\right)\nonumber\\
&=\frac{(\widehat{p}_2-p_2^{\star})(\widehat{p}_1-\sigma)+(\widehat{p}_1-p_1^{\star})(\sigma-\widehat{p}_2)}{\widehat{p}_1-\widehat{p}_2}.
\end{align}
Although the bisection thresholds are data-dependent, each threshold is fixed conditional on the history before its estimation batch. Conditional on this history, the next-state samples at the first $N_{k+1}$ subsequent visits to $(s,a)$ are independent draws from $P^0_{s,a}$. We apply Bernstein's inequality conditionally to each batch and take a union bound over all bisection updates, state-action pairs, and stages. 
In particular, Bernstein's inequality gives, for $N_{k+1}\gtrsim \frac{\theta_p}{\max\{1-\gamma,{\sigma}\}}$,
\begin{align*}
    |\widehat{p}_2-p_2^{\star}|\lesssim \sqrt{\frac{\theta_p p_2^{\star}(1-p_2^{\star})}{N_{k+1}}} + \frac{\theta_p}{N_{k+1}}
    \lesssim \sqrt{\frac{\theta_p\max\{\sigma,1-\gamma\} }{N_{k+1}}}
\end{align*}
with probability at least $1-\delta$, where $\theta_p = \log\left(\frac{2SAK}{\delta}\log\left(\frac{1}{(1-\gamma)\varepsilon_\tau}\right)\right)$, and we use the fact that $\widehat{p}_2\le \sigma$.
Thus
\begin{align}\label{eq:proof-temp-1}
\left|\frac{(\widehat{p}_2-p_2^{\star})(\widehat{p}_1-\sigma)}{\widehat{p}_1-\widehat{p}_2}\right|
\le \left|\widehat{p}_2-p_2^{\star}\right|\lesssim \sqrt{\frac{\theta_p \max\{\sigma,1-\gamma\}}{N_{k+1}}}.
\end{align}
To bound the remaining term in \eqref{eq:proof-temp-0}, we distinguish two cases according to the size of the estimated interval mass $\widehat{p}_1-\widehat{p}_2$.
\begin{itemize}
\item
If $\widehat{p}_1-\widehat{p}_2\le10\max\{\sigma,1-\gamma\}$, then we have
\begin{align*}
p_1^{\star} &\le \widehat{p}_1 + \sqrt{\frac{C\theta_p}{N_{k+1}}} \le \widehat{p}_1 - \widehat{p}_2 + \widehat{p}_2 + \sqrt{\frac{C\theta_p}{N_{k+1}}}\nonumber\\
& \le 11\max\left\{\sigma,1-\gamma\right\} + \sqrt{\frac{C\theta_p}{N_{k+1}}} \le 12\max\left\{\sigma,1-\gamma\right\},
\end{align*}
where we use the fact that $\widehat{p}_2\le\sigma$ and $\sqrt{\frac{C\theta_p}{N_{k+1}}}\le \max\left\{\sigma,1-\gamma\right\}$.
Thus we have
\begin{align}\label{eq:proof-temp-2}
\left|\frac{(\widehat{p}_1-p_1^{\star})(\sigma-\widehat{p}_2)}{\widehat{p}_1-\widehat{p}_2}\right|\le \left|\widehat{p}_1-p_1^{\star}\right| \lesssim \sqrt{\frac{\theta_p p_1^{\star}}{N_{k+1}}} + \frac{\theta_p}{N_{k+1}} \lesssim \sqrt{\frac{\theta_p \max\{\sigma,1-\gamma\}}{N_{k+1}}}. 
\end{align}
\item 
If $\widehat{p}_1-\widehat{p}_2>10\max\left\{\sigma,1-\gamma\right\}$, then we have
\begin{align*}
p_1^{\star}= \widehat{p}_1 - \widehat{p}_2 + p_1^{\star} - \widehat{p}_1 + \widehat{p}_2 \le \widehat{p}_1 - \widehat{p}_2 + \max\left\{\sigma,1-\gamma\right\} + \sigma 
\le 1.2(\widehat{p}_1 - \widehat{p}_2),
\end{align*}
where we use the fact that
$
p_1^{\star} - \widehat{p}_1\le \sqrt{\frac{\theta_p}{N_{k+1}}}\le \max\{1-\gamma,\sigma\}
$
for $N_{k+1}\gtrsim\frac{\theta_p}{\max\{1-\gamma,\sigma\}^2}$.
Thus we have
\begin{align}\label{eq:proof-temp-3}
\left|\frac{(\widehat{p}_1-p_1^{\star})(\sigma-\widehat{p}_2)}{\widehat{p}_1-\widehat{p}_2}\right|
&\le \left|\frac{1.2(\widehat{p}_1-p_1^{\star})(\sigma-\widehat{p}_2)}{{p}_1^{\star}}\right|
\le \frac{1.2\sigma}{{p}_1^{\star}}\left|\widehat{p}_1-p_1^{\star}\right|\nonumber\\
&\lesssim \frac{1.2\sigma}{{p}_1^{\star}}\left(\sqrt{\frac{\theta_p p_1^{\star}}{N_{k+1}}} + \frac{\theta_p}{N_{k+1}}\right) 
\lesssim \sqrt{\frac{\theta_p \sigma^2}{N_{k+1} p_1^{\star}}} + \frac{\theta_p\sigma}{N_{k+1}p_1^{\star}}\nonumber\\
&\lesssim \sqrt{\frac{\theta_p \max\{\sigma,1-\gamma\}}{N_{k+1}}}, 
\end{align}
where we use the fact that $p_1^{\star}\ge \widehat{p}_1 + p_1^{\star} - \widehat{p}_1 \ge 10\max\{\sigma,1-\gamma\} - \sqrt{C\theta_p/N_{k+1}} \ge 9\max\{\sigma,1-\gamma\}$.
\end{itemize}
Inserting \eqref{eq:proof-temp-3}, \eqref{eq:proof-temp-2}, and \eqref{eq:proof-temp-1} into \eqref{eq:proof-temp-0}, we have
\begin{align*}
    \left|\left(\frac{\widehat{p}_1-\sigma}{\widehat{p}_1-\widehat{p}_2} -\frac{{p}_1^{\star}-\sigma}{p_1^{\star}-p_2^{\star}}\right)(p_1^{\star}-p_2^{\star})\right|\lesssim \sqrt{\frac{\theta_p \max\{\sigma,1-\gamma\}}{N_{k+1}}}.
\end{align*}
Inserting it into \eqref{eq:proof-temp-4}, we have
\begin{align*}
\left|\overline{P}_{s,a}^{V_{k}}V_{k}-{P}_{s,a}^{V_{k}}V_{k}\right|\lesssim \sqrt{\frac{\theta_p \max\{\sigma,1-\gamma\}}{N_{k+1}}}B +\varepsilon_\tau.
\end{align*}

\textbf{Case 2: $\tau^{\star}\le{\tau}_1$}, implying that $p_2^{\star}\le p_1^{\star}\le\sigma$.
By \eqref{eq:proof-barPV}, we further decompose the expectation under $\overline{P}_{s,a}^{V_k}$ as
\begin{align}
\overline{P}_{s,a}^{V_k} V_k &= \sum_{u:V_k(u)<\tau^{\star}}P^0(u|s,a)V_k(u) + \sum_{u:V_k(u)=\tau^{\star}}P^0(u|s,a)V_k(u)\notag\\
&\qquad+ \sum_{u:\tau^{\star}<V_k(u)\le \tau_1}P^0(u|s,a)V_k(u) + \sum_{u:\tau_1<V_k(u)\le \tau_2}P^0(u|s,a)((1-p)V_k(u)+pV_{\min})\notag\\
&\qquad+\sum_{u:V_k(u)> \tau_2}P^0(u|s,a)V_{\min}.
\end{align}
Similarly, by \eqref{eq:proof-PV}, we further decompose the expectation under ${P}_{s,a}^{V_k}$ as
\begin{align}
{P}_{s,a}^{V_k} V_k &= \sum_{u:V_k(u)<\tau^{\star}}P^0(u|s,a)V_k(u) + \sum_{u:V_k(u)=\tau^{\star}}P^0(u|s,a)((1-p^{\star})V_k(u) + p^{\star}V_{\min})\notag\\
&\qquad+ \sum_{u:\tau^{\star}<V_k(u)\le \tau_1}P^0(u|s,a)V_{\min} + \sum_{u:\tau_1<V_k(u)\le \tau_2}P^0(u|s,a)V_{\min}\notag\\
&\qquad+\sum_{u:V_k(u)> \tau_2}P^0(u|s,a)V_{\min}.
\end{align}
Subtracting the two expressions gives
\begin{align}
\overline{P}_{s,a}^{V_k} V_k - {P}_{s,a}^{V_k} V_k
&=\sum_{u:V_k(u)=\tau^{\star}}P^0(u|s,a)p^{\star}(V_k(u)-V_{\min})+\sum_{u:\tau^{\star}<V_k(u)\le \tau_1}P^0(u|s,a)(V_k(u)-V_{\min})\notag\\
&\qquad + \sum_{u:\tau_1<V_k(u)\le \tau_2}P^0(u|s,a)(1-p)(V_k(u)-V_{\min})\label{eq:proof-temp-revision-2}\\
&\le \Big(\sum_{u:V_k(u)=\tau^{\star}}P^0(u|s,a)p^{\star}+\sum_{u:\tau^{\star}<V_k(u)\le \tau_1}P^0(u|s,a)\Big)(\tau_1-V_{\min})\notag\\
&\qquad + \sum_{u:\tau_1<V_k(u)\le \tau_2}P^0(u|s,a)(1-p)(\tau_2-V_{\min})\notag\\
&=(\sigma-p_1^{\star})(\tau_1-V_{\min}) + (1-p)(p_1^{\star}-p_2^{\star})(\tau_2-V_{\min}),\notag
\end{align}
where the last line uses \eqref{eq:proof-temp-revision-1} and the fact that
\begin{align*}
\sum_{u:V_k(u)=\tau^{\star}}P^0(u|s,a)p^{\star}+\sum_{u:\tau^{\star}<V_k(u)\le \tau_1}P^0(u|s,a) &= \sigma - \sum_{u:V_k(u)>\tau^{\star}}P^0(u|s,a)+ \sum_{u:\tau^{\star}<V_k(u)\le \tau_1}P^0(u|s,a)\notag\\
&= \sigma - \sum_{u:V_k(u)>\tau_1}P^0(u|s,a) = \sigma-p_1^{\star}.
\end{align*}
All terms in \eqref{eq:proof-temp-revision-2} are nonnegative. Thus the expectation difference is nonnegative, and its absolute value satisfies
\begin{align*}
\left|\overline{P}_{s,a}^{V_{k}}V_{k}-{P}_{s,a}^{V_{k}}V_{k}\right|
&\le \left|\frac{\widehat{p}_1-\sigma}{\widehat{p}_1-\widehat{p}_2}(p_1^{\star}-p_2^{\star})\left(\tau_2 - V_{\min}\right) + (\sigma-p_1^{\star})\left(\tau_1 - V_{\min}\right)\right|\nonumber\\
&\le \left|\frac{\widehat{p}_1-\sigma}{\widehat{p}_1-\widehat{p}_2}(p_1^{\star}-p_2^{\star})\right|B + (\sigma-p_1^{\star})B.
\end{align*}
The preceding display assumes $\widehat{p}_1\neq\widehat{p}_2$. When $\widehat{p}_1=\widehat{p}_2$, the estimated interval has collapsed, and the bound reduces to
\begin{align*}
\left|\overline{P}_{s,a}^{V_{k}}V_{k}-{P}_{s,a}^{V_{k}}V_{k}\right|\le \left|(\sigma-p_1^{\star})\left(\tau_1 - V_{\min}\right)\right|
\le  (\sigma-p_1^{\star})B.
\end{align*}
Before bounding the above terms, we first notice that provided $N_{k+1}\max\{\sigma,1-\gamma\}\ge \theta_p$,
\begin{align}\label{eq:proof-temp-5}
\widehat{p}_1\le p_1^{\star} + \sqrt{\frac{C\theta_p p_1^{\star}}{4N_{k+1}}} +\frac{\sqrt{C}\theta_p}{2N_{k+1}}\le \sigma + \sqrt{\frac{C\theta_p \max\{\sigma,1-\gamma\}}{N_{k+1}}}.
\end{align}
We make the following discussions.
\begin{itemize}
\item 
If $p_1^{\star}-p_2^{\star}\le 4\sqrt{\frac{C\theta_p \max\{\sigma,1-\gamma\}}{N_{k+1}}}$, then we have
$
\left|\frac{\widehat{p}_1-\sigma}{\widehat{p}_1-\widehat{p}_2}(p_1^{\star}-p_2^{\star})\right| \le p_1^{\star}-p_2^{\star}\lesssim \sqrt{\frac{\theta_p \max\{\sigma,1-\gamma\}}{N_{k+1}}}.
$
\item
If $p_1^{\star}-p_2^{\star}> 4\sqrt{\frac{C\theta_p \max\{\sigma,1-\gamma\}}{N_{k+1}}}$, then we have
$
\widehat{p}_1-\widehat{p}_2\ge p_1^{\star}-p_2^{\star} - 2\sqrt{\frac{C\theta_p\max\{\sigma,1-\gamma\}}{N_{k+1}}} \ge \frac12(p_1^{\star}-p_2^{\star}),
$
and 
$
\left|\frac{\widehat{p}_1-\sigma}{\widehat{p}_1-\widehat{p}_2}(p_1^{\star}-p_2^{\star})\right| \lesssim \widehat{p}_1-\sigma \overset{}{\lesssim} \sqrt{\frac{\theta_p \max\{\sigma,1-\gamma\}}{N_{k+1}}},
$
where the last inequality uses \eqref{eq:proof-temp-5}.
\item
In addition, we have that
$
0\le \sigma - p_1^{\star} = \sigma - \widehat{p}_1 + \widehat{p}_1 - p_1^{\star}
\le \widehat{p}_1 - p_1^{\star} \lesssim \sqrt{\frac{\theta_p \max\{\sigma,1-\gamma\}}{N_{k+1}}},
$
which comes from $\sigma - \widehat{p}_1\le 0$.
\end{itemize}
Combining the above conclusions, we have
\begin{align*}
\left|\overline{P}_{s,a}^{V_{k}}V_{k}-{P}_{s,a}^{V_{k}}V_{k}\right|\lesssim  \sqrt{\frac{\theta_p \max\{\sigma,1-\gamma\}}{N_{k+1}}}B.
\end{align*}

\textbf{Case 3: $\tau^{\star}>{\tau}_2$}, implying that $p_1^{\star}\ge p_2^{\star}\ge\sigma$. 
By \eqref{eq:proof-barPV} and \eqref{eq:proof-PV}, we have
\begin{align*}
{P}_{s,a}^{V_{k}}V_{k}-\overline{P}_{s,a}^{V_{k}}V_{k} &= \sum_{u:\tau_1<V_k(u)\le \tau_2}P^0(u|s,a)p(V_k(u)-V_{\min}) \notag\\
&\qquad+\sum_{u:V_k(u)=\tau^{\star}}P^0(u|s,a)(1-p^{\star})(V_k(u)-V_{\min})+\sum_{u:\tau_2<V_k(u)< \tau^{\star}}P^0(u|s,a)(V_k(u)-V_{\min})\notag\\
&\le \sum_{u:\tau_1<V_k(u)\le \tau_2}P^0(u|s,a)p(\tau_2-V_{\min}) \notag\\
&\qquad +\Big(\sum_{u:V_k(u)=\tau^{\star}}P^0(u|s,a)(1-p^{\star})+\sum_{u:\tau_2<V_k(u)< \tau^{\star}}P^0(u|s,a)\Big)(\tau^{\star}-V_{\min})\notag\\
&\le p(p_1^{\star}-p_2^{\star})(\tau_2-V_{\min}) + (p_2^{\star}-\sigma)(\tau^{\star}-V_{\min}),\notag
\end{align*}
where the last line arises from
\begin{align*}
\sum_{u:V_k(u)=\tau^{\star}}P^0(u|s,a)(1-p^{\star})+\sum_{u:\tau_2<V_k(u)< \tau^{\star}}P^0(u|s,a) &= \sum_{u:V_k(u)\ge \tau^{\star}}P^0(u|s,a) - \sigma + \sum_{u:\tau_2<V_k(u)< \tau^{\star}}P^0(u|s,a)\notag\\
& = \sum_{u:V_k(u)> \tau_2}P^0(u|s,a)- \sigma = p_2^{\star} - \sigma.
\end{align*}
Then we have
\begin{align*}
\left|\overline{P}_{s,a}^{V_{k}}V_{k}-{P}_{s,a}^{V_{k}}V_{k}\right|
&\le \left|\frac{\sigma-\widehat{p}_2}{\widehat{p}_1-\widehat{p}_2}(p_1^{\star}-p_2^{\star})\left(\tau_2 - V_{\min}\right) + (p_2^{\star}-\sigma)\left(\tau^{\star} - V_{\min}\right)\right|\nonumber\\
&\le \left|\frac{\sigma - \widehat{p}_2}{\widehat{p}_1-\widehat{p}_2}(p_1^{\star}-p_2^{\star})\right|B + (p_2^{\star}-\sigma)B.
\end{align*}
Similarly, we assume that $\widehat{p}_1-\widehat{p}_2\neq 0$.
Otherwise we have
\begin{align*}
\left|\overline{P}_{s,a}^{V_{k}}V_{k}-{P}_{s,a}^{V_{k}}V_{k}\right|
\le \left|(p_2^{\star}-\sigma)\left(\tau^{\star} - V_{\min}\right)\right|
\le(p_2^{\star}-\sigma)B.
\end{align*}
Moreover, we notice that
\begin{align*}
p_2^{\star}\le \widehat{p}_2 + |p_2^{\star} - \widehat{p}_2| \le \sigma +\sqrt{\frac{C\theta_p p_2^{\star}}{4N_{k+1}}}+ \frac13\sqrt{\frac{C\theta_p\max\{\sigma,1-\gamma\}}{N_{k+1}}},
\end{align*}
where we use the fact that $\widehat{p}_2\le \sigma$, and further leads to
\begin{align*}
  p_2^{\star}&\le \frac32\max\{\sigma,1-\gamma\} + \frac{C\theta_p}{4N_{k+1}} \le 2\max\{\sigma,1-\gamma\},\notag\\
  p_2^{\star}&\le\widehat{p}_2+ \sqrt{\frac{C\theta_p\max\{\sigma,1-\gamma\}}{N_{k+1}}} \le \sigma + \sqrt{\frac{C\theta_p\max\{\sigma,1-\gamma\}}{N_{k+1}}}.
\end{align*}
Then we have:
\begin{itemize}
\item If $p_1^{\star} - p_2^{\star} >4\sqrt{\frac{C\theta_p}{N_{k+1}}}$, then we have
$
\widehat{p}_1-\widehat{p}_2 \ge p_1^{\star} - p_2^{\star} -2\sqrt{\frac{C\theta_p}{N_{k+1}}} \ge \frac{1}{2}(p_1^{\star} - p_2^{\star}),
$
and 
$
\frac{\sigma - \widehat{p}_2}{\widehat{p}_1-\widehat{p}_2}(p_1^{\star}-p_2^{\star}) \le 2(\sigma - \widehat{p}_2) \lesssim \sqrt{\theta_p\frac{\max\{\sigma,1-\gamma\}}{N_{k+1}}}.
$
\item If $p_1^{\star} - p_2^{\star} \le 4\sqrt{\frac{C\theta_p}{N_{k+1}}}$, then we have
\begin{align*}
p_1^{\star} \le p_2^{\star} + 4\sqrt{\frac{C\theta_p}{N_{k+1}}} \le 6\max\{\sigma,1-\gamma\}.
\end{align*}
If $p_1^{\star} - p_2^{\star} \le 20\sqrt{\frac{C\theta_p\max\{\sigma,1-\gamma\}}{N_{k+1}}}$, then we have
$
\frac{\sigma - \widehat{p}_2}{\widehat{p}_1-\widehat{p}_2}(p_1^{\star}-p_2^{\star}) \le p_1^{\star}-p_2^{\star} \lesssim \sqrt{\frac{\theta_p\max\{\sigma,1-\gamma\}}{N_{k+1}}}.
$
Otherwise, that is $ 20\sqrt{\frac{C\theta_p\max\{\sigma,1-\gamma\}}{N_{k+1}}}< p_1^{\star} - p_2^{\star} \le 4\sqrt{\frac{C\theta_p}{N_{k+1}}}$, then we have
$
\widehat{p}_1 - \widehat{p}_2 \ge p_1^{\star} - p_2^{\star} - (\sqrt{2}+\sqrt{6})\sqrt{\frac{C\theta_p\max\{\sigma,1-\gamma\}}{N_{k+1}}}  \ge \frac{4}{5}(p_1^{\star} - p_2^{\star})
$
and 
$
\frac{\sigma - \widehat{p}_2}{\widehat{p}_1-\widehat{p}_2}(p_1^{\star}-p_2^{\star}) \le 2(\sigma - \widehat{p}_2) \le 2(p_2^{\star} - \widehat{p}_2) \lesssim \sqrt{\frac{\theta_p\max\{\sigma,1-\gamma\}}{N_{k+1}}}.
$
\item In addition, we have
$
\sigma\ge \widehat{p}_2\ge p_2^{\star} - \sqrt{\frac{C\theta_p\max\{\sigma,1-\gamma\} }{N_{k+1}}}.
$
Thus we have
$
p_2^{\star} - \sigma \lesssim  \sqrt{\frac{\theta_p\max\{\sigma,1-\gamma\} }{N_{k+1}}}.
$
\end{itemize}
The three cases establish \eqref{eq:lem-bound-alpha-1}.

We next prove the variance bound \eqref{eq:lem-bound-alpha-2}, using the same case distinction. We separate the effect of changing the centering mean from that of redistributing probability mass. The former is controlled by the expectation bound just established, while the latter is controlled by the probability-mass estimates obtained in the corresponding cases above.

For $\tau^{\star}\in(\tau_1,\tau_2]$, we have
\begin{align}\label{eq:proof-revision-3}
\left|\mathsf{Var}_{\overline{P}_{s,a}^{V_{k}}}V_{k} -\mathsf{Var}_{P_{s,a}^{V_{k}}}V_{k}\right| 
&\le \max_{\min V_k\le \tau\le \max V_k}|\left(\tau-\overline{P}_{s,a}^{V_k}V_k\right)^2-\left(\tau-{P}_{s,a}^{V_k}V_k\right)^2|\notag\\&\quad + \max_{\tau_1< \tau'\le\tau_2,\tau_1< \tau''\le\tau^{\star}}\bigg|\frac{\widehat{p}_1-\sigma}{\widehat{p}_1-\widehat{p}_2}\left(\tau'-\overline{P}_{s,a}^{V_k}V_k\right)^2 + \frac{\sigma - \widehat{p}_2}{\widehat{p}_1-\widehat{p}_2}\left(V_{\min} -\overline{P}_{s,a}^{V_k}V_k\right)^2\nonumber\\
&\quad -\frac{{p}_1^{\star}-\sigma}{p_1^{\star}-p_2^{\star}}\left(\tau''-{P}_{s,a}^{V_k}V_k\right)^2 - \frac{\sigma - {p}_2^{\star}}{{p}_1^{\star}-{p}_2^{\star}}\left(V_{\min}-{P}_{s,a}^{V_k}V_k\right)^2\bigg|(p_1^{\star}-p_2^{\star}) \nonumber\\
&\le  2B\left|\overline{P}_{s,a}^{V_k}V_k-{P}_{s,a}^{V_k}V_k\right| + \left|\frac{\widehat{p}_1-\sigma}{\widehat{p}_1-\widehat{p}_2} -\frac{{p}_1^{\star}-\sigma}{p_1^{\star}-p_2^{\star}}\right|\left(\tau_2-V_{\min}\right)^2(p_1^{\star}-p_2^{\star})\nonumber\\
&\quad+\max_{\tau_1< \tau'\le\tau_2,\tau_1< \tau''\le\tau^{\star}}\left|\left(\tau'-\overline{P}_{s,a}^{V_k}V_k\right)^2 -\left(\tau''-{P}_{s,a}^{V_k}V_k\right)^2 \right|\nonumber\\
&\quad +\left|\left(V_{\min}-\overline{P}_{s,a}^{V_k}V_k\right)^2-\left(V_{\min}-{P}_{s,a}^{V_k}V_k\right)^2\right|\nonumber\\
&\overset{\text{(i)}}{\le} \left|\frac{\widehat{p}_1-\sigma}{\widehat{p}_1-\widehat{p}_2} -\frac{{p}_1^{\star}-\sigma}{p_1^{\star}-p_2^{\star}}\right|(p_1^{\star}-p_2^{\star})B^2 + 2B\left(\varepsilon_\tau + |\overline{P}_{s,a}^{V_k}V_k-{P}_{s,a}^{V_k}V_k|\right)\nonumber\\
& \quad+ 4B\left|\overline{P}_{s,a}^{V_k}V_k-{P}_{s,a}^{V_k}V_k\right|\nonumber\\
&\overset{\text{(ii)}}{\lesssim}\sqrt{\frac{\theta_p \max\{\sigma,1-\gamma\}}{N_{k+1}}}B^2 + \varepsilon_\tau B,
\end{align}
where (i) uses the fact that $|\tau'-\overline{P}_{s,a}^{V_k}V_k| \le B$, $|\tau''-{P}_{s,a}^{V_k}V_k| \le B$, $|V_{\min}-\overline{P}_{s,a}^{V_k}V_k| \le B$, and $|V_{\min}-{P}_{s,a}^{V_k}V_k| \le B$,
and (ii) uses \eqref{eq:lem-bound-alpha-1}.
.

For $\tau^{\star}\le{\tau}_1$, we assume $\widehat{p}_1\neq\widehat{p}_2$ and have
\begin{align}\label{eq:proof-revision-4}
\left|\mathsf{Var}_{\overline{P}_{s,a}^{V_{k}}}V_{k} -\mathsf{Var}_{P_{s,a}^{V_{k}}}V_{k}\right| 
&\le\max_{\min V_k\le \tau\le \max V_k}|\left(\tau-\overline{P}_{s,a}^{V_k}V_k\right)^2-\left(\tau-{P}_{s,a}^{V_k}V_k\right)^2|\notag\\&\quad + \max_{\tau_1\le \tau'\le\tau_2}\frac{\widehat{p}_1-\sigma}{\widehat{p}_1-\widehat{p}_2}(p_1^{\star}-p_2^{\star})\left|\left(\tau'-\overline{P}_{s,a}^{V_k}V_k\right)^2 - \left(V_{\min}-{P}_{s,a}^{V_k}V_k\right)^2\right|\nonumber\\
&\quad+ (\sigma-p_1^{\star})\max_{\tau^{\star}\le \tau''\le\tau_1}\left|\left(\tau'' - \overline{P}_{s,a}^{V_k}V_k\right)^2 + \left(V_{\min}-{P}_{s,a}^{V_k}V_k\right)^2\right|\nonumber\\
&\lesssim B|\overline{P}_{s,a}^{V_k}V_k-{P}_{s,a}^{V_k}V_k|  + \left|\frac{\widehat{p}_1-\sigma}{\widehat{p}_1-\widehat{p}_2}(p_1^{\star}-p_2^{\star})\right|B^2 + (\sigma-p_1^{\star})B^2\notag\\
&\lesssim \sqrt{\frac{\theta_p \max\{\sigma,1-\gamma\}}{N_{k+1}}}B^2.
\end{align}
The case of $\widehat{p}_1-\widehat{p}_2= 0$ can be handled in a similar way by replacing the factor $\frac{\widehat{p}_1-\sigma}{\widehat{p}_1-\widehat{p}_2}$ with $0$ in \eqref{eq:proof-revision-4}.

For $\tau^{\star} > \tau_2$, we assume $\widehat{p}_1\neq\widehat{p}_2$ and have
\begin{align}\label{eq:proof-revision-5}
\left|\mathsf{Var}_{\overline{P}_{s,a}^{V_{k}}}V_{k} -\mathsf{Var}_{P_{s,a}^{V_{k}}}V_{k}\right| 
&\le \max_{\min V_k\le \tau\le \max V_k}|\left(\tau-\overline{P}_{s,a}^{V_k}V_k\right)^2-\left(\tau-{P}_{s,a}^{V_k}V_k\right)^2|\notag\\&\quad +\frac{\sigma-\widehat{p}_2}{\widehat{p}_1-\widehat{p}_2}(p_1^{\star}-p_2^{\star})\max_{\tau_1\le\tau'\le\tau_2}\left|\left(\tau'-{P}_{s,a}^{V_{k}}V_k\right)^2 - \left(V_{\min}-\overline{P}_{s,a}^{V_{k}}V_k\right)^2\right|\nonumber\\
&\quad + (p_2^{\star}-\sigma)\max_{\tau_2\le\tau''\le\tau^{\star}}\left|\left(\tau''-{P}_{s,a}^{V_{k}}V_k\right)^2 -\left(V_{\min}-\overline{P}_{s,a}^{V_{k}}V_k\right)^2\right|\nonumber\\
&\lesssim B|\overline{P}_{s,a}^{V_k}V_k-{P}_{s,a}^{V_k}V_k|+\left|\frac{\sigma - \widehat{p}_2}{\widehat{p}_1-\widehat{p}_2}(p_1^{\star}-p_2^{\star})\right|B^2 + (p_2^{\star}-\sigma)B^2\notag\\
&\lesssim\sqrt{\frac{\theta_p \max\{\sigma,1-\gamma\}}{N_{k+1}}}B^2.
\end{align}
The case of $\widehat{p}_1-\widehat{p}_2= 0$ can be handled in a similar way by replacing the factor $\frac{\sigma-\widehat{p}_2}{\widehat{p}_1-\widehat{p}_2}$ with $0$ in \eqref{eq:proof-revision-5}.
Combining the three cases establishes \eqref{eq:lem-bound-alpha-2} and completes the proof.

\subsection{Proof of Lemma \ref{lem:upper-bound-V}}
\label{subsec:proof-lem-upper-bound-V}

The proof of \eqref{eq:lem-upper-bound-V-star} is similar to that of Lemma 6 in \cite{shi2024curious}.
According to the robust Bellman consistency equation \eqref{eq:robust-Bellman}, we have
\begin{align*}
\max_{s,a} Q^{\star,\sigma}&\le 1 + \gamma\left((1-\sigma)\max_{s,a} Q^{\star,\sigma} + \sigma \min_{s,a} V^{\star,\sigma}\right),\quad
\min_{s,a} Q^{\star,\sigma}\ge\gamma \min_s V^{\star,\sigma} \ge \gamma \min_{s,a} Q^{\star,\sigma}.
\end{align*}
Combining the above two inequalities, we have
\begin{align*}
\max_{s,a} Q^{\star,\sigma} - \min_{s,a} Q^{\star,\sigma} &\le 1 + \gamma\left((1-\sigma)\max_{s,a} Q^{\star,\sigma} + \frac{\sigma}{\gamma} \min_{s,a} Q^{\star,\sigma}\right) - \min_{s,a} Q^{\star,\sigma}\notag\\ &\le 1 + \gamma(1-\sigma)\max_{s,a} Q^{\star,\sigma} - (1-\sigma) \min_{s,a} Q^{\star,\sigma}\notag\\ &\le 1 + \gamma (1-\sigma)\left(\max_{s,a} Q^{\star,\sigma} - \min_{s,a} Q^{\star,\sigma}\right).
\end{align*}
Thus we have
\begin{align*}
\max_{s,a} Q^{\star,\sigma} - \min_{s,a} Q^{\star,\sigma}
\le \frac{1}{1-\gamma(1-\sigma)}\le \frac{1}{\max\{\sigma,1-\gamma\}}.
\end{align*}

Next we intend to prove \eqref{eq:lem-upper-bound-Q-T0}.
According to the update rule of $Q_t$ in \eqref{eq:alg-proof-update}, we have
\begin{align*}
Q_{\lfloor(1-\beta)T_1\rfloor} 
&= \prod_{t=1}^{(1-\beta)T_1}(1-\Lambda_t)Q_{0}  + \sum_{t=1}^{(1-\beta)T_1}\Lambda_t^{(1-\beta)T_1}r + \gamma\sum_{t=1}^{(1-\beta)T_1}\Lambda_t^{(1-\beta)T_1}P_tV_{-1}
= \sum_{t=1}^{(1-\beta)T_1}\Lambda_t^{(1-\beta)T_1}r,
\end{align*}
which is due to the fact that $Q_0 =0$ and $V_{-1}=0$.
Thus we have
\begin{align*}
0\le Q_{(1-\beta)T_1}\le \left\|\sum_{t=1}^{(1-\beta)T_1}\Lambda_t^{(1-\beta)T_1}\right\|\le 1
\end{align*}
and complete the proof.

It remains to prove \eqref{eq:lem-upper-bound-V-k}. We first express the stage-end update as an ideal robust update plus an error term, and then control the resulting span recursively. Expanding \eqref{eq:alg-proof-update} over stage $k$ gives
\begin{align*}
Q_{T_k} 
&= \prod_{t=T_{k-1}+1}^{T_k}(I-\Lambda_t)Q_{T_{k-1}} + \gamma\sum_{t=T_{k-1}+1}^{T_k}\Lambda_t^{T_k}P^{V_{k-2}}V_{k-2} + \sum_{t=T_{k-1}+1}^{T_k}\Lambda_t^{T_k}r\nonumber\\
&\quad + \gamma\sum_{t=T_{k-1}+1}^{T_k}\Lambda_t^{T_k}\left(P_t - \overline{P}^{V_{k-2}}\right)V_{k-2} + \gamma\sum_{t=T_{k-1}+1}^{T_k}\Lambda_t^{T_k}\left( \overline{P}^{V_{k-2}}- P^{V_{k-2}}\right)V_{k-2}\nonumber\\
&=(I-\widetilde{\Lambda}_k)Q_{T_{k-1}}  + \widetilde{\Lambda}_k (r + \gamma e_k)+ \gamma\widetilde{\Lambda}_kP^{V_{k-2}}V_{k-2},
\end{align*}
where $\widetilde{\Lambda}_k$ collects the update weights and $e_k$ combines the sampling noise and the resampling bias:
\begin{align*}
\widetilde{\Lambda}_k 
&= \sum_{t=T_{k-1}+1}^{T_k}\Lambda_t^{T_k},\qquad 
e_k = (\widetilde{\Lambda}_k)^{-1}\sum_{t=T_{k-1}+1}^{T_k}\Lambda_t^{T_k}\left(P_t - \overline{P}^{V_{k-2}}\right)V_{k-2} + \left( \overline{P}^{V_{k-2}}- P^{V_{k-2}}\right)V_{k-2}.
\end{align*}
Define $Q_k^{\max} = \max_{s,a}Q_{T_k}$ and $Q_k^{\min} = \min_{s,a}Q_{T_k}$ for $k\ge 1$,
and $Q_0^{\max} = \max_{s,a}Q_{(1-\beta)T_1}$ and $Q_0^{\min} = \min_{s,a}Q_{(1-\beta)T_1}$.
Then we have
\begin{align*}
Q_{T_k}
&\le (I-\widetilde{\Lambda}_k)\mathsf{1}Q_{k-1}^{\max} + \gamma\widetilde{\Lambda}_k\mathsf{1}((1-\sigma)\max_sV_{k-2}+\sigma \min_s V_{k-2}) + \widetilde{\Lambda}_k\mathsf{1} + \gamma\widetilde{\Lambda}_k\mathsf{1}\max_{s,a}e_k,\nonumber\\
Q_{T_k}
&\ge (I-\widetilde{\Lambda}_k)\mathsf{1}Q_{k-1}^{\min} + \gamma\widetilde{\Lambda}_k\mathsf{1}\min_s V_{k-2} + \gamma\widetilde{\Lambda}_k\mathsf{1}\min_{s,a} e_k.
\end{align*}
Notice that 
\begin{align}
\left\|I-\widetilde{\Lambda}_k\right\|
&=\left\|\prod_{t=T_{k-1}+1}^{T_k}(I-\Lambda_t)\right\|\overset{\text{(i)}}{\lesssim} \frac{1}{T^2},\label{eq:proof-temp-45}\\
\left\|\widetilde{\Lambda}_k\right\| 
&\ge 1 - \left\|I-\widetilde{\Lambda}_k\right\| \ge 1-\frac{C}{T^2} \overset{\text{(ii)}}{\ge} \frac12,\label{eq:proof-temp-18}
\end{align}
where $C$ is a sufficiently large constant, (i) comes from \eqref{eq:lem-bound-eta-1} in Lemma \ref{lem:eta}, and (ii) holds as long as $T\ge 2C$.
Then we further have
\begin{align*}
Q_k^{\max}
&\le \frac{C}{T^2}Q_{k-1}^{\max} + \gamma((1-\sigma)\max_sV_{k-2}+\sigma \min_s V_{k-2}) + 1 + \gamma\left\|e_k\right\|_{\infty},\nonumber\\
Q_k^{\min}
&\ge \gamma\left(1-\frac{C}{T^2}\right)\min_s V_{k-2}(s) - \gamma\left\|e_k\right\|_{\infty}.
\end{align*}
Combining the above two inequalities, we have
\begin{align}
Q_k^{\max} - Q_k^{\min}
&\le \gamma(1-\sigma)(\max_sV_{k-2}-\min_sV_{k-2}) + 1+2\gamma\left\|e_k\right\|_{\infty}+\frac{C}{T^2}Q_{k-1}^{\max} + \frac{C\gamma}{T^2}\min_sV_{k-2}\nonumber\\
&\overset{\text{(i)}}{\le} \gamma(1-\sigma)(\max_sV_{k-2}-\min_sV_{k-2}) + 2+2\gamma\left\|e_k\right\|_{\infty}\nonumber\\
&\overset{\text{(ii)}}{\le} \gamma(1-\sigma)(Q_{k-2}^{\max}-Q_{k-2}^{\min}) + 2 + 2\gamma\left\|e_k\right\|_{\infty},\label{eq:proof-Qmax-Qmin-decom}
\end{align}
where (i) uses the fact that 
$$
\frac{C}{T^2}Q_{k-1}^{\max}\le \frac{C}{T^2(1-\gamma)}\le \frac14\le\frac{\gamma}{2},\qquad \frac{C}{T^2}\min_sV_{k-2}\le \frac{C}{T^2(1-\gamma)}\le \frac14\le\frac{\gamma}{2},
$$
and (ii) uses the fact that $\max_sV_{k-2}-\min_sV_{k-2}\le Q_{k-2}^{\max} - Q_{k-2}^{\min}$.
We shall use mathematical induction to prove that for any $k\ge 0$,
\begin{align}\label{eq:proof-temp-44}
    Q_{k}^{\max} - Q_k^{\min} \le \frac{4}{1-\gamma(1-\sigma)},
\end{align}
which implies \eqref{eq:lem-upper-bound-V-k}.
According to \eqref{eq:lem-upper-bound-Q-T0}, we have \eqref{eq:proof-temp-44} holds for $k=0$ and $k=1$ because
\begin{align*}
Q_{0}^{\max} - Q_0^{\min} \le 1 \le \frac{4}{1-\gamma(1-\sigma)},\qquad Q_{1}^{\max} - Q_1^{\min} \le 1 \le \frac{4}{1-\gamma(1-\sigma)}.
\end{align*}
For the induction step, fix $k\ge 2$ and assume that \eqref{eq:proof-temp-44} holds for all $0\le j\le k-1$. In view of \eqref{eq:proof-Qmax-Qmin-decom}, it suffices to show that $\|e_k\|_{\infty}\le 1$. To this end, we bound the sampling noise and the resampling bias separately. Following the derivations of \eqref{eq:proof-temp-47} and \eqref{eq:proof-temp-48}, the update weights satisfy
\begin{align*}
   \sum_{t=T_{k-1}+1}^{T_k}(\Lambda_t^{T_k}(s,a))^2 
   &\lesssim \frac{\log^2T}{\mu_{\mathsf{min}}(1-\gamma)T_k},
\quad   \max_{T_{k-1}< t\le T_k}\left\|\Lambda_t^{T_k} \right\|
   \lesssim \frac{\log^2T}{\mu_{\mathsf{min}}(1-\gamma)T_k},\notag\\
    \left\|\widetilde{\Lambda}_k^{-1}\right\|&\le \frac{1}{1-\|I-\widetilde{\Lambda}_k\|}\le 2.
   \end{align*}
By the induction hypothesis, $\max_{s\in\mathcal{S}}V_{k-2}(s)-\min_{s\in\mathcal{S}}V_{k-2}(s)\lesssim \max\{\sigma,1-\gamma\}^{-1}$. Combining Lemma \ref{lem:bound-alpha} with Bernstein's inequality, as in the derivation of \eqref{eq:proof-lem-sample-2}, yields
\begin{align}
\left\|e_k\right\|_{\infty}&\lesssim\sqrt{\log\frac{SAT}{\delta}}\sqrt{\frac{\log^2T}{\mu_{\mathsf{min}}(1-\gamma)T_k}}\frac{1}{1-\gamma} + \frac{\log\frac{SAT}{\delta}\log^2 T}{\mu_{\mathsf{min}}(1-\gamma)^2T_k}+ \sqrt{\frac{\theta_p}{N_{k-1}\max\{\sigma,1-\gamma\}}}+ \varepsilon_{\tau}\nonumber\\
&\overset{\text{(i)}}{\lesssim}\sqrt{\log\frac{SAT}{\delta}}\sqrt{\frac{\log^2T}{\mu_{\mathsf{min}}(1-\gamma)^3 T}} + \frac{\log\frac{SAT}{\delta}\log^2 T}{\mu_{\mathsf{min}}(1-\gamma)^2 T}+ \sqrt{\frac{\theta_p\log T\log_2\frac{2}{(1-\gamma)\varepsilon_{\tau}}}{T\mu_{\mathsf{min}}(1-\gamma)\max\{\sigma,1-\gamma\}}} + \varepsilon_{\tau},\label{eq:proof-temp-16}
\end{align}
where (i) uses the fact that $T_k\ge T_1\gtrsim {\rm e}^{-1}T$ in \eqref{eq:lem-bound-T1} and 
$$
N_{k-1}=\Big\lfloor\frac12\mu_{\mathsf{min}}(T_{k-1}-T_{k-2})\log_2^{-1}\frac{2}{(1-\gamma)\varepsilon_{\tau}}\Big\rfloor\gtrsim \mu_{\mathsf{min}}\beta T_{1}\log_2^{-1}\frac{2}{(1-\gamma)\varepsilon_{\tau}}\gtrsim T\mu_{\mathsf{min}}(1-\gamma)\log^{-1}T\log_2^{-1}\frac{2}{(1-\gamma)\varepsilon_{\tau}}.
$$
Recalling that $T\ge \frac{C_2\theta_p\log^2 T\log\frac{2}{(1-\gamma)\varepsilon_{\tau}}}{\mu_{\mathsf{min}}(1-\gamma)^3\varepsilon^2\max\{\sigma,1-\gamma\}}$, we have
\begin{align*}
&\sqrt{\log\frac{SAT}{\delta}}\sqrt{\frac{\log^2T}{\mu_{\mathsf{min}}(1-\gamma)^3T}} 
\lesssim \varepsilon\le 1,
\quad \frac{\log\frac{SAT}{\delta}\log^2 T}{\mu_{\mathsf{min}}(1-\gamma)^2 T} \lesssim \varepsilon^2\le 1,\nonumber\\
&
\varepsilon_{\tau}
\lesssim(1-\gamma)\varepsilon \lesssim1,\quad 
\sqrt{\frac{\theta_p\log T\log_2\frac{2}{(1-\gamma)\varepsilon_{\tau}}}{T\mu_{\mathsf{min}}(1-\gamma)\max\{\sigma,1-\gamma\}}} \lesssim \varepsilon\le 1.
\end{align*}
Inserting into \eqref{eq:proof-temp-16}, as long as $C_2$ is sufficiently large, we have
\begin{align}\label{eq:proof-temp-17}
\left\|e_k\right\|_{\infty}\le 1.
\end{align}
Substituting \eqref{eq:proof-temp-17} into \eqref{eq:proof-Qmax-Qmin-decom} and applying the induction hypothesis yields
\begin{align*}
Q_k^{\max} - Q_k^{\min}
&\le\gamma(1-\sigma)(Q_{k-2}^{\max}-Q_{k-2}^{\min}) +  4 \nonumber\\
&\le\gamma(1-\sigma)\frac{4}{1-\gamma(1-\sigma)} + 4 
= \frac{4}{1-\gamma(1-\sigma)}.
\end{align*}
Noticing that 
$$
\frac{4}{1-\gamma(1-\sigma)}\lesssim \frac{1}{\max\{\sigma,1-\gamma\}},
$$
we complete the proof.

\section{Proof of Preliminary Lemmas}

\subsection{Proof of Lemma \ref{lem:frac-k-Tk}}
\label{subsec:proof-lem-frac-k-Tk}

We first prove \eqref{eq:lem-bound-T1}.
Considering that $\beta\in(0,1/2)$, we have
\begin{align*}
    (1-\beta)^{K}T&\ge  \exp\left(-2\beta K\right)T \ge \exp\left(-\frac{1-\gamma}{\log T}\frac{\log(\frac{C_1}{(1-\gamma)\varepsilon})}{1-\gamma}\right)T \nonumber\\
    &\overset{\text{(i)}}{\ge} \exp\left(-1\right)T
    \ge \frac{33\log^2 T}{\mu_{\mathsf{min}}(1-\gamma)},
\end{align*}
where (i) arises from the fact that $\log T\ge \log\frac{C_1}{(1-\gamma)\varepsilon}$ for sufficiently large $C_2$.
Then we have
\begin{align*}
    (1-\beta)T_1\ge(1-\beta)((1-\beta)^{K-1}T-1) = (1-\beta)^KT - (1-\beta)\ge \frac{33\log^2 T}{\mu_{\mathsf{min}}(1-\gamma)} - 1\ge \frac{32\log^2 T}{\mu_{\mathsf{min}}(1-\gamma)}.
\end{align*}

Now we intend to prove \eqref{eq:lem-frac-k-Tk}.
According to the definition of $T_k = (1-\beta)^{K-k}T $, we have
\begin{align*}
\frac{k}{T_k} = \frac{k(1-\beta)^k}{(1-\beta)^KT},\quad k\ge 1.
\end{align*}
We shall show that
\begin{align}\label{eq:proof-temp-50}
\frac{k-1}{T_{k-1}} \le \frac{k}{T_k},\quad 2\le k\le K.
\end{align}
It suffices to prove that
$\frac{k-1}{k}\frac{T_k}{T_{k-1}} = \left(1-\frac1k\right)\frac{1}{1-\beta}\le 1,
$
that is
\begin{align*}
\beta\le \frac1k,\quad 2\le k\le K.
\end{align*}
Notice that 
$$
K \le \frac{2\log\frac{C_1}{(1-\gamma)\varepsilon}}{1-\gamma} \overset{\text{(i)}}{\le} \frac{2\log T}{1-\gamma}\le \frac{1}{\beta},
$$
where (i) uses the lower bound of $T$ in \eqref{eq:thm-T}.
Thus we prove \eqref{eq:proof-temp-50} and have
\begin{align*}
\frac{k}{T_k} \le \frac{K}{T_K} \le \frac{2\log\frac{C_1}{(1-\gamma)\varepsilon}}{(1-\gamma)T}
\end{align*}
and complete the proof.

\subsection{Proof of Lemma \ref{lem:markovian}}
\label{subsec:proof-lem-markovian}

According to Lemma 8 in \citet{li2021sample}, it suffices to prove that
$$
T_k-T_h \ge\frac{443t_{\mathsf{mix}}}{\mu_{\mathsf{min}}}\log \frac{4SAK^2}{\delta}.
$$
According to the definition of $T_k$, we have
$$
T_k-T_h \ge T_k-T_{k-1} \ge \frac{\beta}{2} T_k \ge \frac{\beta}{2} T_1 \ge \frac{(1-\gamma)T}{8{\rm e}\log T} \ge\frac{443t_{\mathsf{mix}}}{\mu_{\mathsf{min}}}\log\frac{4SAK^2}{\delta},
$$
for sufficiently large $C_2$ in \eqref{eq:thm-T}.

\subsection{Proof of Lemma \ref{lem:eta}}
\label{subsec:proof-lem-eta}

According to \eqref{eq:lem-bound-T1}, we have
\begin{align}
    T_k\ge (1-\beta)T_1\ge {\rm e}^{-1} T-1 \overset{\text{(i)}}{\ge} \frac{32\log^2 T}{\mu_{\mathsf{min}}(1-\gamma)} = \frac{8\log T}{\mu_{\mathsf{min}}\beta},\label{eq:bound-T1}
\end{align}
where (i) comes from \eqref{eq:thm-T} with $C_2\ge 100{\rm e}$.
Based on this property, for $0\le t\le (1-\beta)T_k$, we have
\begin{align*}
\Lambda_t^{T_k}(s_{t-1},a_{t-1}) &= \prod_{i=t+1}^{T_k}(1-\Lambda_i(s_{t-1},a_{t-1}))\Lambda_t(s_{t-1},a_{t-1})\le \prod_{i=(1-\beta)T_k+1}^{T_k}(1-\Lambda_i(s_{t-1},a_{t-1}))\nonumber\\
&\le (1-\eta_k)^{L_{T_k}(s_{t-1},a_{t-1})-L_{(1-\beta)T_k}(s_{t-1},a_{t-1})}\le (1-\eta_k)^{\frac12\mu_{\mathsf{min}}\beta T_k}\nonumber\\
&=\left(1-\frac{8\log T}{8\log T+\mu_{\mathsf{min}}\beta T_k}\right)^{\frac12\mu_{\mathsf{min}}\beta T_k}
\le\exp\left(-\frac{8\log T}{8\log T+\mu_{\mathsf{min}}\beta T_k}\frac12\mu_{\mathsf{min}}\beta T_k\right)\nonumber\\
&\overset{\text{(i)}}{\le} \exp\left(-\frac{4\log T}{\mu_{\mathsf{min}}\beta T_k}\frac12\mu_{\mathsf{min}}\beta T_k\right)
=\frac{1}{T^2},
\end{align*}
where (i) uses \eqref{eq:bound-T1}.
For $(s,a)\neq(s_{t-1},a_{t-1})$, the $\Lambda_t^{T_k}(s,a)=0$, and \eqref{eq:lem-bound-eta-1} holds obviously.

Moreover, for $(1-\beta)T_k<t\le T_k$, we have
\begin{align*}
\Lambda_t^{T_k}(s_{t-1},a_{t-1}) \le \Lambda_t(s_{t-1},a_{t-1}) = \eta_k= \frac{8\log T}{8\log T+\mu_{\mathsf{min}}\beta T_k}\overset{\text{}}{\le} \frac{8\log T}{\mu_{\mathsf{min}}\beta T_k},
\end{align*}
For $(s,a)\neq(s_{t-1},a_{t-1})$, the $\Lambda_t^{T_k}(s,a)=0$, and \eqref{eq:lem-bound-eta-2} holds obviously.

The last inequality holds because for $1\le t< T_k$,
\begin{align*}
I &= 1-\Lambda_{T_k} + \Lambda_{T_k} = (1 - \Lambda_{T_k})(1-\Lambda_{T_{k-1}} + \Lambda_{T_{k-1}}) + \Lambda_{T_k}
\nonumber\\
&=\prod_{t=1}^{T_k}(1 - \Lambda_t) + \sum_{t=1}^{T_k}\prod_{i=t+1}^{T_k}(1-\Lambda_i)\Lambda_t
= \Lambda_0^{T_k} + \sum_{t=1}^{T_k}\Lambda_t^{T_k} = \sum_{t=0}^{T_k}\Lambda_t^{T_k}.
\end{align*}

\subsection{Proof of Lemma \ref{lem:gamma-beta}}
\label{subsec:proof-lem-gamma-beta}

According to the definition of $\gamma$ and $\beta$, we have
\begin{align*}
\frac{1}{1-\frac{\gamma}{1-\beta}} 
&= \frac{4\log T-1+\gamma}{4\log T-1+\gamma-4\gamma\log T} = \frac{4\log T-1+\gamma}{(4\log T - 1)(1-\gamma)}\overset{\text{(i)}}{\lesssim} \frac{1}{1-\gamma},\\
\frac{1}{1-\frac{\gamma}{\sqrt{1-\beta}}} 
&=\left(1+\frac{\gamma}{\sqrt{1-\beta}}\right)\frac{1}{1-\frac{\gamma^2}{1-\beta}}\le \frac{3}{1-\frac{\gamma^2}{1-\beta}} =\frac{3(4\log T-1+\gamma)}{4\log T-1+\gamma-4\gamma^2\log T}\nonumber\\
&=\frac{3(4\log T-1+\gamma)}{(4\log T(1+\gamma) -1)(1-\gamma)}
\overset{\text{(ii)}}{\lesssim} \frac{1}{1-\gamma}
\end{align*}
where (i) and (ii) hold as long as $\log T\ge 1$. 

\subsection{Proof of Lemma \ref{lem:infinity-norm}}
\label{subsec:proof-lem-infinity-norm}

Define $\varepsilon = \left\|V - V'\right\|_{\infty}$.
Notice that for all $s\in\mathcal{S}$, we have $V(s)\le V'(s) + \varepsilon$.
Then
\begin{align*}
P_{s,a}^{V}V &= \inf_{P_{s,a}\in\mathcal{U}^{\sigma}(P^0_{s,a})}P_{s,a}V \le \inf_{P_{s,a}\in\mathcal{U}^{\sigma}(P^0_{s,a})}P_{s,a}(V'+\varepsilon\mathsf{1})\\
&= \inf_{P_{s,a}\in\mathcal{U}^{\sigma}(P^0_{s,a})}P_{s,a}V' + \varepsilon\mathsf{1}
=P_{s,a}^{V'}V' + \varepsilon\mathsf{1}.
\end{align*}
Similarly, we have
$P_{s,a}^{V'}V'
\le P_{s,a}^{V}V + \varepsilon\mathsf{1}.$
Thus we have 
$    \left\|P^{V}V - P^{V'}V'\right\|_{\infty} \le \varepsilon = \left\|V - V'\right\|_{\infty}.$

\section{Algorithm for synchronous setting}
\label{subsec:alg-syn}

The efficient Q-learning algorithm for robust RL in the synchronous setting is presented in Algorithm \ref{alg}. 
This algorithm extends Algorithm \ref{alg:asyn} by simultaneously updating $Q_t(s,a)$ and the parameter $\widehat{p}_{s,a}^{\mathsf{mid}}$ for all state-action pairs in each iteration, with parameters
\begin{align}
T_0 &= 0,\qquad T_k = \lfloor(1-\beta)^{K-k}T\rfloor,\quad 1\le k\le K;\label{eq:def-Tk-syn}\\
N_k &=\left\lfloor (T_{k}-T_{k-1})\log_2^{-1}\frac{2}{(1-\gamma)\varepsilon_{\tau}} \right\rfloor.\label{eq:def-Np-syn}
\end{align}

Algorithm \ref{alg} maintains the same set of variables as Algorithm \ref{alg:asyn}, resulting in an identical space complexity of $O(SA)$.

\begin{algorithm}[p]
\begingroup
\SetAlgoLined
\DontPrintSemicolon
\SetAlFnt{\small\linespread{1.12}\selectfont}
\renewcommand{\CommentSty}[1]{\textnormal{\textcolor{blue}{\texttt{#1}}}\unskip}
\SetAlgoNlRelativeSize{-1}
\SetInd{0.6em}{0.9em}
\SetKwInOut{AlgInput}{Input}
\SetKwInOut{AlgOutput}{Output}
\caption{Efficient synchronous Q-learning for robust RL.}
\label{alg}
\AlgInput{Learning rates $\{\eta_k\}$, total budget $T$, number of stages $K$, tolerance $\varepsilon_\tau$, discount factor $\gamma$, uncertainty level $\sigma$, stage boundaries $\{T_k\}$ from \eqref{eq:def-Tk-syn}, and sample counts $\{N_k\}$ from \eqref{eq:def-Np-syn}.}
\AlgOutput{Q-function $Q_T$.}
Initialize $Q_0\leftarrow0$, $V_{-1},V_0\leftarrow0$, and $\tau_{0,s,a}^{\mathsf{L}},\tau_{0,s,a}^{\mathsf{R}},p_{0,s,a}\leftarrow0$ for all $(s,a)\in\mathcal{S}\times\mathcal{A}$\;
\For{$k=1,\ldots,K$}{
  Initialize the following for all state-action pairs:\;
  $\begin{aligned}[t]
    &\tau_{k,s,a}^{\mathsf{R}}\leftarrow(1-\gamma)^{-1},\quad
      \tau_{k,s,a}^{\mathsf{L}}\leftarrow-0.1,\quad
      \widehat p_{k,s,a}^{\mathsf{L}}\leftarrow1,\quad
      \widehat p_{k,s,a}^{\mathsf{R}}\leftarrow0,\\
    &\tau_{s,a}^{\mathsf{mid}}\leftarrow0.5/(1-\gamma)-0.05,\quad
      \widehat p_{s,a}^{\mathsf{mid}}\leftarrow0,\quad n_{s,a}\leftarrow0
  \end{aligned}$\;
  \For{$t=T_{k-1},\ldots,T_k-1$}{
    \For{$(s,a)\in\mathcal S\times\mathcal A$}{
    \tcp{Transition observation}
    Draw $s_{t+1}(s,a)\sim P^0(\cdot\mid s,a)$\;
    \tcp{Resampling}
    $\begin{aligned}[t]
      s_{t+1}^{\prime}(s,a)\leftarrow{}&\textnormal{\textsc{Resample}}\bigl(V_{k-2},s_{t+1}(s,a),
      \tau_{k-1,s,a}^{\mathsf{L}},\tau_{k-1,s,a}^{\mathsf{R}},p_{k-1,s,a}\bigr)
    \end{aligned}$\;
    \tcp{Q-function update}
    $\begin{aligned}[t]
      Q_{t+1}(s,a)\leftarrow{}&(1-\eta_k)Q_t(s,a)+\eta_k\bigl(r(s,a)+\gamma V_{k-2}(s_{t+1}^{\prime}(s,a))\bigr)
    \end{aligned}$\;
    \tcp{Online parameter estimation}
        \If{$\tau_{k,s,a}^{\mathsf{R}}-\tau_{k,s,a}^{\mathsf{L}}>\varepsilon_\tau$}{
$\begin{aligned}[t]
          &\bigl(\Theta_{s,a},n_{s,a}\bigr)\leftarrow\textnormal{\textsc{ResampParamsUpd}}\bigl(
            V_{k-1},s_{t+1}(s,a),\sigma,\Theta_{s,a},n_{s,a},N_k\bigr),
        \end{aligned}$
        where $\Theta_{s,a} = (\tau_{s,a}^{\mathsf{mid}},\widehat p_{s,a}^{\mathsf{mid}}, \tau_{k,s,a}^{\mathsf{L}},\tau_{k,s,a}^{\mathsf{R}},
            \widehat p_{k,s,a}^{\mathsf{L}},\widehat p_{k,s,a}^{\mathsf{R}})$, and assignments to $\Theta_{s,a}$ update its components accordingly.
        }
    }
  }
  \tcp{Value function update}
  Compute $V_k(s)\leftarrow\max_{a'\in\mathcal A}Q_{T_k}(s,a')$ for all $s\in\mathcal S$\;
  \tcp{Resampling probability computation}
  \For{$(s,a)\in\mathcal S\times\mathcal A$}{
    \eIf{$\widehat p_{k,s,a}^{\mathsf{L}}\le\widehat p_{k,s,a}^{\mathsf{R}}$}{
      $p_{k,s,a}\leftarrow0$\;
    }{
      $p_{k,s,a}\leftarrow(\sigma-\widehat p_{k,s,a}^{\mathsf{R}})/(\widehat p_{k,s,a}^{\mathsf{L}}-\widehat p_{k,s,a}^{\mathsf{R}})$\;
    }
  }
}
\endgroup
\end{algorithm}

\section{Numerical simulations}
\label{sec:exp}

\begin{figure}[t]
    \centering
    \includegraphics[width=0.45\textwidth]{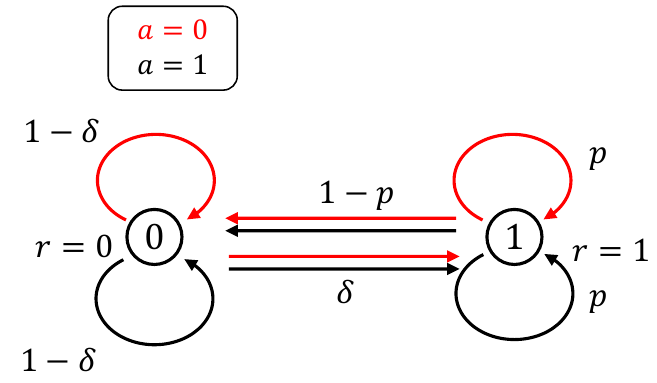}
    \caption{Illustrations of the considered RMDP.}
    \label{fig:RMDP}
\end{figure}

\revision{
This section presents numerical simulations to illustrate the parameter dependence predicted by our theoretical bounds in Section \ref{sec:mainthm}. 
We consider both the asynchronous and the synchronous setting and run Algorithms \ref{alg:asyn} and \ref{alg} on the RMDP \(\mathcal{M} = \{\mathcal{S}, \mathcal{A},\gamma,\mathcal{U}^{\sigma}(P^0), r\}\) shown in Figure \ref{fig:RMDP}. 
This RMDP has two states \(\mathcal{S} = \{0,1\}\) and two actions \(\mathcal{A} = \{0,1\}\) at each state. The nominal transition kernel \(P^0\) satisfies 
$$
P^0(0|0,0) = P^0(0|0,1) = 1-\delta, \qquad \text{and}\qquad P^0(1|1,0) =P^0(1|1,1) = p,
$$
with $\delta=0$ for synchronous settings and $\delta=10^{-4}$ for asynchronous settings,
and the reward satisfies 
$$
r(0,0) = r(0,1) = 0, \qquad \text{and}\qquad r(1,0) = r(1,1) = 1.
$$
Since $p>\sigma$ and $\delta<\sigma$ for all parameter settings considered, the optimal robust $Q$-function is given by
\begin{align*}
Q^{\star,\sigma}(0,0) = Q^{\star,\sigma}(0,1) = 0,\qquad Q^{\star,\sigma}(1,0) = Q^{\star,\sigma}(1,1) = \frac{1}{1-\gamma(p-\sigma)}.
\end{align*}
In all experiments, we set $p=(4\gamma-1)/(3\gamma)$ and fix the target accuracy at $\varepsilon=0.5$. For a given budget $T$ of iterations, a trial is successful if $\|Q_T-Q^{\star,\sigma}\|_\infty\le\varepsilon$. For each parameter setting, we use bisection to estimate the minimum budget at which at least $48$ of $50$ Monte Carlo trials are successful. We report this budget as the empirical sample complexity divided by $SA$.
Parameters $\beta$, $\eta_k$, $T_k$, and $K$ are set as they are in Section \ref{sec:alg} and Theroem \ref{thm:infinite-horizon}, Corollary \ref{thm:infinite-horizon-syn} with $C_1=100$, and $\varepsilon_\tau=0.5(1-\gamma)\varepsilon$.
}

\begin{figure}[t]
    \centering
    \begin{subfigure}[b]{0.48\textwidth}
        \centering
        \includegraphics[width=\textwidth]{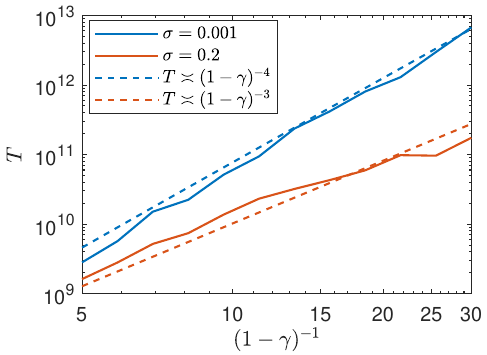}
        \caption{}
        \label{fig:exp_syn_gamma}
    \end{subfigure}
    \hfill
    \begin{subfigure}[b]{0.48\textwidth}
        \centering
        \includegraphics[width=\textwidth]{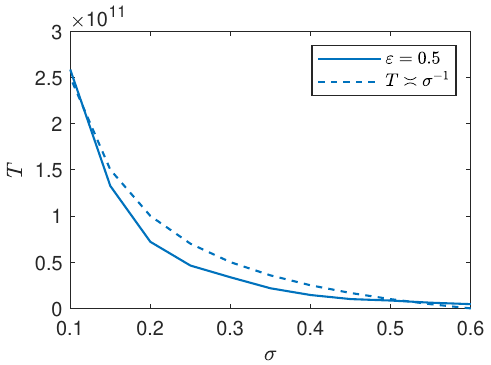}
        \caption{}
        \label{fig:exp_syn_sigma}
    \end{subfigure}
    \caption{Sample complexity of the proposed algorithm with respect to the effective horizon $1/(1-\gamma)$ and $\sigma$ in asynchronous settings. Target error $\varepsilon=0.5$.}
    \label{fig:exp_asyn}
\end{figure}

\revision{
We first examine the dependence of \(T\) on the effective horizon \(1/(1-\gamma)\) in asynchronous settings. We consider a small and a medium uncertainty level, namely \(\sigma=0.001\) and \(\sigma=0.2\), which correspond to the regimes \(\sigma\lesssim 1-\gamma\) and \(\sigma\gtrsim 1-\gamma\), respectively. We vary $(1-\gamma)^{-1}$ from $5$ to $30$.
The results are shown in Figure \ref{fig:exp_asyn}(a). It is observed that when \(\sigma\) is small, \(T\) exhibits a trend consistent with \(1/(1-\gamma)^4\) as \((1-\gamma)^{-1}\) varies, whereas for $\sigma=0.2$, the observed trend is consistent with $(1-\gamma)^{-3}$. These observations are consistent with our theoretical results.
}

\begin{figure}[t]
    \centering
    \begin{subfigure}[b]{0.48\textwidth}
        \centering
        \includegraphics[width=\textwidth]{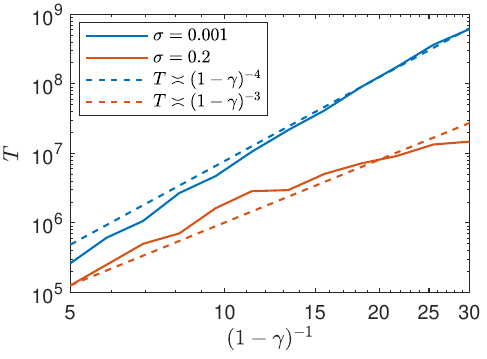}
        \caption{}
        \label{fig:exp_syn_gamma}
    \end{subfigure}
    \hfill
    \begin{subfigure}[b]{0.48\textwidth}
        \centering
        \includegraphics[width=\textwidth]{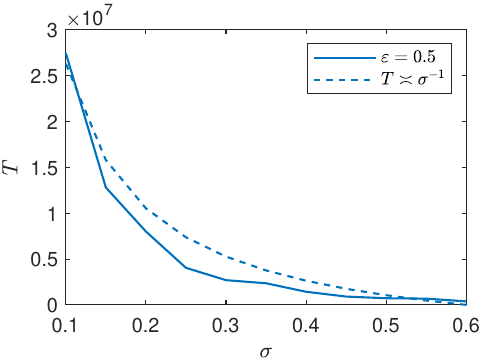}
        \caption{}
        \label{fig:exp_syn_sigma}
    \end{subfigure}
    \caption{Sample complexity of the proposed algorithm with respect to the effective horizon $1/(1-\gamma)$ and $\sigma$ in synchronous settings. Target error $\varepsilon=0.5$.}
    \label{fig:exp_syn}
\end{figure}

\revision{
We then examine the dependence of \(T\) on the uncertainty level \(\sigma\), with results shown in Figure \ref{fig:exp_asyn}(b). We fix the discount factor \(\gamma=0.95\) and vary \(\sigma\) from \(0.1\) to \(0.6\). The dotted line represents the curve \(c_1/\sigma+c_2\), with fitted coefficients \(c_1\) and \(c_2\). As \(\sigma\) increases, the empirical complexity \(T\) is well approximated by the fitted curve, which aligns with our theoretical results.
Results in synchronous settings are shown in Figure \ref{fig:exp_syn}. The observed trends are consistent with those in asynchronous settings, which further validates our theoretical results.
}

\bibliographystyle{apalike}
\bibliography{sample}

\end{document}